\RequirePackage[l2tabu,orthodox]{nag}

\documentclass[headsepline,footsepline,footinclude=false,oneside,fontsize=11pt,paper=a4,listof=totoc,bibliography=totoc,DIV=12]{scrbook} 

\PassOptionsToPackage{table,svgnames,dvipsnames}{xcolor}

\usepackage[utf8]{inputenc}
\usepackage[T1]{fontenc}
\usepackage[sc]{mathpazo}
\usepackage[ngerman,english]{babel} 
\usepackage[autostyle]{csquotes}
\usepackage[%
  backend=biber,
  url=true,
  style=numeric, 
  sorting=none, 
  maxnames=4,
  minnames=3,
  maxbibnames=99,
  giveninits,
  uniquename=init]{biblatex} 
\usepackage{graphicx}
\usepackage{scrhack} 
\usepackage{listings}
\usepackage{lstautogobble}
\usepackage{tikz}
\usepackage{pgfplots}
\usepackage{pgfplotstable}
\usepackage{booktabs} 
\usepackage[final]{microtype}
\usepackage{caption}
\usepackage[hidelinks]{hyperref} 
\usepackage{ifthen} 
\usepackage{pdftexcmds} 
\usepackage{paralist} 
\usepackage{subfig} 
\usepackage{siunitx} 
\usepackage{multirow} 
\usepackage{array} 
\usepackage{makecell} 
\usepackage{pdfpages} 
\usepackage{adjustbox} 
\usepackage{tablefootnote} 
\usepackage{threeparttable} 

\usepackage{amssymb}
\usepackage{pifont}

\usepackage[acronym,xindy,toc]{glossaries} 
\makeglossaries

\newacronym[shortplural={D$_{\text{dye}}$}, longplural={donor dye, ex. Alexa 488}]{ddye}{D$_{\text{dye}}$}{donor dye, ex. Alexa 488}

\newacronym[description={\glslink{r0}{F\"{o}rster distance}}]{R0}{$R_{0}$}{F\"{o}rster distance}

\newglossaryentry{r0}
{
  name=\glslink{R0}{\ensuremath{R_{0}}},
  text=F\"{o}rster distance,
  description={F\"{o}rster distance, where 50\% ...}, 
  sort=R
}

\newglossaryentry{kdeac}
{
  name=\glslink{R0}{\ensuremath{k_{DEAC}}},
  text=$k_{DEAC}$, 
  description={is the rate of deactivation from ... and emission)}, 
  sort=k
}

\newacronym[shortplural={TUM}, longplural={Technical University of Munich}]{tuma}{TUM}
{Technical University of Munich}

\newglossaryentry{tum}
{
  name = TUM,
  description = {A university in Germany, Bavaria, in the city of Munich},
  plural = TUM
}

\newglossaryentry{computer}
{
  name=computer,
  description={is a programmable machine that receives input,
               stores and manipulates data, and provides
               output in a useful format}
} 

\bibliography{bibliography}

\setkomafont{disposition}{\normalfont\bfseries} 
\BeforeTOCHead[toc]{{\cleardoublepage\pdfbookmark[0]{\contentsname}{toc}}}

\definecolor{TUMBlue}{HTML}{0065BD}
\definecolor{TUMSecondaryBlue}{HTML}{005293}
\definecolor{TUMSecondaryBlue2}{HTML}{003359}
\definecolor{TUMBlack}{HTML}{000000}
\definecolor{TUMWhite}{HTML}{FFFFFF}
\definecolor{TUMDarkGray}{HTML}{333333}
\definecolor{TUMGray}{HTML}{808080}
\definecolor{TUMLightGray}{HTML}{CCCCC6}
\definecolor{TUMAccentGray}{HTML}{DAD7CB}
\definecolor{TUMAccentOrange}{HTML}{E37222}
\definecolor{TUMAccentGreen}{HTML}{A2AD00}
\definecolor{TUMAccentLightBlue}{HTML}{98C6EA}
\definecolor{TUMAccentBlue}{HTML}{64A0C8}

\pgfplotsset{compat=newest}
\pgfplotsset{
  cycle list={TUMBlue\\TUMAccentOrange\\TUMAccentGreen\\TUMSecondaryBlue2\\TUMDarkGray\\},
}

\graphicspath{
    {logos/}
    {figures/}
}

\newcolumntype{P}[1]{>{\centering\arraybackslash}p{#1}} 
\newcolumntype{M}[1]{>{\centering\arraybackslash}m{#1}} 
\newcolumntype{L}[1]{>{\raggedright\arraybackslash}m{#1}} 
\newcolumntype{R}[1]{>{\raggedleft\arraybackslash}m{#1}} 
\usepackage{algorithm}
\usepackage{algpseudocode}

\usepackage{booktabs}
\usepackage{multirow}
\usepackage{graphicx}
\usepackage{adjustbox}   
\usepackage{makecell}    
\usepackage{subcaption}  

\newcommand*{\getUniversity}{Technische Universität München}
\newcommand*{\getFaculty}{School of Computation, Information and Technology - Informatics}
\newcommand*{\getTitle}{3D Multi-View Stylization with Pose-Free Correspondences Matching for Robust 3D Geometry Preservation}
\newcommand*{\getTitleGer}{3D-Multi-View-Stilisierung mit Posefreiem Korrespondenzabgleich für Robuste 3D Geometrieerhaltung}
\newcommand*{\getAuthor}{Shirsha Bose}
\newcommand*{\getDoctype}{Master's Thesis in Informatics}
\newcommand*{\getSupervisor}{Muhammad Qadeer Ahmed Khan, Abhishek Saroha}
\newcommand*{\getAdvisor}{Prof. Dr. Daniel Cremers}
\newcommand*{\getSubmissionDate}{11/01/2026}
\newcommand*{\getSubmissionLocation}{Munich}

\begin{document}

\selectlanguage{english}

\pagenumbering{alph}
\begin{titlepage}
  \oddsidemargin=\evensidemargin\relax
  \textwidth=\dimexpr\paperwidth-2\evensidemargin-2in\relax
  \hsize=\textwidth\relax

  \centering

  \IfFileExists{logos/tum.pdf}{%
    \includegraphics[height=20mm]{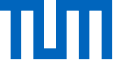}
  }{%
    \vspace*{20mm}
  }

  \vspace{5mm}
  {\huge\MakeUppercase{\getFaculty{}}}\\

  \vspace{5mm}
  {\large\MakeUppercase{\getUniversity{}}}\\

  \vspace{20mm}
  {\Large \getDoctype{}}

  \vspace{15mm}
  \makeatletter
  \ifthenelse{\pdf@strcmp{\languagename}{english}=0}
  {\huge\bfseries \getTitle{}}
  {\huge\bfseries \getTitleGer{}}
  \makeatother

  \vspace{15mm}
  {\LARGE \getAuthor{}}

  \IfFileExists{logos/faculty.png}{%
    \vfill{}
    \includegraphics[height=20mm]{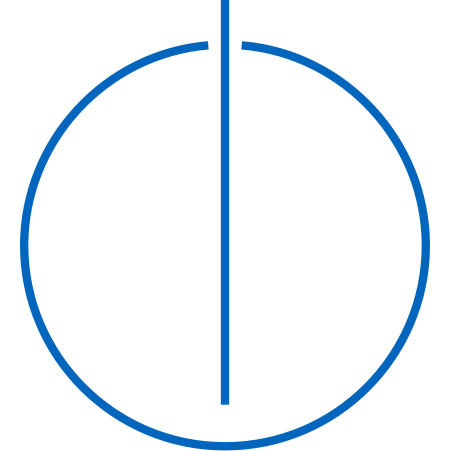}
  }{}
\end{titlepage}

\frontmatter{}

\begin{titlepage}
  \centering

  \IfFileExists{logos/tum.pdf}{%
    \includegraphics[height=20mm]{logos/tum.pdf}
  }{%
    \vspace*{20mm}
  }

  \vspace{5mm}
  {\huge\MakeUppercase{\getFaculty{}}}\\

  \vspace{5mm}
  {\large\MakeUppercase{\getUniversity{}}}\\

  \vspace{20mm}
  {\Large \getDoctype{}}

  \makeatletter
  \vspace{15mm}
  \ifthenelse{\pdf@strcmp{\languagename}{english}=0}
  {
  {\huge\bfseries \getTitle{}}

  \vspace{10mm}
  {\huge\bfseries \foreignlanguage{ngerman}{\getTitleGer{}}}
  }
  {
  {\huge\bfseries \getTitleGer{}}

  \vspace{10mm}
  {\huge\bfseries \foreignlanguage{english}{\getTitle{}}}
  }
  \makeatother

  \vspace{15mm}
  \begin{tabular}{l l}
    Author:          & \getAuthor{} \\
    Supervisor:      & \getSupervisor{} \\
    Advisor:         & \getAdvisor{} \\
    Submission Date: & \getSubmissionDate{} \\
  \end{tabular}

  \IfFileExists{logos/faculty.png}{%
    \vfill{}
    \includegraphics[height=20mm]{logos/faculty.png}
  }{}
\end{titlepage}

\cleardoublepage{}

\thispagestyle{empty}
\vspace*{0.8\textheight}
\noindent
\makeatletter
\ifthenelse{\pdf@strcmp{\languagename}{english}=0}
{I confirm that this \MakeLowercase{\getDoctype{}} is my own work and I have documented all sources and material used.}
{Ich versichere, dass ich diese \getDoctype{} selbstständig verfasst und nur die angegebenen Quellen und Hilfsmittel verwendet habe.}
\makeatother

\vspace{15mm}
\noindent
\getSubmissionLocation{}, \getSubmissionDate{} \hspace{50mm} \getAuthor{}

\cleardoublepage{}

\makeatletter
\ifthenelse{\pdf@strcmp{\languagename}{english}=0}
{\addcontentsline{toc}{chapter}{Acknowledgments}}
{\addcontentsline{toc}{chapter}{Danksagungen}}
\makeatother
\thispagestyle{empty}

\vspace*{20mm}

\begin{center}
\makeatletter
\ifthenelse{\pdf@strcmp{\languagename}{english}=0}
{\usekomafont{section} Acknowledgments}
{\usekomafont{section} Danksagungen}
\makeatother
\end{center}

\vspace{10mm}

I would like to express my sincere gratitude to Prof.\ Dr.\ Daniel Cremers at the Technical University of Munich, whose Computer Vision Lab provided an outstanding research environment for this thesis. The courses and academic training I received in the group, particularly in 2D computer vision, deep learning, and the foundations of 3D vision and learning were essential in building the conceptual and practical groundwork required for this thesis. I also gratefully acknowledge the computational resources and infrastructure provided by the lab, which enabled extensive experimentation and development throughout the project.

I am especially thankful to my supervisors, Muhammad Qadeer Ahmed Khan and Abhishek Saroha, for their continuous guidance, constructive feedback, and critical analysis at every stage of the work. Their ability to identify key technical challenges, suggest clear directions when I was stuck, and encourage rigorous thinking was invaluable. Many important design decisions and improvements in the final framework emerged directly from their thoughtful discussions and detailed feedback.

I would also like to thank my friends and colleagues for their support during my thesis period for the motivating conversations, encouragement during difficult phases, and for making the process more enjoyable.

Finally, I am deeply grateful to my parents for their unwavering belief in me. Their constant support, patience, and encouragement gave me the stability and confidence to persist through the demanding phases of this work. This thesis would not have been possible without them.

\cleardoublepage{}
\chapter{\abstractname}

Artistic style transfer is well studied for single images and videos, but extending it to multi-view captures of a 3D scene remains difficult because stylization can break the cross-view correspondences that geometry-aware pipelines rely on. Independent per-view stylization often causes view-dependent texture drift, warped edges, and inconsistent shading, which in turn degrades SLAM tracking, monocular depth prediction, and multi-view reconstruction. This thesis targets multi-view stylization that remains geometrically usable for downstream 3D tasks, without assuming camera poses or an explicit 3D representation during training.

We introduce a feed-forward stylization network trained with per-scene test-time optimization under a composite objective that couples appearance transfer with geometry preservation. Stylization is driven by an AdaIN-inspired feature-statistics loss computed from a frozen VGG-19 encoder, matching multi-scale channel-wise moments of the output to a reference style image. To stabilize structure across viewpoints, we propose a correspondence-based geometric consistency loss using SuperPoint detections and SuperGlue matching: descriptors from a stylized anchor view are constrained to remain consistent with descriptors from the original multi-view set under confident matches, directly regularizing repeatable local structure. We also impose a depth-preservation loss using a frozen MiDaS/DPT model, matching normalized depth predictions between original and stylized views; a dataset-to-style global color alignment step reduces depth-model domain shift. A staged weight schedule (warmup + ramp) progressively introduces geometry and depth constraints to avoid early over-regularization.

We evaluate on multi-view scenes from Tanks and Temples and Mip-NeRF 360 using both image-level and reconstruction-level metrics. Static measures quantify style adherence and structure retention via Color Histogram Distance (CHD) and a DINO-based Structure Distance (DSD). For 3D consistency, we use monocular DROID-SLAM to compare similarity-aligned trajectories (ATE/RTE) and back-projected point clouds via symmetric Chamfer distance. Across ablations, correspondence and depth regularization reduce structural distortion and substantially improve SLAM stability and reconstructed geometry; on scenes with MuVieCAST baselines, our method yields markedly stronger trajectory and point-cloud consistency while maintaining competitive stylization.

\makeatletter
\ifthenelse{\pdf@strcmp{\languagename}{english}=0}
{\renewcommand{\abstractname}{Kurzfassung}}
{\renewcommand{\abstractname}{Abstract}}
\makeatother




\makeatletter
\ifthenelse{\pdf@strcmp{\languagename}{english}=0}
{\renewcommand{\abstractname}{Abstract}}
{\renewcommand{\abstractname}{Kurzfassung}}
\makeatother
\microtypesetup{protrusion=false}
\tableofcontents{}
\microtypesetup{protrusion=true}

\mainmatter{}

\chapter{Introduction}

\section{Motivation}
Artistic neural style transfer studies how to render a photograph in the appearance of an artwork while preserving the underlying scene content. Early neural style transfer methods showed that deep convolutional features separate \emph{content} (high-level structure) from \emph{style} (texture statistics), enabling compelling stylizations via perceptual optimization \cite{gatys2016image}. Subsequent work introduced feed-forward networks that approximate this optimization for real-time stylization \cite{johnson2016perceptual}, and normalization-based formulations that enable arbitrary style transfer by aligning feature statistics, such as Adaptive Instance Normalization (AdaIN) \cite{huang2017arbitrary}. These approaches are highly effective for \emph{single} 2D images, where each output can be judged independently by visual plausibility.

However, when applying image style transfer to videos or multi-view captures of the same 3D scene, naive per-frame (or per-view) stylization typically introduces \emph{temporal flicker} and \emph{view-dependent artifacts}: edges bend, textures ``swim'' across surfaces, and fine structures appear or disappear depending on viewpoint. The core reason is that classic 2D objectives optimize each frame in isolation and do not explicitly enforce that the \emph{same 3D surface geometry} receives a stable appearance across views. Video style transfer methods address this partially through temporal regularization and warping-based constraints \cite{ruder2016artistic, gao2018reconet}, but they commonly assume smooth frame-to-frame motion (dense temporal overlap) and are not designed for unordered or wide-baseline multi-view imagery.

This limitation has motivated a growing body of \emph{3D-aware stylization} methods that operate on explicit or implicit 3D scene representations. Point-cloud and proxy-geometry approaches stylize a reconstructed 3D representation and then render stylized views, aiming to improve cross-view coherence \cite{cao2020psnet, huang2021learning}. Mesh-based methods like StyleMesh\cite{hollein2022stylemesh} transfer style onto reconstructed indoor meshes by optimizing textures while using depth/normal cues for 3D-aware regularization. In parallel, neural radiance field (NeRF) stylization approaches learn a radiance-field representation and optimize it to match a style while retaining multi-view consistency; notable examples include StylizedNeRF\cite{huang2022stylizednerf} (2D--3D mutual learning) and ARF\cite{zhang2022arf} (nearest-neighbor style losses for sharper artistic details). More recently, 3D Gaussian Splatting (3DGS)\cite{kerbl20233d} has become a popular real-time radiance-field alternative, and several works extend it to scene stylization by embedding or transforming features inside the Gaussian representation \cite{liu2024stylegaussian, galerne2025sgsst, kovacs2024g}.

Despite strong visual results, many 3D stylization pipelines rely on additional inputs and preprocessing: camera poses and structure-from-motion (SfM) reconstructions (for NeRF/3DGS), proxy geometry, or careful capture assumptions. Furthermore, optimization-based 3D stylization often exhibits oversmoothing or loss of high-frequency structure, especially near edges and thin structures, which are precisely the cues needed by downstream 3D vision systems. In practical scenarios (casual capture, wide baselines, sparse views), we want stylization methods that remain coherent \emph{without} requiring accurate camera poses, dense temporal sampling, or heavy reconstruction pipelines.

A second, closely related challenge is that stylization can disrupt the very image cues used in geometry estimation. SLAM and 3D multi-view reconstruction heavily depend on repeatable keypoints and matchable descriptors; stylization may alter edges, shading, and local contrast, reducing correspondence quality and destabilizing camera tracking. Likewise, stylization can shift the image statistics away from the training domain of monocular depth priors, leading to inconsistent depth estimates that negatively affect fusion. These observations motivate a stylization objective that explicitly constrains (i) \emph{feature-level correspondence stability} and (ii) \emph{depth consistency} during training.

Finally, evaluation of stylization for 3D usability remains underdeveloped. Standard style-transfer metrics such as FID \cite{heusel2017gans} capture distribution-level similarity of generated images to a target domain, but they do not directly measure multi-view geometric integrity. Recent multi-view stylization works often rely on human studies like  MuVieCAST\cite{ibrahimli2024muviecast} or per-image structure/style distances such as color-histogram distances and DINO-based structure distances in \cite{zuo2024towards} are useful, but still fundamentally \emph{single-image} or perceptual. For downstream 3D tasks, we need an evaluation protocol that measures how well the stylized views preserve the geometry needed for reconstruction and tracking.

In this thesis, we propose a pose-free multi-view stylization framework trained per-scene from multi-view RGB images. The core idea is to preserve geometry by penalizing \emph{descriptor drift} on matched correspondences: we introduce a SuperPoint\cite{detone2018superpoint}/SuperGlue\cite{sarlin2020superglue}-based multi-view geometry loss that encourages the stylized output to remain matchable across views. To further stabilize geometry-related cues, we add a depth-preservation loss using a frozen MiDaS/DPT\cite{ranftl2020towards, ranftl2021vision} teacher network and introduce a practical domain-alignment step to reduce depth-model shift. We also extend the framework to a conditional U-Net stylizer that supports \emph{multiple} styles within a single model and demonstrate generalization to unseen scenes at test time. To quantify geometric integrity after stylization, we propose an evaluation protocol based on monocular DROID-SLAM \cite{teed2021droid}: we compare camera trajectories (after similarity alignment) and compute a symmetric Chamfer distance between reconstructed point clouds.

\begin{quote}
\textit{How can we obtain multi-view stylization that is visually stylized yet preserves geometric consistency sufficiently for downstream 3D reconstruction? And how can we measure, quantitatively, to what extent geometry is preserved after stylization?}
\end{quote}

\section{Problem Statement}
Given a set of RGB images $\{I_i\}_{i=1}^N$ of a static scene captured from different viewpoints and a style reference image $S$, our goal is to learn a neural stylization function $g_{\theta}$ such that:
\begin{equation}
\hat{I}_i = g_{\theta}(I_i, S) \quad \text{for } i \in \{1,\dots,N\},
\end{equation}
(or, in the multi-style setting, $\hat{I}_i = g_{\theta}(I_i, y)$ where $y$ is a discrete style id).
The outputs $\{\hat{I}_i\}$ should satisfy the following properties:
\begin{enumerate}
    \item \textbf{Stylization:} each $\hat{I}_i$ adopts the appearance (texture statistics and artistic patterns) of the target style.
    \item \textbf{Structure preservation:} each $\hat{I}_i$ retains the semantic layout and geometric structure of $I_i$.
    \item \textbf{Multi-view consistency:} corresponding 3D surface points observed in different views should map to compatible appearances in $\{\hat{I}_i\}$, enabling stable feature correspondences and consistent geometry for downstream reconstruction.
\end{enumerate}
In other words, we aim to learn $g_{\theta}$ such that stylization is achieved \emph{without destroying the geometric information} required for multi-view matching, camera tracking, and 3D fusion.

\section{Thesis Contributions}
This thesis makes the following contributions:
\begin{enumerate}
    \item \textbf{Pose-free multi-view stylization with correspondence-regularized geometry preservation.}
    We propose a lightweight CNN stylizer trained per-scene with an AdaIN-inspired feature-statistics loss \cite{huang2017arbitrary}, augmented by a novel correspondence-based geometry loss built on SuperPoint/SuperGlue matches \cite{detone2018superpoint, sarlin2020superglue}. The geometry loss penalizes stylization-induced drift in local descriptors on matched keypoints across views, encouraging the stylized images to remain matchable and thus reducing view-dependent geometric flicker.

    \item \textbf{Depth-preserving regularization with practical domain alignment.}
    To mitigate stylization artifacts that confuse monocular depth priors, we introduce a depth-consistency loss using a frozen MiDaS/DPT model \cite{ranftl2020towards,ranftl2021vision}. We additionally apply a simple dataset-to-style color alignment step to reduce domain shift when computing depth-based supervision under stylized appearance.

    \item \textbf{A quantitative geometry-consistency evaluation protocol using DROID-SLAM.}
    We propose an evaluation method that runs DROID-SLAM \cite{teed2021droid} on original and stylized image sequences and measures:
    (i) trajectory consistency via similarity-aligned translation and rotation errors, and
    (ii) 3D structural consistency via symmetric Chamfer distance between point clouds reconstructed from DROID disparities, intrinsics, and poses.

    \item \textbf{Multi-style extension and cross-scene transfer with a conditional U-Net stylizer.}
    We extend the stylizer to a conditional U-Net with style conditioning (via conditional instance normalization / learned embeddings), enabling a single model to handle multiple styles. We show that the learned loss framework transfers to unseen scenes at test time while maintaining a strong degree of geometric stability.
\end{enumerate}

\section{Why This Direction Matters}
Beyond perceptual quality, many applications require stylized outputs to remain \emph{geometrically usable}. Stylized 3D capture for AR/VR, artistic scene documentation, and stylized novel-view synthesis all rely on stable correspondences, depth cues, and consistent structure across viewpoints. A method that produces visually pleasing single images but breaks cross-view coherence can degrade or even prevent reconstruction. 

To test geometric usability directly, we evaluate stylized sequences \emph{through} a full SLAM pipeline rather than relying only on image-level scores. We run DROID-SLAM\cite{teed2021droid} on the stylized views and report \emph{ATE/RTE} to quantify camera pose drift, along with \emph{Chamfer Distance (ChD)} between reconstructed point clouds and the reference geometry. In practice, these metrics act as a stress test: if stylization disrupts correspondences or depth cues, SLAM quickly diverges, yielding higher pose error and degraded 3D structure. This thesis therefore targets stylization that is not only artistic, but also demonstrably compatible with downstream 3D pipelines.

\section{Thesis Outline}
Chapter~2 surveys related work on neural style transfer, video/multi-view consistency, and 3D scene stylization. Chapter~3 presents the proposed methodology, including the stylizer architecture, the SuperPoint/SuperGlue geometry loss, and the MiDaS/DPT depth regularization. Chapter~4 describes the experimental setup (datasets, training details, and baselines). Chapter~5 details the evaluation metrics, including perceptual/style measures and the proposed DROID-SLAM-based geometry protocol. Chapters~6 present results, ablations, and discussion. Chapter~7 concludes with limitations and future research directions.

\chapter{Related Work}
\section{Image Style Transfer}
Image style transfer started with classic optimization-based neural style transfer(NST). \cite{gatys2016image,simonyan2014very} showed that a pretrained CNN (commonly VGG) can separate content (deep feature activations) from style (feature correlations via Gram matrices), and produce a stylized image by iterative pixel optimization that matches content features while matching style statistics across multiple layers. This formulation established the standard “content–style trade-off” and remains a conceptual baseline, but it is slow at inference because it requires many gradient steps per image.

To make style transfer practical, later work moved the optimization into training using feed-forward transformation networks. \cite{johnson2016perceptual} trained a generator with perceptual (feature-space) losses to achieve real-time stylization with results comparable to optimization-based NST, with faster inference. A key enabling detail for high-quality fast stylization is normalization design: \cite{ulyanov2016instance}
showed that replacing batch normalization with instance normalization significantly improves stylization quality and stability in feed-forward networks.

A major limitation of early fast models is that they were often tied to a fixed style or small style set. This led to arbitrary style transfer, where the style image is provided at test time. AdaIN \cite{huang2017arbitrary}
 aligns the channel-wise mean and variance of content features to those of the style features, enabling real-time arbitrary stylization with a single model. In parallel, feature-transform approaches such as WCT \cite{li2017universal} match higher-order statistics (feature covariance) via whitening and coloring transforms, improving quality for diverse styles without training a separate network per style. These methods emphasize that much of “style” can be captured by aligning feature statistics, with the remaining challenge being how to preserve spatial structure.

To better transfer local style patterns (e.g., brush strokes) while preserving structure, later methods introduced attention and transformers. SANet \cite{park2019arbitrary} uses a style-attention module to inject locally relevant style features according to the content’s spatial layout, aiming to balance global style and local texture richness. Transformer-based stylization such as StyTr² \cite{deng2022stytr2} further targets limitations of CNN locality by modeling long-range dependencies between content and style tokens, improving global coherence in some cases.

A separate branch studies photorealistic style transfer, where both content and “style” are photos and the goal is to change appearance (color tone, lighting mood) without painterly distortions. Deep Photo Style Transfer \cite{luan2017deep} explicitly highlights that standard NST can bend straight lines and introduce wavy textures, and proposes additional constraints to preserve photorealism.

Most recently, diffusion models have been adopted for style transfer because of their strong generative priors and ability to produce rich textures. InST \cite{zhang2023inversion} frames style transfer through diffusion inversion and learns style guidance from a single painting to drive stylized synthesis without requiring a complex textual description. Diffusion-based methods often deliver strong perceptual results, but they introduce new trade-offs around compute, controllability, and strict structural preservation issues that become even more critical when extending style transfer to videos or multi-view/3D settings.

\section{Multi-View Correspondences}
Given two images of the same scene with sufficient overlap, a local feature pipeline detects keypoints and computes descriptors, producing tentative correspondences
$\mathcal{M}=\{(x^A_k,x^B_k)\}_{k=1}^{K}$.
Such correspondences are a cornerstone of multi-view geometry: they enable robust estimation of epipolar relations and camera motion (often with RANSAC for outlier rejection), and they are the basic ``glue'' behind SfM/SLAM systems that triangulate and refine 3D structure from multi-view observations \cite{lowe2004distinctive,fischler1981random,schonberger2016structure}.
In the context of stylization, correspondences become equally important as a \emph{consistency signal}: if $x^A_k$ and $x^B_k$ observe the same physical point, the stylized appearance around these locations should remain compatible, otherwise downstream 3D reconstruction can lose track due to unstable visual evidence across views.

A practical way to exploit matches for stylization is to define a correspondence-consistency regularizer that compares local features at matched locations in the stylized outputs, e.g.,
\begin{equation}
\mathcal{L}_{\text{corr}}
= \frac{1}{|\mathcal{M}|}\sum_{(x^A,x^B)\in\mathcal{M}}
\rho\!\left(\left\|\phi(\hat{I}^A)(x^A)-\phi(\hat{I}^B)(x^B)\right\|_2\right),
\end{equation}
where $\phi(\cdot)$ is a feature extractor (descriptor map) and $\rho(\cdot)$ is a robust penalty (e.g., Huber) to reduce sensitivity to occasional mismatches.
Unlike purely photometric constraints, correspondence-based constraints can remain meaningful even when stylization introduces large color/texture changes, because they tie together \emph{semantically corresponding} locations rather than raw RGB values.

\paragraph{Learned keypoints and descriptors: SuperPoint\cite{detone2018superpoint}.}
Classical pipelines rely on hand-crafted features such as SIFT \cite{lowe2004distinctive}, which are robust but can degrade when appearance is heavily transformed.
SuperPoint is a learned alternative that jointly predicts keypoints and descriptors using a self-supervised training strategy (homographic adaptation), yielding repeatable interest points and compact descriptors suitable for real-time matching. For stylization pipelines, SuperPoint is attractive because it tends to provide stable keypoints under viewpoint changes and moderate illumination variation, forming a stronger basis for enforcing cross-view constraints than raw pixel-space alignment.

\paragraph{Learned matching with context: SuperGlue\cite{sarlin2020superglue}.}
Even with good descriptors, naive nearest-neighbor matching can be brittle in repetitive textures or low-texture regions.
SuperGlue improves robustness by treating matching as a context-aware assignment problem: it uses a graph neural network with attention to reason over sets of keypoints and their descriptors, and solves a differentiable matching objective via optimal transport (Sinkhorn normalization) while allowing ``no-match'' outcomes. This ability to model global context and explicitly reject ambiguous matches is particularly relevant for stylization, where appearance changes may shift local gradients/texture phases and increase matching ambiguity. In practice, higher-quality correspondences translate into more reliable multi-view constraints, which better preserves cross-view object correspondence and stabilizes downstream 3D tasks.

\paragraph{Related learned correspondence families.}
The broader literature includes detector-free matchers that directly predict dense/semidense correspondences, such as LoFTR\cite{sun2021loftr}, which replaces explicit keypoint detection with transformer-based coarse-to-fine matching and can be strong in low-texture regimes.
More recent work such as LightGlue\cite{lindenberger2023lightglue} focuses on making learned matching efficient and scalable while retaining SuperGlue-like context reasoning.
These developments reinforce an important theme: multi-view consistency objectives benefit disproportionately from correspondence quality, and modern learned matchers often outperform classical heuristics when scenes include repeated patterns, motion blur, or challenging viewpoint changes.

\paragraph{Limitations under stylization and implications.}
Although learned matchers are significantly more robust than classical ones, correspondence estimation can still fail under strong non-photorealistic effects (e.g., large texture hallucination, edge suppression, or highly abstract brush patterns).
When correspondences break, enforcing consistency via warping or match-based losses can introduce artifacts (e.g., ghosting or incorrect constraints), motivating robust penalties, confidence thresholding, and careful selection of match sets.
Inspite of these limitations, as we treat SuperPoint/SuperGlue consistency as an \emph{explicit loss term} that the stylizer must learn to satisfy, therefore we introduce it \emph{gradually} via a warmup--ramp schedule. Concretely, we first optimize the stylizer using only the appearance-driven losses so that outputs become reasonably stylized and contextually stable. We then ramp up the correspondence loss weight smoothly, so the network learns to preserve matchable structures \emph{without destabilizing early training} when correspondences are most fragile. This staged optimization acts as a practical safeguard: it reduces the risk of the geometry loss failure or overpowering the style objective in the initial iterations, while still converging to solutions where stylization and cross-view geometric consistency are jointly satisfied.

\section{Video Style Transfer and Temporal Consistency}
Applying an image stylization model independently to each frame typically produces \textbf{temporal flicker} because the stylizer is not constrained to make consistent decisions across time (e.g., texture strokes ``crawl'' and colors fluctuate).
Early video style transfer methods addressed this by adding \textbf{temporal consistency losses} based on \textbf{optical flow warping}: the stylized output at time $t\!-\!1$ is warped into frame $t$, and the stylized frame $t$ is encouraged to match the warped prediction (often with occlusion handling and improved initialization).
This formulation was popularized by \cite{ruder2016artistic, ruder2018artistic}, who extend neural style transfer with flow-based temporal regularization to stabilize appearance over time.

Later work aimed to make temporally consistent video stylization \textbf{real-time} and more robust.
ReCoNet\cite{gao2018reconet} introduces a feed-forward video stylization network with temporal losses applied both at the output level (including warping constraints that account for luminance changes) and at intermediate feature levels to stabilize traceable regions across frames.
Another direction is to avoid optical flow at inference: Learning Blind Video Temporal Consistency\cite{lai2018learning} trains a recurrent model with \textbf{short-term and long-term} temporal objectives that takes the original video and the per-frame processed video as inputs and outputs a temporally stabilized sequence, enabling real-time deployment without flow computation at test time.

A complementary family reduces flicker via keyframe-based propagation. EbSynth\cite{jamrivska2019stylizing} (“Stylizing Video by Example”) takes one or more user-stylized keyframes and propagates their appearance to other frames, prioritizing high fidelity to the exemplar and user control; it can look extremely stable when correspondences are good, but still benefits from re-anchoring with additional keyframes and can fail when matching/warping becomes unreliable.

Despite these advances, temporal-consistency methods have two important limitations for our setting.
First, \textbf{optical flow can be unreliable} under occlusions, fast motion, motion blur, and large viewpoint changes, which can inject incorrect supervision into warping-based losses and cause ghosting or over-smoothing.
More fundamentally, temporal consistency only enforces stability along the \textbf{time axis} of a single video and does not directly enforce \emph{multi-view} geometric consistency for a set of images representing a 3D scene (e.g., wide-baseline viewpoints or unordered view sets).
As a result, a method can be temporally stable yet still break the cross-view correspondences needed for downstream 3D reconstruction, motivating explicit geometric constraints beyond frame-to-frame smoothing.

\section{NeRF-based stylization.}
NeRF-based approaches embed style into a \emph{radiance field} so that renderings from arbitrary camera poses remain view-consistent by construction: every view is rendered from the same 3D representation rather than stylized independently in 2D.
A common baseline is to first reconstruct a photorealistic NeRF from posed multi-view captures and then fine-tune the appearance/color components with style losses computed on rendered views. This often yields strong long-range consistency, but it inherits the cost and fragility of NeRF optimization (per-scene training and volumetric rendering) and becomes brittle when capture coverage is limited.

Several papers differ in \emph{how} they inject style and \emph{what} parts of the NeRF they allow to change.
\textbf{StylizedNeRF}\cite{huang2022stylizednerf} addresses the 2D--3D domain gap by replacing the NeRF color head with a style module and training it in a \emph{2D--3D mutual-learning} loop: consistency priors are distilled from NeRF into a 2D stylizer via a consistency loss, while a mimic loss aligns the 2D and 3D stylization outputs, with style-conditioned latent codes to model ambiguity.
For spatial control, \textbf{Locally Stylized NeRF}\cite{pang2023locally} separates geometry and appearance (e.g., via hash-grid encodings) and performs stylization by optimizing only the appearance branch while keeping geometry fixed, which tends to stabilize multi-view results and supports region-level controllability.

Recent pipelines often leverage \textbf{diffusion priors} for richer styles than classic perceptual losses, but must explicitly fight diffusion stochasticity and multi-view drift.
\textbf{Instruct-NeRF2NeRF}\cite{haque2023instruct} alternates between rendering training views, editing them with an instruction-conditioned diffusion model, and continuing NeRF training on the updated images, effectively performing iterative ``dataset updates'' that consolidate 2D edits into a coherent 3D field.
\textbf{ViCA-NeRF}\cite{dong2023vica} promotes view consistency more explicitly by propagating edits using NeRF-derived depth correspondences and aligning diffusion latents between edited and unedited images, using a two-stage edit/refine procedure.
For stylization, \textbf{3D Style Transfer from Style-Aligned Multi-View Images (Style-NeRF2NeRF)\cite{fujiwara2024style}} follows a ``generate-then-train'' strategy: it first generates style-aligned multi-view images via depth-conditioned diffusion (with attention sharing) and then refines the NeRF using distribution/feature matching to transfer style into the radiance field.
Text-driven stylization/editing is another branch; for example, \textbf{NeRF-Art}\cite{wang2023nerf} stylizes a pre-trained NeRF from text prompts and discusses the challenge of jointly changing appearance and geometry without introducing density/geometry artifacts, motivating dedicated regularization.

Finally, the literature splits between \textbf{per-scene optimization} (best quality but slow) and \textbf{generalizable/feed-forward} models (faster and more scalable).
\textbf{StyleRF}\cite{liu2023stylerf} targets zero-shot 3D style transfer by transforming features inside a radiance-field representation, trading some peak scene-specific fidelity for generality across styles.
\textbf{FPRF}\cite{kim2024fprf} and \textbf{G3DST}\cite{meric2024g3dst} reduce or remove per-scene/per-style optimization while introducing explicit multi-view consistency mechanisms (e.g., 3D feature-space transfer, flow-based consistency).
For photorealistic ``look transfer,'' \textbf{IPRF}\cite{koh2025intrinsic} leverages intrinsic decomposition (albedo vs.\ shading) to better preserve structure and illumination behavior compared to purely RGB-space objectives.
When capture coverage is sparse, reconstruction artifacts can dominate stylization; \textbf{Stylizing Sparse-View 3D Scenes with Hierarchical Neural Representation}\cite{wang2025stylizing} proposes a coarse-to-fine representation and optimization strategy (e.g., content-strength annealing) to better disentangle content semantics from style textures.

Most pipelines assume accurate camera poses (and often intrinsics) to reconstruct the scene before stylization; when poses are unavailable, an additional SfM/SLAM stage is required and its errors propagate directly into the stylized result. NeRF optimization is also computationally heavy and typically per-scene, requiring iterative volumetric rendering and long runtimes compared to lightweight feed-forward stylizers. Finally, when the reconstruction is imperfect due to sparse coverage, low texture regions, structural peculiarities, or occlusion—geometry artifacts (e.g., floaters or depth layering errors) become embedded in the representation and can be amplified by stylization, which is undesirable when the stylized views must remain matchable and usable for downstream 3D reconstruction.


\section{Gaussian-splatting-based stylization.}
Methods in this family operate on an explicit \textbf{3D Gaussian Splatting (3DGS)}\cite{kerbl20233d} scene representation, where a scene is encoded as a set of anisotropic 3D Gaussians and rendered through a fast, differentiable splatting/rasterization pipeline.
Compared to NeRF-style volumetric rendering, 3DGS typically enables \textbf{much faster training and real-time rendering}, making stylization and editing workflows more interactive while preserving \textbf{multi-view consistency by construction} (all views are rendered from the same 3D primitive set).

In practice, the main challenge is that 3DGS \textbf{tightly couples appearance and geometry}: ``color-only'' updates can underfit continuous, high-frequency textures or create artifacts because the discrete Gaussian set may not have sufficient capacity where stylized detail is needed.
Consequently, many approaches introduce \textbf{geometry-aware control} (e.g., filtering problematic Gaussians, densification/splitting, or depth/structure regularization) to better support stylized texture while keeping the scene stable.
For instance, \textbf{StylizedGS}\cite{zhang2025stylizedgs} combines nearest-neighbor feature matching style losses with depth preservation and regularization, and further enables controllability such as region/scale control.
\textbf{G-Style}\cite{kovacs2024g} argues that fixed geometry is often insufficient and proposes splitting Gaussians guided by stylization gradients to increase detail where necessary.
Similarly, \textbf{ReGS}\cite{mei2024regs} motivates that appearance-only optimization is insufficient for reference textures and introduces a texture-guided mechanism that adaptively adjusts responsible Gaussians while using depth regularization to preserve geometry.

A second line of work targets \textbf{feed-forward or near-instant style switching}.
\textbf{StyleGaussian}\cite{liu2024stylegaussian} embeds 2D VGG features into Gaussians, applies a style transform, and decodes back to images, reporting interactive stylization while maintaining multi-view consistency.
\textbf{A3GS}\cite{fang2025a3gs} moves toward \textbf{zero-shot, feed-forward} 3DGS stylization (seconds per scene) using graph-based feature aggregation on the Gaussian structure, aiming to avoid per-style optimization.

3D Gaussian-splatting-based stylization inherits several constraints that are misaligned with our pose-free, image-only objective. First, most 3DGS pipelines require accurate camera poses (typically from SfM/SLAM) to build the Gaussian scene in the first place; if pose estimation is unreliable or unavailable, the entire stylization pipeline becomes brittle. Second, because appearance and geometry are tightly coupled in the Gaussian parameters, strong stylization gradients often interact with densification/splitting heuristics and regularizers, which can lead to over blurring, loss of sharp edges, or geometry “drift” (e.g., floaters) when the method tries to accommodate high-frequency artistic textures. Third, many approaches remain per-scene optimization procedures with non-trivial memory/runtime costs and tuning (densification schedules, pruning thresholds, depth regularization strength), which limits scalability compared to a lightweight feed-forward stylizer trained directly on multi-view images. In contrast, our approach enforces feature correspondences and depth stability directly in image space without requiring explicit 3D reconstruction or camera poses during training, while still evaluating downstream geometric usability via SLAM-based metrics.

\section{Multi-View and 3D-Aware Stylization}
Multi-view stylization goes beyond producing a visually pleasing stylized image: it must preserve \textbf{cross-view coherence} so that the resulting set of stylized views still describes a single, stable 3D scene. This matters because many downstream tasks---feature matching, MVS/SLAM, point cloud fusion, mesh extraction, and novel-view synthesis---implicitly assume that appearance changes across views are explained primarily by geometry and viewpoint, not by view-dependent ``style drift.'' A practical way to address this (without committing to a full 3D scene model) is to operate directly on multi-view images but add explicit \textbf{multi-view consistency constraints} so stylization decisions are coupled across viewpoints.

\paragraph{MuVieCAST\cite{ibrahimli2024muviecast} (2D-first multi-view consistent stylization baseline).}
MuVieCAST is a representative 2D-first framework designed for \emph{multi-view consistent} artistic style transfer, and it serves as our primary state-of-the-art baseline. The method couples a feed-forward image stylizer (TransferNet) with content/style feature extraction and an explicit \emph{multi-view stereo (MVS) geometry learning module}, using \emph{calibrated multi-view images and camera poses} to estimate scene geometry and enforce cross-view consistency during training. A further practical aspect of MuVieCAST is that, besides content and style losses, it introduces dedicated geometry-guidance terms (e.g., gradient/edge-based constraints and MVS derived depth/volumetric supervision) to preserve the dominant geometric structures while stylizing.
At the same time, this design implies limitations that are important for our setting. First, the MVS-based consistency pathway fundamentally depends on pose availability and reasonable geometric estimates; this can be restrictive in scenarios where poses are missing, noisy, or difficult to compute reliably. Second, enforcing multi-view agreement through MVS depth/volumes and strong edge/gradient preservation can bias optimization toward conservative image changes: in challenging viewpoints (wide baselines, occlusions, repeated textures, low-texture regions), geometry supervision and edge constraints can become fragile or overly constraining, which may trade off against stylization strength and may not directly optimize \emph{matchability} for correspondence-driven 3D pipelines. 

\paragraph{Diffusion-based multi-view stylization (strong priors, harder determinism).}
Diffusion models provide strong image priors and can yield high-quality stylization, but multi-view consistency is challenging because \textbf{stochastic generation} can introduce subtle view-to-view drift (e.g., texture phase shifts) that may be perceptually acceptable in isolation yet harmful for correspondence-heavy 3D tasks.
\textbf{OSDiffST}\cite{zuo2024towards} addresses multi-view style transfer by adapting a pre-trained one-step diffusion model (SD-Turbo) using a \textbf{vision-conditioning} module to encode the reference style and \textbf{LoRA} for parameter-efficient adaptation.
It further introduces explicit losses for \textbf{color alignment} and \textbf{structure preservation} to reduce view-dependent changes while maintaining stylization quality.
The trade-off is that diffusion-based pipelines are typically heavier in compute/memory than lightweight feed-forward stylizers, and achieving \textbf{deterministic} multi-view consistency remains difficult: small inconsistencies in fine texture or local details can still accumulate across views and degrade 3D reconstruction even when the stylization looks convincing per frame.


\section{Positioning of This Thesis: Why Our Method Helps}
We position the proposed approach as a \textbf{geometry-aware, pose-free multi-view stylization framework} that is designed to remain usable for downstream 3D pipelines while still producing strong artistic appearance. Concretely, our method offers:

\begin{itemize}
    \item \textbf{Lightweight and practical optimization:} instead of relying on heavy 3D scene optimization or iterative diffusion sampling, we train a small feed-forward stylizer (Residual CNN or U-Net) per scene/style using only frozen auxiliary networks (VGG, MiDaS/DPT, SuperPoint/SuperGlue). This keeps training stable and computationally feasible while avoiding the overhead of volumetric rendering or denoising trajectories.

    \item \textbf{Explicit geometry preservation without camera poses:} unlike pose-dependent multi-view supervision, we enforce cross-view stability directly in image space. The SuperPoint/SuperGlue correspondence loss $\mathcal{L}_{SG}$ penalizes drift in matchable local structure (corners, edges, repeated texture cues), while the depth consistency loss $\mathcal{L}_{Depth}$ discourages stylization-induced appearance changes that destabilize monocular depth predictions. Together, these losses target the exact failure modes that break SLAM/3D reconstruction under stylization.

    \item \textbf{Generalization beyond a single style or scene:} although the base formulation optimizes per scene/style, the same loss design extends naturally to multi-style training via conditional normalization (conditional U-Net), enabling a \emph{single} stylizer to render multiple artistic styles from a style ID. Moreover, our conditional model exhibits transfer to unseen scenes, suggesting that the network learns style-specific transformations while retaining a persistent bias toward preserving high-frequency geometric cues required for correspondence and reconstruction.

    \item \textbf{Direct evaluation for 3D usability:} rather than judging stylization only by per-image appearance, we evaluate the outputs by running DROID-SLAM and reporting trajectory and reconstruction consistency (ATE/RTE and point-cloud distances). This aligns the objective of stylization with the intended downstream use: producing view sets that remain trackable and geometrically coherent.
\end{itemize}

In practice, this yields strong improvements in 3D consistency metrics in our experiments, while introducing a controllable trade-off in pure color/style alignment.

\chapter{Methodology}

\section{Overview}
\label{sec:method_overview}

\begin{figure}[t]
    \centering
    \includegraphics[width=\linewidth]{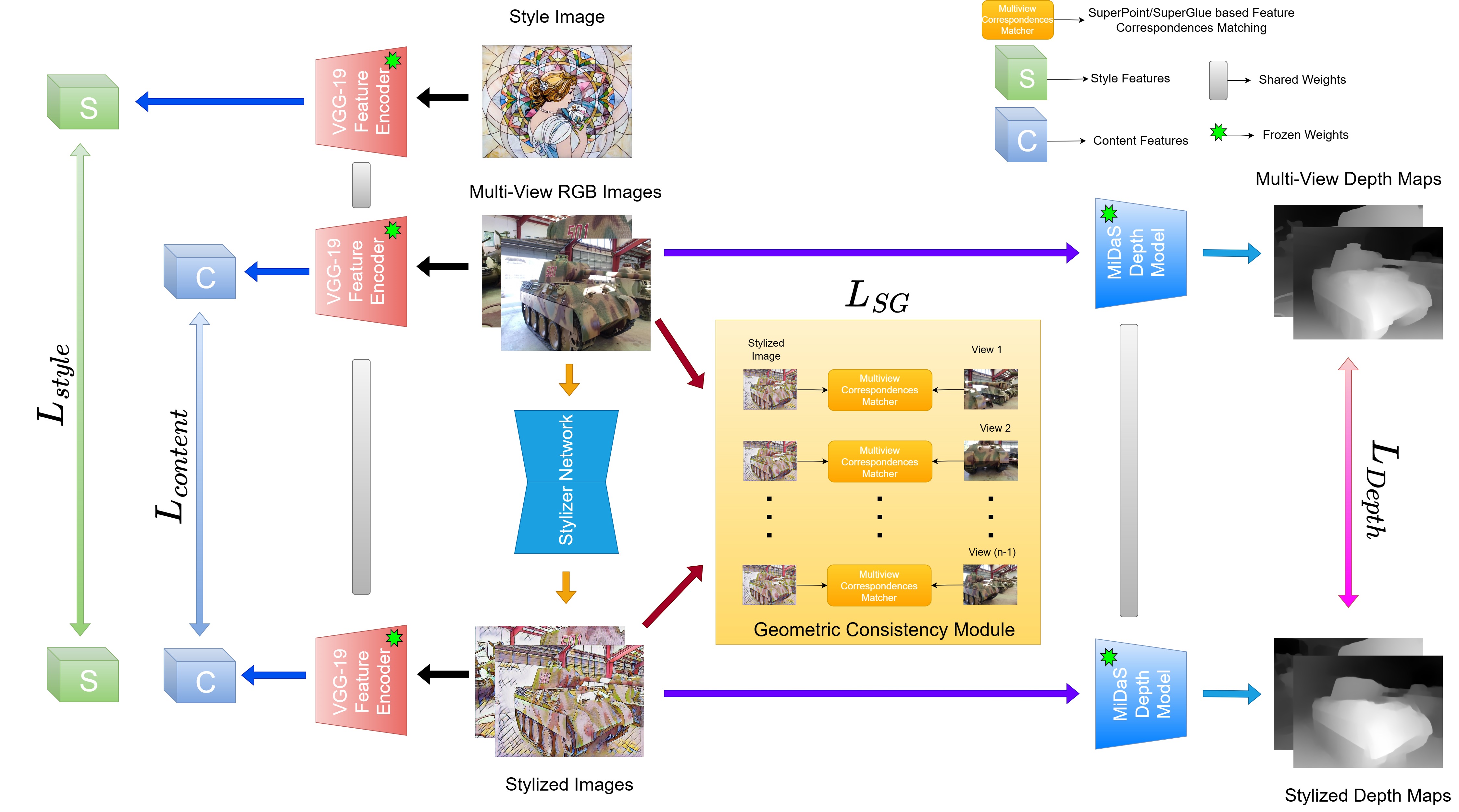}
    \caption{Proposed pose-free multi-view stylization pipeline. The stylizer network produces stylized views, while training is guided by VGG-based content/style losses, a SuperPoint/SuperGlue correspondence-based geometry loss $\mathcal{L}_{SG}$, and a MiDaS/DPT depth-preservation loss $\mathcal{L}_{Depth}$.}
    \label{fig:mv_pipeline}
\end{figure}

Given a set of multi-view images of a static scene $\{I_i\}_{i=1}^{N}$ and a reference style image $S$, our goal is to learn a feed-forward stylization network $g_{\theta}$ that produces stylized views
\begin{equation}
\hat{I}_i = g_{\theta}(I_i,S), \quad i \in \{1,\dots,N\},
\end{equation}
such that each $\hat{I}_i$ reflects the style appearance of $S$ while preserving the underlying 3D scene geometry sufficiently for cross-view correspondence, depth stability, and downstream 3D reconstruction.

Figure~\ref{fig:mv_pipeline} illustrates the training workflow. The inputs to the system are (i) the multi-view RGB images $\{I_i\}$ and (ii) a single style reference image $S$. The style image is first processed by a frozen VGG-19 feature encoder to extract multi-layer style statistics (channel-wise mean and standard deviation). These statistics serve as fixed style targets during training. In parallel, each training iteration samples a batch of multi-view images and forwards them through the stylizer $g_{\theta}$ to obtain stylized outputs $\{\hat{I}_i\}$. The original batch images and the corresponding stylized outputs are also passed through the same frozen VGG-19 encoder to compute content features (for structure preservation) and stylized feature statistics (for style matching). This yields the perceptual objectives $\mathcal{L}_{content}$ and $\mathcal{L}_{style}$ shown on the left side of Figure~\ref{fig:mv_pipeline}.

To explicitly enforce multi-view geometric consistency, the stylized outputs are additionally guided by a correspondence-based module built on SuperPoint and SuperGlue. As shown in the center of Figure~\ref{fig:mv_pipeline}, we select a stylized \emph{anchor} view and compute SuperPoint keypoints and descriptors on the stylized anchor image. We then establish correspondences between this anchor and all other views of the same scene using SuperGlue matching. The resulting correspondences define a set of cross-view feature pairs, and the SuperGlue loss $\mathcal{L}_{SG}$ penalizes deviations between matched descriptors (and optionally discourages the suppression of salient keypoints). Intuitively, this term prevents the stylizer from introducing view-dependent distortions that would break repeatable correspondences, e.g., bending edges, shifting textures across surfaces, or removing thin structures and thereby improving viewpoint stability and reducing geometric flicker.

In addition to correspondence stability, we incorporate depth preservation using a frozen MiDaS/DPT depth estimator. As shown on the right side of Figure~\ref{fig:mv_pipeline}, the original images produce reference depth maps, while the stylized images produce stylized depth maps. To make depth supervision more compatible with the stylized domain, the original RGB images are optionally normalized toward the style color statistics prior to depth inference (a practical domain-alignment step). The depth loss $\mathcal{L}_{Depth}$ then penalizes discrepancies between reference and stylized depth predictions (after normalization to handle scale ambiguity), discouraging stylization-induced changes that would corrupt monocular depth cues and destabilize downstream 3D fusion.

Finally, the total training objective is formed as a weighted sum of the perceptual stylization losses, the correspondence-based geometry loss, the depth-preservation loss, and standard regularizers (e.g., total variation and optional RGB reconstruction):
\begin{equation}
\mathcal{L} \;=\;
\lambda_c \mathcal{L}_{content} +
\lambda_s \mathcal{L}_{style} +
\lambda_{sg}\mathcal{L}_{SG} +
\lambda_d \mathcal{L}_{Depth}
\end{equation}
This loss is backpropagated through $g_{\theta}$ (while VGG-19, SuperPoint/SuperGlue, and MiDaS/DPT remain frozen), and the stylizer parameters $\theta$ are optimized using Adam. In our experiments, we instantiate $g_{\theta}$ either as a series of Residual Blocks CNN or as a U-Net backbone; both operate per-image at inference time, enabling fast stylization while the training-time constraints enforce multi-view geometric integrity.

\section{Stylizer Network}
\label{sec:stylizer_network}

The stylizer $g_{\theta}$ is a feed-forward CNN that maps an input view $I_i \in [0,1]^{3\times H\times W}$ to a stylized output $\hat{I}_i \in [0,1]^{3\times H\times W}$. In this thesis we experiment with two architectural variants for $g_{\theta}$: (i) a residual CNN\cite{he2016deep} and (ii) a U-Net encoder--decoder \cite{ronneberger2015u}. Both variants are trained per-scene using the same loss framework (AdaIN-style perceptual losses, SuperPoint/SuperGlue geometrical consistency, and MiDaS/DPT depth preservation). The motivation for evaluating both is twofold: the residual CNN provides a simple, series-connectivity baseline that is fast to train, while the U-Net introduces multi-scale features and skip connections that can better capture style as well as preserve edges and high-frequency structures, which are crucial for stable correspondences across views.

\subsection{Residual CNN stylizer}
Our first stylizer variant is a compact residual network composed of an input convolution, $B$ residual blocks, and an output convolution. Each residual block maintains the same spatial resolution and channel dimension, making the network lightweight and well-suited for per-scene optimization. The forward pass can be written as
\begin{align}
h_0 &= \phi(\mathrm{Conv}_{3\times 3}^{(in)}(I)),\\
h_{k+1} &= \phi\!\left(h_k + \mathrm{Conv}_{3\times 3}^{(2)}\!\left(\phi(\mathrm{Conv}_{3\times 3}^{(1)}(h_k))\right)\right), \quad k = 0,\dots,B-1,\\
\hat{I} &= \mathrm{clip}_{[0,1]}\!\left(\mathrm{Conv}_{3\times 3}^{(out)}(h_B)\right),
\end{align}
where $\phi$ is ReLU and $B{=}5$ in our experiments. Notably, this model does \emph{not} use batch normalization. In our setting, training is performed with small batch sizes and on a single scene; batch normalization would make the mapping depend on batch statistics, which can introduce undesirable appearance fluctuations and instability when batch composition changes. Instead, the residual CNN relies on its shallow depth and the perceptual supervision to learn a stable stylization mapping.

\paragraph{Design rationale.}
The residual CNN encourages a near-identity mapping early in training (via skip connections), which helps preserve coarse geometry. This is important because our geometry objective relies on maintaining matchable local structures. Moreover, the low model capacity acts as an implicit regularizer, rather than hallucinating new structures, the network tends to transfer style primarily through texture and color statistics, aligning well with the goals of view-consistent stylization.

\subsection{U-Net stylizer}
To better preserve fine details and reduce view-dependent flicker, we also employ a U-Net\cite{ronneberger2015u} architecture. U-Nets use an encoder--decoder structure with skip connections between corresponding resolutions, allowing high-frequency information (edges, thin structures) to bypass the bottleneck. This property is particularly beneficial in our framework because stable edges and local gradients support repeatable keypoints and more consistent correspondence matching.

Our UNet consists of three downsampling stages, a bottleneck block, and three upsampling stages. Let $\mathrm{Pool}(\cdot)$ denote $2\times2$ max pooling and $\mathrm{Up}(\cdot)$ denote bilinear upsampling by a factor of two. Each stage uses a convolutional block:
\begin{equation}
\mathrm{Block}(x) = \phi\!\left(\mathrm{IN}(\mathrm{Conv}( \phi(\mathrm{IN}(\mathrm{Conv}(x))) ))\right),
\end{equation}
where $\mathrm{IN}$ is instance normalization and $\phi$ is ReLU. Instance normalization has been shown to be particularly effective for stylization as it normalizes feature statistics per image and reduces sensitivity to global contrast/illumination variations \cite{ulyanov2016instance}. In our implementation, we use affine instance normalization (learnable scale and bias) inside each block.

The network produces intermediate feature maps
\begin{align}
d_1 &= \mathrm{Block}_{1}(I), \\
d_2 &= \mathrm{Block}_{2}(\mathrm{Pool}(d_1)), \\
d_3 &= \mathrm{Block}_{3}(\mathrm{Pool}(d_2)), \\
m   &= \mathrm{Block}_{mid}(\mathrm{Pool}(d_3)),
\end{align}
followed by the decoder with skip concatenations:
\begin{align}
u_3 &= \mathrm{Block}_{up3}\!\left([\mathrm{Up}(m), d_3]\right), \\
u_2 &= \mathrm{Block}_{up2}\!\left([\mathrm{Up}(u_3), d_2]\right), \\
u_1 &= \mathrm{Block}_{up1}\!\left([\mathrm{Up}(u_2), d_1]\right),
\end{align}
where $[\cdot,\cdot]$ denotes channel-wise concatenation. The final output is obtained by a $3\times3$ convolution and a sigmoid activation:
\begin{equation}
\hat{I} = \sigma\!\left(\mathrm{Conv}_{3\times 3}^{(out)}(u_1)\right),
\end{equation}
ensuring $\hat{I}\in[0,1]$.

\paragraph{Normalization choice: no BatchNorm, InstanceNorm in U-Net.}
We avoid batch normalization in both stylizers because our training regime uses small batches and per-scene optimization; batch-dependent statistics can cause inconsistent output appearance and unstable training dynamics. In contrast, the U-Net variant uses instance normalization inside each convolutional block. Instance normalization operates per image (not across the batch), making it robust to batch size and well aligned with stylization objectives, as shown in prior work on fast style transfer \cite{ulyanov2016instance}. Empirically, we found that this improves training stability and supports stronger style transfer without sacrificing the local structures required for geometric consistency.


\section{Stylization}
\label{sec:stylization}

Our stylization objective follows the standard perceptual-loss formulation introduced in neural style transfer, where a fixed, pretrained classification network provides feature spaces in which \emph{content} and \emph{style} can be compared \cite{gatys2016image,johnson2016perceptual}. Concretely, we use a frozen VGG-19 feature encoder \cite{simonyan2014very} and supervise the stylizer outputs using (i) a content reconstruction loss at a deeper layer and (ii) an AdaIN-style statistics loss that matches feature-channel moments to a reference style image \cite{huang2017arbitrary}.

\subsection{Feature encoder and preprocessing}
Let $\Phi_{\ell}(\cdot)$ denote the activation map extracted from layer $\ell$ of VGG-19.
All images are first normalized using the ImageNet channel statistics:
\begin{equation}
\tilde{I} \;=\; \frac{I - \mu}{\sigma},
\end{equation}
where $\mu=(0.485,0.456,0.406)$ and $\sigma=(0.229,0.224,0.225)$ are applied per-channel.
We keep VGG-19 fixed during training and only optimize the stylizer parameters $\theta$.

In our implementation, we extract features at the layers
\begin{equation}
\mathcal{S}_{\ell}=\{\texttt{relu1\_1},\texttt{relu2\_1},\texttt{relu3\_1},\texttt{relu4\_1}\},
\quad\text{and}\quad
\ell_c=\texttt{relu4\_2},
\end{equation}
where $\ell_c$ is used for content supervision and $\mathcal{S}_{\ell}$ is used for style statistics matching.

\subsection{Content loss}
Following the perceptual content constraint used in neural style transfer \cite{gatys2016image,johnson2016perceptual}, we preserve scene structure by matching deep VGG features between the input view $I$ and its stylized output $\hat{I}$. Using the chosen content layer $\ell_c=\texttt{relu4\_2}$, we define
\begin{equation}
\mathcal{L}_{content}(I,\hat{I})
\;=\;
\left\|\Phi_{\ell_c}(\tilde{\hat{I}}) - \Phi_{\ell_c}(\tilde{I})\right\|_2^2,
\end{equation}
which encourages the stylized output to retain the semantic layout and coarse geometry of the original image while allowing low-level appearance to change.

\subsection{AdaIN-style statistics loss}
\label{sec:adain_style_loss}

For style transfer, we adopt an AdaIN\cite{huang2017arbitrary}-inspired objective that matches feature-channel statistics of the stylized output to those of the (normalized) style reference image $\tilde{S}$. Compared to Gram-matrix style losses \cite{gatys2016image}, moment matching is computationally lightweight and aligns with the view that artistic style is largely captured by feature distribution statistics.

Let $F \in \mathbb{R}^{C \times H \times W}$ be a feature map. We compute channel-wise mean and standard deviation as
\begin{align}
\mu(F)_c &= \frac{1}{HW}\sum_{u=1}^{H}\sum_{v=1}^{W} F_{cuv},\\
\sigma(F)_c &= \sqrt{\frac{1}{HW}\sum_{u=1}^{H}\sum_{v=1}^{W}\big(F_{cuv}-\mu(F)_c\big)^2 + \epsilon},
\end{align}
with $\epsilon=10^{-5}$ for numerical stability.

We define the set of VGG layers used for style statistics matching as
\begin{equation}
\mathcal{S}_{\ell}=\{\texttt{relu1\_1},\texttt{relu2\_1},\texttt{relu3\_1},\texttt{relu4\_1}\}.
\end{equation}
For each style layer $\ell \in \mathcal{S}_{\ell}$, we precompute the target style statistics from the style image:
\begin{equation}
\mu_{S}^{\ell}=\mu\!\left(\Phi_{\ell}(\tilde{S})\right),
\qquad
\sigma_{S}^{\ell}=\sigma\!\left(\Phi_{\ell}(\tilde{S})\right).
\end{equation}
The AdaIN-style statistics loss for a stylized output $\hat{I}$ is then
\begin{equation}
\mathcal{L}_{style}(S,\hat{I})
=
\sum_{\ell \in \mathcal{S}_{\ell}}
\left(
\left\|\mu\!\left(\Phi_{\ell}(\tilde{\hat{I}})\right)-\mu_{S}^{\ell}\right\|_{2}^{2}
+
\left\|\sigma\!\left(\Phi_{\ell}(\tilde{\hat{I}})\right)-\sigma_{S}^{\ell}\right\|_{2}^{2}
\right).
\end{equation}
By minimizing $\mathcal{L}_{style}$, the stylizer learns to reproduce the style image’s texture statistics across multiple VGG feature scales (from low-level appearance to more global patterns), while $\mathcal{L}_{content}$ ensures that the underlying scene structure remains aligned to the original multi-view inputs.

\paragraph{Single-style and multi-style usage.}
In the single-style setting, $(\mu_{S}^{\ell},\sigma_{S}^{\ell})$ are computed once from the chosen style image $S$ and reused throughout training. In the multi-style extension, we compute and store statistics for each style image and select the corresponding targets based on the active style identifier during training.

\section{Depth Loss}
\label{sec:depth_loss}

While perceptual style losses can produce visually pleasing results, they do not explicitly constrain cues that are important for 3D reasoning. In practice, stylization often modifies local contrast, shading, and mid-frequency gradients in a way that can confuse monocular depth priors, producing depth estimates that vary across views even for the same underlying geometry. Such depth inconsistencies can propagate to downstream tasks such as multi-view fusion, mapping, and pose estimation. To mitigate this effect, we introduce a depth-preservation loss that regularizes the stylizer outputs to maintain a depth structure consistent with the original views under a frozen teacher depth model.

\subsection{Depth-domain alignment via global color transfer}
Monocular depth networks are sensitive to appearance and domain shift. If the stylized outputs have a very different color distribution than the original inputs, the depth predictions may change due to photometric bias rather than true geometric differences. To reduce this effect and obtain more stable reference targets, we apply a simple global color transfer that aligns the dataset’s RGB statistics to those of the style image, inspired by classic color-transfer normalization \cite{reinhard2002color}.

Let $(\mu_c, \sigma_c)$ be the per-channel mean and standard deviation computed over the \emph{multi-view dataset} $\{I_i\}$, and let $(\mu_s,\sigma_s)$ be the per-channel statistics computed from the style image $S$. For each input view $I_i$, we compute a color-aligned image
\begin{equation}
I_i^{ct} =
\mathrm{clip}_{[0,1]}\!\left(
 (I_i - \mu_c)\odot \frac{\sigma_s}{\sigma_c} + \mu_s
\right),
\label{eq:color_transfer}
\end{equation}
where $\odot$ denotes channel-wise multiplication. This step does not aim to stylize the image; rather, it shifts the \emph{global} color distribution toward the style domain so that depth supervision is less affected by color/contrast differences.

\subsection{Depth preservation with a frozen MiDaS/DPT teacher}
Let $D(\cdot)$ denote a frozen monocular depth estimator (MiDaS/DPT) \cite{ranftl2020towards, ranftl2021vision}. We precompute a reference depth map for each view using the color-aligned image:
\begin{equation}
D_i^{ref} = D(I_i^{ct}).
\label{eq:depth_ref}
\end{equation}
During training, we predict depth from the stylized output:
\begin{equation}
D_i^{sty} = D(\hat{I}_i).
\label{eq:depth_sty}
\end{equation}

Since monocular depth is ambiguous up to an affine transformation (scale and shift), we compare depth maps only after per-image normalization:
\begin{equation}
\mathcal{N}(D) = \frac{D - \mathbb{E}[D]}{\sqrt{\mathrm{Var}(D)} + \epsilon},
\label{eq:depth_norm}
\end{equation}
where $\mathbb{E}[\cdot]$ and $\mathrm{Var}(\cdot)$ are computed over all pixels of a depth map and $\epsilon$ is a small constant for numerical stability.

We define the depth loss over a training batch $\mathcal{B}$ using a robust Smooth L1 penalty:
\begin{equation}
\mathcal{L}_{Depth} \;=\;
\frac{1}{|\mathcal{B}|}
\sum_{i \in \mathcal{B}}
\mathrm{SmoothL1}\!\left(
\mathcal{N}(D_i^{sty}),
\mathcal{N}(D_i^{ref})
\right).
\label{eq:depth_loss}
\end{equation}

\paragraph{Why depth helps multi-view stylization.}
The depth loss complements the correspondence-based geometry loss in two ways. First, it constrains \emph{global} structure: even if local keypoint correspondences remain stable, stylization may still introduce low-frequency distortions or contrast changes that cause depth priors to predict inconsistent scene relief. Depth regularization discourages such deviations and stabilizes volumetric cues. Second, it acts as a balancing force against over-stylization: strong style supervision may encourage texture patterns that overwhelm shading gradients and structural boundaries; enforcing depth consistency encourages the stylizer to preserve cues correlated with 3D layout. Empirically, we observe that adding $\mathcal{L}_{Depth}$ reduces view-dependent changes in large planar regions and improves the stability of reconstructed geometry in downstream SLAM-based evaluation (Chapter~\ref{ch:quant_eval}).

\section{Multi-View Correspondence Consistency via SuperPoint/SuperGlue}
\label{sec:sg_loss}

A key challenge in multi-view stylization is that perceptual style objectives do not enforce that the same 3D surface point retains a stable appearance across viewpoints. As a result, per-view stylization can introduce view-dependent distortions: edges bend, textures drift across surfaces, and thin structures appear or disappear, which in turn degrades feature repeatability and harms downstream 3D tasks. To explicitly regularize cross-view geometric stability \emph{without requiring camera poses}, we introduce a correspondence-based loss built on SuperPoint keypoints and SuperGlue matching \cite{detone2018superpoint,sarlin2020superglue}. The intuition is simple: if stylization preserves the local structures that modern feature pipelines rely on, then keypoints should remain matchable across views, and their descriptors should remain consistent.

\subsection{Precomputing SuperPoint features on the original views}
Given the multi-view dataset $\{I_i\}_{i=1}^{N}$, we first convert each image to grayscale and extract SuperPoint keypoints, scores, and descriptors. For each view $I_j$, we cache:
\begin{equation}
\mathcal{C}_j = \left\{\mathbf{K}_j,\ \mathbf{s}_j,\ \mathbf{F}_j\right\},
\end{equation}
where $\mathbf{K}_j\in\mathbb{R}^{N_j\times 2}$ are keypoint coordinates, $\mathbf{s}_j\in\mathbb{R}^{N_j}$ are keypoint confidence scores, and $\mathbf{F}_j\in\mathbb{R}^{D\times N_j}$ are descriptors (with descriptor dimension $D=256$ in SuperPoint). This cache is computed once before training and remains fixed.

\subsection{Anchor-to-all matching during training}
At each training iteration, we randomly select an \emph{anchor} index $a$ and compute a stylized anchor image $\hat{I}_a=g_{\theta}(I_a)$. We then extract SuperPoint features on the stylized anchor (this branch is differentiable with respect to $\hat{I}_a$, hence gradients flow back to $g_{\theta}$). For every other view $j\neq a$, we run SuperGlue to match the anchor features to the cached features of $I_j$:
\begin{equation}
m_{aj}(\cdot),\ \alpha_{aj}(\cdot) \;=\; 
\mathrm{SuperGlue}\!\left(\hat{I}_a,\ \mathcal{C}_j\right),
\end{equation}
where $m_{aj}(k)$ denotes the matched index in view $j$ for anchor keypoint $k$ (or $-1$ if unmatched), and $\alpha_{aj}(k)$ is the match confidence score. We retain only confident matches:
\begin{equation}
\mathcal{V}_{aj}
=
\Bigl\{\, k \in \{1,\dots,N_a\}\; \Big|\; m_{aj}(k)\neq -1 \ \wedge\  \alpha_{aj}(k)\ge \tau \Bigr\},
\label{eq:valid_matches}
\end{equation}
with confidence threshold $\tau$ (we use $\tau=0.2$ in our experiments).

\subsection{Per-pair descriptor consistency loss}
For each accepted match $k\in\mathcal{V}_{aj}$, we compare L2-normalized descriptors between the stylized anchor and the original target view:
\begin{align}
\widehat{\mathbf{f}}_{a,k} &= \frac{\mathbf{f}_{a,k}}{\|\mathbf{f}_{a,k}\|_2}, \qquad
\widehat{\mathbf{f}}_{j,m_{aj}(k)} = \frac{\mathbf{f}_{j,m_{aj}(k)}}{\|\mathbf{f}_{j,m_{aj}(k)}\|_2},\\
d_{aj}(k) &= 1 - \left\langle \widehat{\mathbf{f}}_{a,k},\ \widehat{\mathbf{f}}_{j,m_{aj}(k)} \right\rangle,
\end{align}
where $d_{aj}(k)$ is the cosine distance (in $[0,2]$). We apply a robust Huber penalty to reduce sensitivity to mismatches and outliers:
\begin{equation}
\rho_{\delta}(x)
=
\begin{cases}
\frac{1}{2}x^{2}, & |x|\le \delta,\\[4pt]
\delta\left(|x|-\frac{1}{2}\delta\right), & |x|>\delta,
\end{cases}
\label{eq:huber}
\end{equation}
with $\delta=0.5$ in our experiments. The descriptor-consistency loss for the pair $(a,j)$ is:
\begin{equation}
\ell^{\mathrm{desc}}_{aj}
=
\frac{1}{|\mathcal{V}_{aj}|}
\sum_{k\in \mathcal{V}_{aj}}
\rho_{\delta}\!\left(d_{aj}(k)\right),
\qquad
\text{and}\quad
|\mathcal{V}_{aj}|=0 \Rightarrow \ell^{\mathrm{desc}}_{aj}=0.
\label{eq:desc_loss_pair}
\end{equation}

\subsection{Optional saliency regularizer}
We optionally encourage the stylized anchor to preserve salient keypoints by penalizing low SuperPoint confidence scores for the matched anchor keypoints:
\begin{equation}
\ell^{\mathrm{sal}}_{aj}
=
\frac{1}{|\mathcal{V}_{aj}|}
\sum_{k\in \mathcal{V}_{aj}}
\left(1 - s_{a,k}\right)^{2},
\qquad
|\mathcal{V}_{aj}|=0 \Rightarrow \ell^{\mathrm{sal}}_{aj}=0,
\label{eq:sal_loss_pair}
\end{equation}
where $s_{a,k}$ is the SuperPoint score at the anchor keypoint $k$. In our experiments, the saliency weight is $\lambda_{\mathrm{sal}}=0.1$.

\subsection{Per-image anchor loss and total dataset geometry loss}
For a fixed anchor $a$, we define the \emph{per-anchor} SuperGlue loss (this corresponds to the per-image basis loss requested) as the sum over all other views:
\begin{equation}
\ell_{SG}(a)
=
\sum_{\substack{j=1\\ j\neq a}}^{N}
\left(
\ell^{\mathrm{desc}}_{aj}
+
\lambda_{\mathrm{sal}}\,\ell^{\mathrm{sal}}_{aj}
\right).
\label{eq:per_anchor_sg}
\end{equation}
This term measures how well the stylized anchor remains matchable, in descriptor space, to the rest of the multi-view set.

Finally, the total multi-view correspondence consistency objective over the full dataset is:
\begin{equation}
\mathcal{L}_{SG}
=
\sum_{a=1}^{N}\ell_{SG}(a).
\label{eq:total_sg}
\end{equation}
In practice, evaluating Eq.~\eqref{eq:total_sg} exactly would be expensive because it requires stylizing every view as an anchor in every iteration. Instead, we approximate it stochastically: at each iteration we sample a single anchor index $a$ uniformly at random and compute $\ell_{SG}(a)$ (anchor-to-all) as shown in Figure~\ref{fig:mv_pipeline}. Over training, this Monte-Carlo estimate provides unbiased coverage of anchors while keeping computation tractable.

\paragraph{Why this loss improves geometry without camera poses.}
The proposed correspondence loss does not require known camera intrinsics or extrinsics. It leverages the fact that multi-view geometry manifests as \emph{repeatable local structures} across views: corners, junctions, and textured regions should remain detectable and matchable. Stylization often disrupts these cues by shifting edges or introducing viewpoint-dependent texture patterns. By penalizing descriptor drift on confident SuperGlue matches, $\mathcal{L}_{SG}$ encourages the stylizer to produce outputs that preserve the local structures that feature-based reconstruction pipelines depend on, thereby reducing geometric flicker and improving downstream SLAM/MVS robustness.

\section{Weight Scheduling (Warmup + Ramp)}
\label{sec:weight_scheduling}

The correspondence-based geometry loss $\mathcal{L}_{SG}$ and the depth loss $\mathcal{L}_{Depth}$ are strong regularizers. If they are applied with full strength from the start, they can dominate the optimization when the stylizer output is still unstable, often driving the network toward a conservative (near-identity) solution that preserves structure but fails to learn the target style. Conversely, training only with perceptual stylization losses can yield visually stylized outputs that deform edges and reduce cross-view matchability. To obtain a stable trade-off, we introduce the geometry and depth constraints gradually using a warmup-and-ramp schedule.

\paragraph{Schedule definition.}
For training step $t$, we use a piecewise-linear weight schedule
\begin{equation}
w(t; t_0, t_r, w_{\max}) \;=\;
\begin{cases}
0, & t \le t_0,\\[4pt]
w_{\max}\cdot \dfrac{t-t_0}{t_r}, & t_0 < t \le t_0+t_r,\\[8pt]
w_{\max}, & t > t_0+t_r,
\end{cases}
\label{eq:weight_schedule}
\end{equation}
and apply it independently to the SuperGlue geometry term and the depth term:
\begin{equation}
\lambda_{sg}(t) = w(t; t_0^{sg}, t_r^{sg}, \lambda_{sg}^{\max}),
\qquad
\lambda_{d}(t)  = w(t; t_0^{d},  t_r^{d},  \lambda_{d}^{\max}).
\end{equation}

\paragraph{Values used in our experiments.}
Unless stated otherwise, we use the following settings throughout the experiments:
\begin{itemize}
    \item \textbf{SuperGlue scheduling:} warmup $t_0^{sg}=200$ iterations, ramp $t_r^{sg}=400$ iterations, final weight $\lambda_{sg}^{\max}=1.0$.
    \item \textbf{Depth scheduling:} warmup $t_0^{d}=200$ iterations, ramp $t_r^{d}=400$ iterations, final weight $\lambda_{d}^{\max}=0.1$.
\end{itemize}
Thus, both constraints are inactive for the first 200 iterations, increase linearly from iterations 201--600, and remain fixed at their final values from iteration 601 onward (for the remainder of training, typically up to 1000 iterations).

\paragraph{Why warmup + ramp is important.}
This scheduling is particularly important in our setting for three practical reasons:
\begin{itemize}
    \item \textbf{Match stability improves after stylization forms.} In early training, stylized outputs may not contain stable corner/edge structures, leading to few confident SuperGlue matches and noisy correspondence supervision. Delaying $\mathcal{L}_{SG}$ until the stylization mapping stabilizes yields more reliable matches and gradients.
    \item \textbf{Prevents ``geometry-only'' collapse.} Descriptor consistency and depth consistency both encourage geometrical conservation. If they are strongly enforced too early, the stylizer can minimize them by remaining close to the input appearance, resulting in weak stylization. Warmup ensures the network first learns the appearance shift required by the target style.
    \item \textbf{Smooth optimization under discrete matching.} The geometry term depends on confidence thresholding and match filtering, which can change abruptly between iterations. A gradual ramp reduces sudden gradient shocks and improves convergence.
\end{itemize}

Overall, the warmup-and-ramp strategy allows the model to first learn the style distribution under $\mathcal{L}_{content}$ and $\mathcal{L}_{style}$, and then progressively adapt this stylization into a multi-view consistent form under $\mathcal{L}_{SG}$ and $\mathcal{L}_{Depth}$.

\section{Final Objective}
\label{sec:final_objective}

In the single-style setting, we train the stylizer $g_{\theta}$ with a weighted combination of the stylization losses (content and AdaIN-style statistics) and the geometry-preservation losses (SuperPoint/SuperGlue correspondence consistency and MiDaS/DPT depth consistency). At iteration $t$, the total objective is:
\begin{equation}
\mathcal{L}(t) \;=\;
\lambda_c\,\mathcal{L}_{content}
\;+\;
\lambda_s\,\mathcal{L}_{style}
\;+\;
\lambda_{sg}(t)\,\ell_{SG}(a)
\;+\;
\lambda_d(t)\,\mathcal{L}_{Depth},
\label{eq:final_objective_single}
\end{equation}
where $\ell_{SG}(a)$ is the anchor-to-all correspondence loss for a randomly sampled anchor view $a$ (Section~\ref{sec:sg_loss}). The weights $\lambda_{sg}(t)$ and $\lambda_d(t)$ follow the warmup-and-ramp schedules described in Section~\ref{sec:weight_scheduling}; in our experiments we use a warmup of 200 iterations and a linear ramp of 400 iterations, with final weights $\lambda_{sg}^{\max}=1.0$ and $\lambda_d^{\max}=0.1$. All losses are differentiable with respect to the stylizer parameters $\theta$, while the VGG encoder, SuperPoint/SuperGlue models, and MiDaS/DPT depth network remain frozen during training. We optimize $\theta$ using Adam.

\section{Multi-style Format}
\label{sec:multistyle}

The single-style formulation trains one stylizer per \emph{(scene, style)} pair. While effective, this becomes expensive when multiple artistic styles are desired: each additional style requires a separate training run and separate network weights. To scale the framework to multiple styles while retaining the same geometry-preservation losses, we introduce a \textbf{style-conditioned} stylizer that can synthesize multiple styles with a single set of parameters.

\subsection{Style-conditioned stylizer with Conditional Instance Normalization}
Let $\{S^{(m)}\}_{m=1}^{M}$ be a set of $M$ reference style images, each assigned a discrete style id $m\in\{1,\dots,M\}$. We train a single network conditioned on the style id:
\begin{equation}
\hat{I}_i \;=\; g_{\theta}(I_i, m).
\end{equation}
We instantiate $g_{\theta}$ as a U-Net backbone (Section~\ref{sec:stylizer_network}) where the convolutional blocks are modulated using \textbf{Conditional Instance Normalization (CIN)} \cite{dumoulin2016learned}. Concretely, for an activation tensor $x \in \mathbb{R}^{B\times C\times H\times W}$, instance normalization is
\begin{equation}
\mathrm{IN}(x) \;=\; \frac{x - \mu(x)}{\sigma(x)+\epsilon},
\end{equation}
where $\mu(\cdot)$ and $\sigma(\cdot)$ are computed per instance and per channel over spatial dimensions. CIN applies an additional \emph{style-dependent} affine transform:
\begin{equation}
\mathrm{CIN}(x; m) \;=\; \gamma^{(m)} \odot \mathrm{IN}(x) \;+\; \beta^{(m)},
\label{eq:cin}
\end{equation}
with $\gamma^{(m)},\beta^{(m)}\in\mathbb{R}^{C}$ broadcast over spatial dimensions.

In our implementation, the style id is mapped to a learnable embedding vector $e_m \in \mathbb{R}^{d}$ (with $d=128$). Each CIN layer predicts its affine parameters from this embedding:
\begin{equation}
\gamma^{(m)} = W_{\gamma} e_m + b_{\gamma}, 
\qquad
\beta^{(m)} = W_{\beta} e_m + b_{\beta},
\end{equation}
where $(W_{\gamma},b_{\gamma},W_{\beta},b_{\beta})$ are learned per CIN layer. This design allows the network to share \emph{content-processing} capacity across styles while using lightweight, style-specific modulation to control the rendered appearance.

\paragraph{Why CIN is a good fit for multi-style stylization.}
CIN is particularly effective for multi-style transfer because (i) instance normalization removes global contrast/illumination variations on a per-image basis, which aligns well with stylization objectives, and (ii) the style-dependent affine parameters re-inject style information as a controllable modulation signal \cite{dumoulin2016learned}. Practically, CIN enables training a single model for multiple styles without duplicating the full network parameters, reducing memory and training overhead compared to training one model per style.

\subsection{Multi-style AdaIN statistics loss}
The content loss remains unchanged from Section~\ref{sec:stylization} (computed between $I_i$ and $\hat{I}_i$ at $\ell_c=\texttt{relu4\_2}$). The style loss is extended by maintaining a separate set of VGG style statistics for each style image. For each style $S^{(m)}$, we precompute
\begin{equation}
\left(\mu_{S^{(m)}}^{\ell}, \sigma_{S^{(m)}}^{\ell}\right), 
\qquad \forall \ell \in \mathcal{S}=\{\texttt{relu1\_1},\texttt{relu2\_1},\texttt{relu3\_1},\texttt{relu4\_1}\}.
\end{equation}
Given the active style id $m$, the AdaIN-style statistics loss becomes
\begin{equation}
\mathcal{L}_{style}^{(m)} \;=\;
\sum_{\ell \in \mathcal{S}}
\left(
\left\|\mu(\Phi_{\ell}(\tilde{\hat{I}})) - \mu_{S^{(m)}}^{\ell}\right\|_2^2
+
\left\|\sigma(\Phi_{\ell}(\tilde{\hat{I}})) - \sigma_{S^{(m)}}^{\ell}\right\|_2^2
\right).
\label{eq:multistyle_styleloss}
\end{equation}

\subsection{Geometry and depth losses in the multi-style setting}
The geometry-preservation components remain structurally identical to the single-style framework. For the correspondence loss, the anchor view is stylized using the same style id $m$, i.e., $\hat{I}_a=g_{\theta}(I_a,m)$, and the anchor-to-all SuperPoint/SuperGlue loss $\ell_{SG}(a)$ is computed as in Section~\ref{sec:sg_loss}. Similarly, the MiDaS/DPT depth consistency loss (Section~\ref{sec:depth_loss}) is applied between depths predicted from stylized outputs and reference depths computed from the original views. In the multi-style implementation we use a single reference depth per view (computed from the original RGB), which avoids an $O(NM)$ depth-cache memory cost and keeps the depth teacher consistent across styles.

\subsection{Overall multi-style training objective}
At each iteration, we sample a style id $m$ and a minibatch of views $\mathcal{B}$, and optimize:
\begin{equation}
\mathcal{L}(t; m) \;=\;
\lambda_c\,\mathcal{L}_{content}
\;+\;
\lambda_s\,\mathcal{L}_{style}^{(m)}
\;+\;
\lambda_{sg}(t)\,\ell_{SG}(a)
\;+\;
\lambda_d(t)\,\mathcal{L}_{Depth},
\label{eq:multistyle_total_loss}
\end{equation}
where $\lambda_{sg}(t)$ and $\lambda_d(t)$ follow the same warmup-and-ramp schedules as in the single-style case (Section~\ref{sec:weight_scheduling}). Importantly, $\ell_{SG}$ and $\mathcal{L}_{Depth}$ are \emph{style-agnostic} losses: they constrain cross-view geometrical stability and volumetric consistency regardless of the chosen artistic style. This allows a single conditioned model to learn multiple appearance mappings while still preserving the geometric integrity required for downstream 3D tasks.

\paragraph{Generalization to unseen scenes.}
Although training is performed per scene, conditioning the stylizer on a discrete style code encourages a separation between \emph{style control} (via CIN modulation) and \emph{content processing} (via shared convolutional features). Empirically, we observe that the trained conditioned network can be applied to unseen scenes at test time—using the same style ids—to produce stylized outputs that maintain better edge/high-frequency stability than per-view independent stylization, while retaining the same geometry regularization mechanisms.

\chapter{Experimental Setup}

\section{Datasets}
We evaluate on two categories of real multi-view captures used in 3D reconstruction and novel-view synthesis.

\paragraph{Tanks and Temples (MuVieCAST version).}
For the Tanks and Temples\cite{Knapitsch2017} dataset we use the same version as used by \textbf{MuVieCAST}\cite{ibrahimli2024muviecast} to enable a fair state-of-the-art comparison on identical scenes and view sets.
For direct comparison with MuVieCAST we use the scenes \textit{Horse}, \textit{Panther}, \textit{Lighthouse}, \textit{Francis}, and \textit{Train}.
For our ablation studies on Tanks and Temples we use the scenes \textit{Train} and \textit{Truck}.
To keep the comparison controlled, Tanks and Temples frames are resized to $640\times480$ for both our method and MuVieCAST.

\paragraph{Mip-NeRF\cite{barron2021mip,barron2022mip} 360 captures.}
For ablations beyond Tanks and Temples we additionally evaluate on scenes from the Mip-NeRF 360 dataset: \textit{Cycle} (Bicycle), \textit{Garden}, and \textit{Counter}.
We keep those original frames whose sizes for these sequences: \textit{Counter} ($389\times260$), \textit{Cycle} ($618\times411$), and \textit{Garden} ($648\times420$).

\section{Styles}
Each experiment is defined by a \textbf{(scene, style)} pair.
For comparisons against MuVieCAST, we use the same style reference images provided in the MuVieCAST repository\cite{ibrahimli2024muviecast}:
\textit{greatWave} for \textit{Horse}, \textit{mosaic} for \textit{Panther}, \textit{abstract} for \textit{Lighthouse}, \textit{abstract2} for \textit{Francis}, and \textit{starry} for \textit{Train}.

For ablation studies (\textit{Train}, \textit{Truck}, \textit{Garden}, \textit{Cycle}, \textit{Counter}), we use a combination of MuVieCAST styles (e.g., \textit{starry}) and additional styles from the \texttt{style-transfer-dataset} repository (by Victor Kitov) \cite{kitov_style_dataset_github}, referenced by style IDs:
\begin{itemize}
    \item \textbf{Train / Truck / Garden:} style\_19 and starry,
    \item \textbf{Cycle:} style\_19, style\_15, style\_41, and starry,
    \item \textbf{Counter:} style\_19, starry, and style\_2.
\end{itemize}



\section{Baselines}
\paragraph{State-of-the-art baseline.}
We compare against \textbf{MuVieCAST} using the authors’ official implementation and the default hyperparameters reported in their paper.

\paragraph{Ablations (ours).}
To isolate the contribution of each design choice in our framework, we evaluate a staged set of variants built on the same \textbf{Residual stylizer} backbone:
\begin{enumerate}
    \item \textbf{Ours (Base):} residual blocks based backbone stylization trained with AdaIN losses $\mathcal{L}_{content}$ and $\mathcal{L}_{style}$.
    \item \textbf{Ours (Base) + $\mathcal{L}_{Depth}$:} adds depth-preservation via the frozen MiDaS/DPT model.
    \item \textbf{Ours (Base) + $\mathcal{L}_{SG}$:} adds correspondence-based geometric consistency via SuperPoint/SuperGlue matching.
    \item \textbf{Ours (Base) + $\mathcal{L}_{Depth}$ + $\mathcal{L}_{SG}$:} combines both depth and correspondence constraints.
\end{enumerate}
Unless stated otherwise, all ablations share the same training protocol (per-scene optimization, learning rate, iteration budget, and weight scheduling) and differ only in the enabled loss terms.

\section{Implementation Details}
We train a residual CNN stylizer/ U-Net stylizer  \textbf{per (scene, style) pair} using Adam Optimization. Images are loaded as RGB and resized so that the maximum side length is at most $768$ pixels (unless otherwise specified for fixed-resolution experiments).
Unless stated otherwise, we use the following hyperparameters:
\begin{itemize}
    \item iterations: $1000$, batch size: $4$, learning rate: $5\times 10^{-4}$,
    \item seed: $123$ (Python and PyTorch; CUDA seeds set), cuDNN benchmark enabled,
    \item preview saving interval: every $50$ iterations.
\end{itemize}

\paragraph{Loss weights.}
We use $w_{\text{content}}=1.0$, $w_{\text{style}}=6.0$, $w_{\text{rgb}}=0.0$ (disabled by default), and $w_{\text{tv}}=10^{-5}$.

\paragraph{VGG features.}
We use a pretrained VGG-19 encoder for perceptual losses. The content layer is relu4\_2 and the style layers are relu1\_1, relu2\_1, relu3\_1, and relu4\_1 (details are provided in the methodology chapter).

\paragraph{SuperPoint/SuperGlue regularization.}
We use SuperPoint and SuperGlue to preserve matchability across views by penalizing descriptor disagreement on confident matches.
At initialization we cache SuperPoint keypoints/descriptors for all original frames (up to $4096$ keypoints per image).
During training we sample one random anchor frame, stylize it, extract SuperPoint features with gradients, and match the anchor against all cached frames (``anchor vs.\ all'').
Hyperparameters are: final weight $w_{\text{sg}}=1.0$, confidence threshold $0.2$, Huber $\delta=0.5$, optional saliency term enabled with weight $0.1$.
We use a schedule with warmup $200$ steps (weight $0$) and ramp $400$ steps (linear increase to the final weight).

\paragraph{Depth preservation (MiDaS/DPT).}
When enabled, we use MiDaS with \texttt{DPT\_Large} as the depth backbone.
For DPT models, depth is inferred at $384\times384$ internally and resized back to the input resolution.
We compute a SmoothL1 loss on normalized depth maps between stylized outputs and cached reference depths.
To reduce domain shift, reference depths are computed after applying a global color transfer that maps the content-set mean/std to the style-image mean/std.
Depth loss weight uses warmup $200$ and ramp $400$ steps up to a final weight $w_{\text{depth}}=0.1$.

\paragraph{Computational Details.}
We trained our overall framework using a single Nividia A-100 GPU of 40 GB memory allocations. Our Base model and the Base + $\mathcal{L}_{Depth}$ took around 30 minutes to complete 1000 iterations of training. For Base + $\mathcal{L}_{SG}$, and Base + $\mathcal{L}_{Depth}$ + $\mathcal{L}_{GS}$ took on an average ~180 minutes to complete 1000 iterations. The training time varies with the number of the images in the multi-view dataset. Scenes with more number of multi-view images took more time to complete training.   

\section{Camera Intrinsics for DROID-SLAM Evaluation}
DROID-SLAM-based metrics require camera intrinsics.
For Tanks and Temples\cite{Knapitsch2017}, the benchmark does not explicitly provide exact intrinsics to encourage individual optimization, but recommends a pinhole model that works well for their camera setups:
\begin{equation}
x_0=\frac{W}{2},\qquad y_0=\frac{H}{2},\qquad f_x=f_y=0.7\,W,
\end{equation}
where $W$ and $H$ are the frame width and height in pixels.
We use this rule to construct intrinsics for our SLAM evaluation runs, consistent with the benchmark guidance and our resized image resolutions.

\chapter{Evaluation Metrics}
\section{Static Metrics}
\subsection{Color Histogram Distance (CHD)}
We use the Color Histogram Distance (CHD)\cite{afifi2021histogan} metric to directly quantify \textbf{how close a stylized output is to the reference style image in ``style space''}, using global color statistics as a stable proxy for high-level stylistic appearance. This is necessary in our thesis because multi-view stylization methods can achieve strong geometric or structural consistency while still failing to adopt the target style (e.g., preserving scene structure but remaining too close to the original content colors). Therefore, alongside structure- and geometry-based evaluations, we require a metric that isolates \textbf{style adherence}. CHD provides this by measuring the distance between the stylized image and the style exemplar in terms of their RGB color distributions, which captures high-level stylistic semantics such as palette bias, saturation, contrast distribution, and overall chromatic mood.

In our interpretation, CHD measures the distance between the stylized image and the style image in terms of style semantics (approximated through global color histograms). A \textbf{high CHD score} indicates that this distance is large, meaning the stylized output is far from the reference style and thus the method did not successfully transfer the style appearance. Conversely, a \textbf{low CHD score} indicates a small distance, meaning the stylized output is closer to the reference style and shares stronger style/texture semantics. In our experiments, we compute CHD for \textbf{every stylized image} in the dataset with respect to the same reference style image, and then report the \textbf{average CHD} over all images. Hence, a \textbf{lower average CHD} indicates that the method produces outputs that are more faithful to the target style.

Let $p_c$ and $q_c$ be normalized histograms of channel $c\in\{R,G,B\}$ with $B$ bins, computed on the stylized image $\hat{I}$ and style image $S$:
\begin{equation}
p_c \in \mathbb{R}^{B},\quad q_c \in \mathbb{R}^{B},\quad \sum_{b=1}^{B} p_c(b)=1,\quad \sum_{b=1}^{B} q_c(b)=1.
\end{equation}
We use the Hellinger distance:
\begin{equation}
H(p,q) = \frac{1}{\sqrt{2}}\left\|\sqrt{p}-\sqrt{q}\right\|_2,
\end{equation}
and define CHD as the mean over channels:
\begin{equation}
\mathrm{CHD}(\hat{I},S)=\frac{1}{3}\sum_{c\in\{R,G,B\}} H(p_c,q_c).
\end{equation}
For a dataset (or scene) containing $N$ stylized images $\{\hat{I}_i\}_{i=1}^{N}$, we report the average CHD:
\begin{equation}
\mathrm{CHD}_{avg} = \frac{1}{N}\sum_{i=1}^{N}\mathrm{CHD}(\hat{I}_i,S).
\end{equation}
In all tables, \textbf{lower CHD is better}, indicating that the stylized outputs are closer to the style reference in style space.

\subsection{DINO Structure Distance (DSD)}
DSD\cite{tumanyan2022splicing} is used in this thesis to quantify \textbf{how well the stylized image preserves the structure of the content image}, independent of how well it matches the style appearance.
This is necessary because a method can transfer style successfully (e.g., matching the style color/texture statistics) while still distorting geometry-relevant structure such as edges, object boundaries, or spatial layout.
Since our goal is to produce stylized views that remain useful for multi-view and 3D pipelines, we require a structure-focused metric in addition to style-adherence metrics.

We base DSD on frozen DINO ViT features because self-supervised ViT representations encode strong semantic and spatial structure, including information related to semantic segmentation \cite{caron2021emerging}.
Furthermore, \cite{tumanyan2022splicing}.\ (Splice) show that structural information can be represented through self-similarity relationships extracted from deep DINO-ViT features, motivating self-similarity as a principled proxy for structure.

We extract patch tokens from a frozen DINO ViT for the content image $I$ and stylized image $\hat{I}$.
Let $T(I)\in\mathbb{R}^{N\times D}$ denote the L2-normalized patch tokens (excluding the CLS token).
We define the self-similarity matrix:
\begin{equation}
S(I)=T(I)T(I)^\top \in \mathbb{R}^{N\times N}.
\end{equation}
DSD is computed as the normalized Frobenius distance between the two self-similarity matrices (scaled by $100$ for readability):
\begin{equation}
\mathrm{DSD}(I,\hat{I}) = 100\cdot \sqrt{\frac{1}{N^2}\|S(I)-S(\hat{I})\|_F^2}.
\end{equation}
A \textbf{lower} DSD indicates that the stylized image preserves content structure more faithfully, whereas a \textbf{higher} DSD indicates larger structural deviation.

For a dataset (or scene) containing $N_{\mathrm{img}}$ content--stylized pairs $\{(I_i,\hat{I}_i)\}_{i=1}^{N_{\mathrm{img}}}$, we report the mean:
\begin{equation}
\mathrm{DSD}_{avg} = \frac{1}{N_{\mathrm{img}}}\sum_{i=1}^{N_{\mathrm{img}}}\mathrm{DSD}(I_i,\hat{I}_i).
\end{equation}
In all tables, \textbf{lower mean DSD is better}, indicating stronger structure preservation across views.

\section{Dynamic / 3D Metrics from SLAM Reconstructions}
\label{sec:slam_metrics}
Most stylization metrics are \emph{image-level}: they compute a score per image and then average over a dataset.
While useful, such metrics do not capture a key failure mode for multi-view and 3D settings: stylization can preserve structure in a single frame while breaking \emph{inter-frame correspondences} required for tracking and reconstruction.
In practice, 3D pipelines rely on consistent keypoint/descriptor correspondences across time and viewpoints.
If stylization introduces view-dependent hallucinations, shifts edges, or alters repeatable patterns, correspondences can be lost and 3D reconstruction quality can degrade even when per-image structure metrics remain favorable.

To evaluate whether an entire stylized view set still supports 3D reconstruction, we introduce a \emph{SLAM-consistency} evaluation based on DROID-SLAM.
DROID-SLAM estimates camera poses and pixelwise depth/disparity via iterative updates and a dense bundle adjustment layer, allowing us to compare reconstructions from the original and stylized sequences.

\subsection{Trajectory Alignment}
Given camera centers $\{c_i\}$ from the original run and $\{\hat{c}_i\}$ from the stylized run, we estimate a similarity transform $(s,R,t)$ minimizing:
\begin{equation}
\min_{s,R,t}\sum_i \left\|c_i - (sR\hat{c}_i + t)\right\|_2^2,
\end{equation}
where $s\in\mathbb{R}^{+}$, $R\in SO(3)$, and $t\in\mathbb{R}^3$.
We compute $(s,R,t)$ using the standard SVD-based closed form (Umeyama alignment).
This alignment is required because monocular trajectories are defined up to a global similarity transform.

\subsection{Absolute Trajectory Error (ATE)}
After alignment, we compute the absolute trajectory error as:
\begin{equation}
\mathrm{ATE}_{RMSE}=\sqrt{\frac{1}{N}\sum_i \left\|c_i - (sR\hat{c}_i + t)\right\|_2^2 }.
\end{equation}
Lower ATE indicates that the stylized sequence yields a camera trajectory closer to the original, implying that geometric correspondences are better preserved across the stylized set.

\subsection{Rotation Trajectory Error (RTE)}
Let $R_i$ be the original camera rotations and $\hat{R}_i$ the stylized camera rotations.
After applying the global alignment rotation $R$, the per-frame rotation error is:
\begin{equation}
R_i^{err} = R_i^\top (R\hat{R}_i).
\end{equation}
We convert this to an angle:
\begin{equation}
\theta_i = \arccos\left(\frac{\mathrm{trace}(R_i^{err})-1}{2}\right),
\end{equation}
reported in degrees (mean/median). Lower is better.

\subsection{Point Cloud Chamfer Distance (ChD)}
Using DROID-SLAM outputs, we reconstruct point clouds from both runs by back-projecting pixels with disparity (inverse depth), intrinsics, and camera poses (predicted/ground-truth).
Let $P=\{p_k\}_{k=1}^{|P|}$ be the point set from the original reconstruction and $\hat{Q}=\{\hat{q}_\ell\}_{\ell=1}^{|\hat{Q}|}$ the point set from the stylized reconstruction (before alignment), where $p_k,\hat{q}_\ell \in \mathbb{R}^3$.

To ensure the comparison is meaningful under monocular scale ambiguity, we align the stylized point cloud using the \emph{same} similarity transform $(s,R,t)$ estimated during trajectory alignment:
\begin{equation}
Q = \mathcal{A}(\hat{Q}) = \left\{\, q_\ell \;\middle|\; q_\ell = sR\hat{q}_\ell + t,\;\; \hat{q}_\ell \in \hat{Q} \right\}.
\end{equation}
Equivalently, for a matrix form $\hat{Q}\in\mathbb{R}^{|\hat{Q}|\times 3}$ containing points as rows, the aligned point cloud is:
\begin{equation}
Q = s \hat{Q} R^\top + \mathbf{1} t^\top,
\end{equation}
where $\mathbf{1}\in\mathbb{R}^{|\hat{Q}|\times 1}$ is an all-ones vector.

We then compute the symmetric Chamfer distance between $P$ and the aligned stylized cloud $Q$:
\begin{equation}
\mathrm{CD}(P,Q)=
\frac{1}{|P|}\sum_{p\in P}\min_{q\in Q}\|p-q\|_2
+
\frac{1}{|Q|}\sum_{q\in Q}\min_{p\in P}\|q-p\|_2.
\end{equation}
Lower Chamfer distance indicates that the 3D geometry reconstructed from stylized images is closer to the geometry reconstructed from the original images, i.e., the stylization preserves scene structure in a reconstruction-relevant sense.


\chapter{Evaluation}
\label{ch:quant_eval}

\paragraph{How to read the tables.}
All reported numbers are scene-level averages (across frames). We use the notation \texttt{CHD/DSD} for static metrics and \texttt{ATE/RTE/ChD} for dynamic metrics. Unless stated otherwise, \textbf{lower is better} for every entry.

\section{Comparison with State of the Art}
\label{sec:sota_comparison}

We compare our approach against \textbf{MuVieCAST} as a representative multi-view consistent stylization baseline. 

\vspace{0.5em}
\noindent\textbf{Method naming (consistent with earlier sections).}
\begin{itemize}
    \item \textbf{MuVieCAST:} baseline method.
    \item \textbf{Ours-Res + $\mathcal{L}_{Depth}$ + $\mathcal{L}_{SG}$:} residual stylizer trained with depth and SuperPoint/SuperGlue correspondence losses.
    \item \textbf{Ours-U-Net + $\mathcal{L}_{Depth}$ + $\mathcal{L}_{SG}$:} U-Net stylizer trained with depth and SuperPoint/SuperGlue correspondence losses.
\end{itemize}

\begin{table}[t]
\centering
\small
\setlength{\tabcolsep}{3.5pt}
\renewcommand{\arraystretch}{1.15}
\caption{Static metrics (lower is better). Entries are \textbf{CHD / DSD}. 
\textbf{Ours-Res}: Residual stylizer trained with AdaIN+$\mathcal{L}_{Depth}$+$\mathcal{L}_{SG}$.
\textbf{Ours-U-Net}: U-Net stylizer trained with AdaIN+$\mathcal{L}_{Depth}$+$\mathcal{L}_{SG}$. Best values are bolded per metric.}
\label{tab:static_metrics_sota}
\begin{tabular}{lccccc}
\hline
\textbf{Method} 
& \textbf{Horse} 
& \textbf{Panther} 
& \textbf{Lighthouse} 
& \textbf{Francis} 
& \textbf{Train} \\
& \textit{greatwave} & \textit{mosaic} & \textit{abstract} & \textit{abstract2} & \textit{starry} \\
& \textbf{CHD / DSD} & \textbf{CHD / DSD} & \textbf{CHD / DSD} & \textbf{CHD / DSD} & \textbf{CHD / DSD} \\
\hline
MuVieCAST 
& \textbf{0.3042} / 10.2075
& \textbf{0.1238} / 13.1499
& 0.3815 / 19.3052
& 0.1623 / 22.2035
& \textbf{0.1538} / 10.4620 \\
Ours-Res 
& 0.3218 / \textbf{8.6087}
& 0.1747 / \textbf{9.0089}
& 0.3990 / \textbf{15.7968}
& 0.2404 / \textbf{17.4681}
& 0.1805 / \textbf{8.0876} \\
Ours-U-Net 
& 0.3309 / 10.6319
& 0.1543 / 10.6255
& \textbf{0.3350} / 16.7642
& \textbf{0.1471} / 19.7963
& 0.1625 / 9.3424 \\
\hline
\end{tabular}
\end{table}

\begin{table}[t]
\centering
\scriptsize
\setlength{\tabcolsep}{1.9pt}
\renewcommand{\arraystretch}{1.2}
\caption{SLAM-based 3D consistency (lower is better). Each entry is \textbf{ATE/RTE/ChD}. Best values are bolded per metric.}
\label{tab:slam_metrics_sota}
\begin{tabular}{lccccc}
\hline
\textbf{Method} 
& \textbf{Horse} 
& \textbf{Panther} 
& \textbf{Lighthouse} 
& \textbf{Francis} 
& \textbf{Train} \\
& \textit{greatwave} & \textit{mosaic} & \textit{abstract} & \textit{abstract2} & \textit{starry} \\
& \shortstack{\textbf{ATE/RTE/ChD}}
& \shortstack{\textbf{ATE/RTE/ChD}}
& \shortstack{\textbf{ATE/RTE/ChD}}
& \shortstack{\textbf{ATE/RTE/ChD}}
& \shortstack{\textbf{ATE/RTE/ChD}} \\
\hline

MuVieCAST 
& \shortstack{0.058/ 0.469/ 126.91}
& \shortstack{0.005/ 1.159/ 10.31}
& \shortstack{1.081/ 81.156/ 588.64}
& \shortstack{0.752/ 27.273/ 387.54}
& \shortstack{0.810/ 10.017/ 287.34} \\

Ours-Res 
& \shortstack{0.051/ \textbf{0.401}/ \textbf{67.27}}
& \shortstack{0.004/ 0.896/ \textbf{7.72}}
& \shortstack{\textbf{0.917}/ \textbf{48.129}/ \textbf{173.38}}
& \shortstack{\textbf{0.050}/ \textbf{0.697}/ 201.73}
& \shortstack{\textbf{0.433}/ \textbf{4.95}/ \textbf{26.23}} \\

Ours-U-Net 
& \shortstack{\textbf{0.049}/ 0.423/ 89.54}
& \shortstack{\textbf{0.003}/ \textbf{0.633}/ 9.29}
& \shortstack{1.039/ 78.886/ 400.32}
& \shortstack{0.061/ 1.168/ \textbf{136.66}}
& \shortstack{0.711/ 8.63/ 46.85} \\

\hline
\end{tabular}
\end{table}

\subsection{Quantitative comparison}
\label{sec:sota_quantitative}

Tables~\ref{tab:static_metrics_sota} and~\ref{tab:slam_metrics_sota} summarize our evaluation over five scenes (Horse, Panther, Lighthouse, Francis, Train), each stylized with a different reference style image (greatwave, mosaic, abstract, abstract2, starry). For the \textbf{static (image-level)} evaluation, we report \textbf{CHD/DSD} (lower is better). For the \textbf{dynamic/3D} evaluation, we report \textbf{ATE/RTE/ChD} from DROID-SLAM (lower is better).

\paragraph{Scene-wise quantitative analysis (relative change vs.\ MuVieCAST).}

\textbf{Horse (\textit{great wave}).}
Our methods improve SLAM stability: \textbf{ATE} drops by $\approx$12--16\% and \textbf{RTE} by $\approx$10--15\%, while \textbf{Chamfer distance} decreases substantially (about \textbf{47\%} for Ours-Res and \textbf{29\%} for Ours-U-Net). On the image side, Ours-Res reduces \textbf{DSD} by $\approx$16\% (better structure), whereas Ours-U-Net is slightly worse in DSD (a small increase), consistent with its more aggressive low-frequency stylization.

\textbf{Panther (\textit{mosaic}).}
Both variants improve 3D consistency: Ours-Res reduces \textbf{ATE/RTE} by $\approx$20--23\% and \textbf{ChD} by $\approx$25\%; Ours-U-Net yields the lowest \textbf{ATE/RTE} in this scene (about \textbf{40--45\%} reduction), but achieves a smaller \textbf{ChD} reduction ($\approx$10\%). Structurally, both reduce \textbf{DSD} notably (Ours-Res by $\approx$32\%, Ours-U-Net by $\approx$19\%), indicating that our constraints help preserve the object layout even under a style with strong edge/region decomposition.

\textbf{Lighthouse (\textit{abstract}).}
This scene shows a clear separation between architectures. Ours-Res improves SLAM metrics meaningfully (\textbf{ATE} $\approx$15\% lower, \textbf{RTE} $\approx$41\% lower) and yields a large \textbf{ChD} reduction (about \textbf{71\%}), suggesting much closer 3D reconstruction to the RGB reference. Ours-U-Net provides only modest ATE/RTE gains (near-neutral), but still improves \textbf{ChD} by $\approx$32\%, indicating partial geometric recovery. On static metrics, Ours-U-Net also improves \textbf{CHD} (better style alignment) while still reducing \textbf{DSD}, reflecting its higher capacity to match style while our losses limit the worst geometric drift.

\textbf{Francis (\textit{abstract2}).}
This is the most decisive case for downstream 3D usability. Compared to MuVieCAST, both of our variants reduce \textbf{ATE} by $\approx$92--93\% and \textbf{RTE} by $\approx$96--97\%, indicating far more stable tracking. \textbf{ChD} also drops sharply (about \textbf{48\%} for Ours-Res and \textbf{65\%} for Ours-U-Net). Meanwhile, both improve \textbf{DSD} (Ours-Res by $\approx$21\%, Ours-U-Net by $\approx$11\%). Notably, Ours-U-Net also improves \textbf{CHD} here (stronger style alignment than MuVieCAST), while still being geometrically constrained by $\mathcal{L}_{SG}$ and $\mathcal{L}_{Depth}$.

\textbf{Train (\textit{starry}).}
Both variants yield large 3D gains, with Ours-Res being the most stable overall: \textbf{ATE} improves by $\approx$47\%, \textbf{RTE} by $\approx$51\%, and \textbf{ChD} by \textbf{$\approx$91\%}. Ours-U-Net also reduces \textbf{ChD} strongly (about \textbf{84\%}), but has smaller ATE/RTE improvements ($\approx$12--14\%). On static structure, both reduce \textbf{DSD} (Ours-Res by $\approx$23\%, Ours-U-Net by $\approx$11\%), consistent with better preservation of trackable edges and local layout.

\paragraph{Summary of Static metrics (CHD/DSD).}
MuVieCAST achieves the lowest CHD in three scenes (Horse, Panther, Train), indicating strong adherence to the style palette in these cases. However, across all five scenes our method consistently improves \textbf{DSD} (structure preservation) relative to MuVieCAST. This behavior aligns with our objective: we explicitly regularize geometric structure via cross-view correspondences ($\mathcal{L}_{SG}$) and depth consistency ($\mathcal{L}_{Depth}$), which favors preserving edges and local layout cues that DINO self-similarity captures, and considers similar to the RGB counterpart.

\paragraph{Summary of SLAM-based 3D metrics (ATE/RTE/ChD).}
The largest separation appears in the SLAM-based metrics. Across scenes, both of our variants (U-Net and Residual Blocks) reduce trajectory error (ATE/RTE) and substantially reduce Chamfer distance (ChD) compared to MuVieCAST. In particular, the SLAM-based results indicate that our stylized sequences remain substantially closer to the original RGB reconstructions than MuVieCAST, both in terms of camera trajectory consistency and reconstructed 3D structure. In particular, we observe a consistent drop in ChD (e.g., Lighthouse and Train), implying that the point clouds reconstructed from our stylized sequences are geometrically closer to those reconstructed from the original RGB sequences. This supports the core thesis claim that explicitly constraining correspondences and depth improves \emph{downstream 3D usability} of stylized view sets.

\paragraph{Residual vs U-Net stylizer.}
Comparing \textbf{Ours-Res} and \textbf{Ours-U-Net}, the residual stylizer tends to achieve the strongest \textbf{3D geometry} metrics overall (lower ATE/RTE/ChD in most scenes). The U-Net stylizer sometimes yields better \textbf{style alignment} (lower CHD in Lighthouse and Francis), but can be slightly less stable in SLAM reconstruction for some sequences as expresses higher style with a
higher risk of viewpoint-dependent artifacts in weakly structured regions. This indicates a practical trade-off: U-Net capacity can improve global appearance matching, while the residual stylizer often preserves sharper local structure that benefits tracking and reconstruction.

\subsection{Qualitative comparison}
\label{sec:sota_qualitative}

\begin{figure*}[p]
\centering
\includegraphics[width=0.9\textwidth]{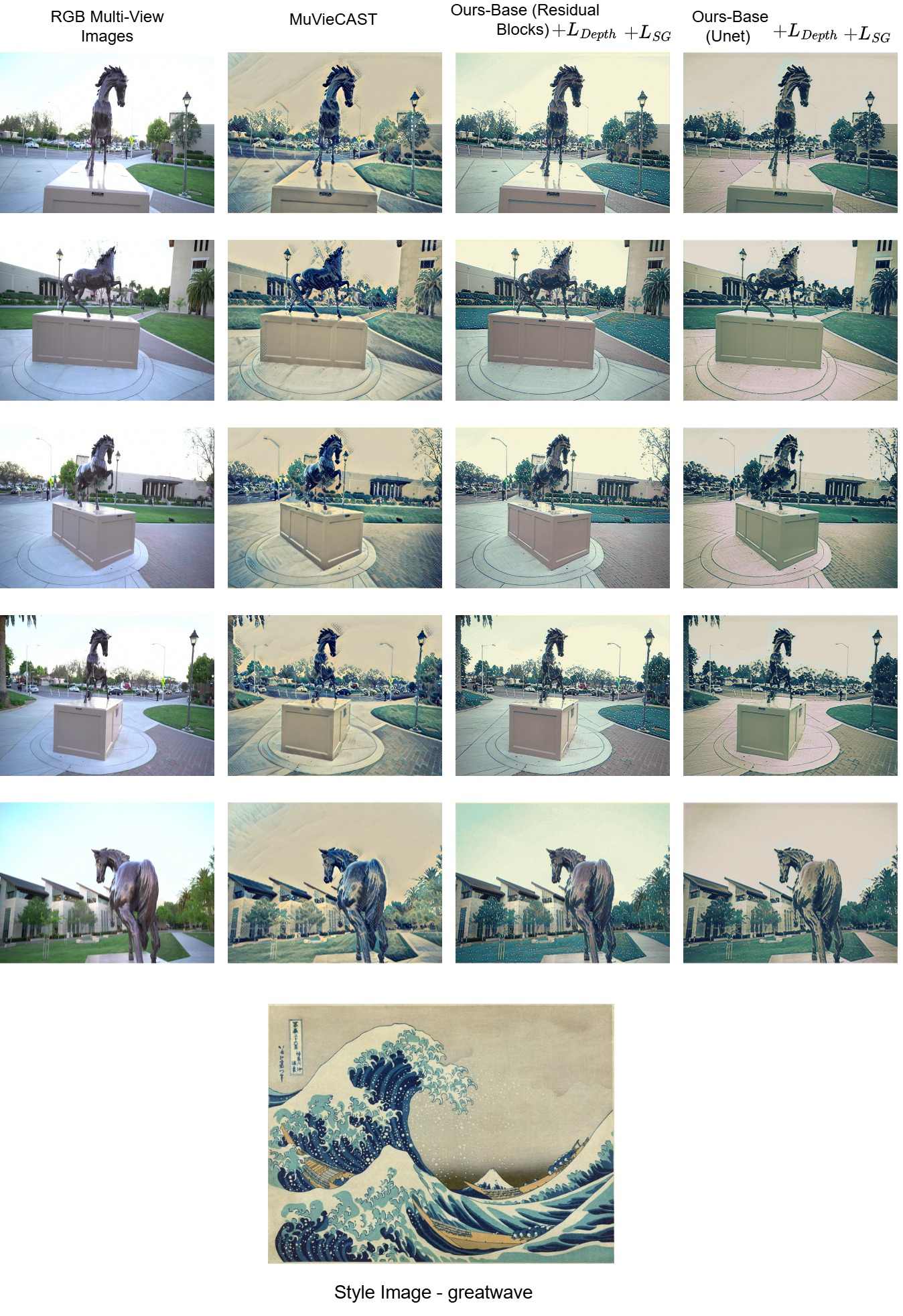}
\caption{Qualitative comparison on \textbf{Horse} (style: \textit{GreatWave}). Columns show the input RGB multi-view images, MuVieCAST, and our variants (Ours-Res and Ours-U-Net) trained with AdaIN+$\mathcal{L}_{Depth}$+$\mathcal{L}_{SG}$.}
\label{fig:qual_horse}
\end{figure*}

\begin{figure*}[p]
\centering
\includegraphics[width=0.9\textwidth]{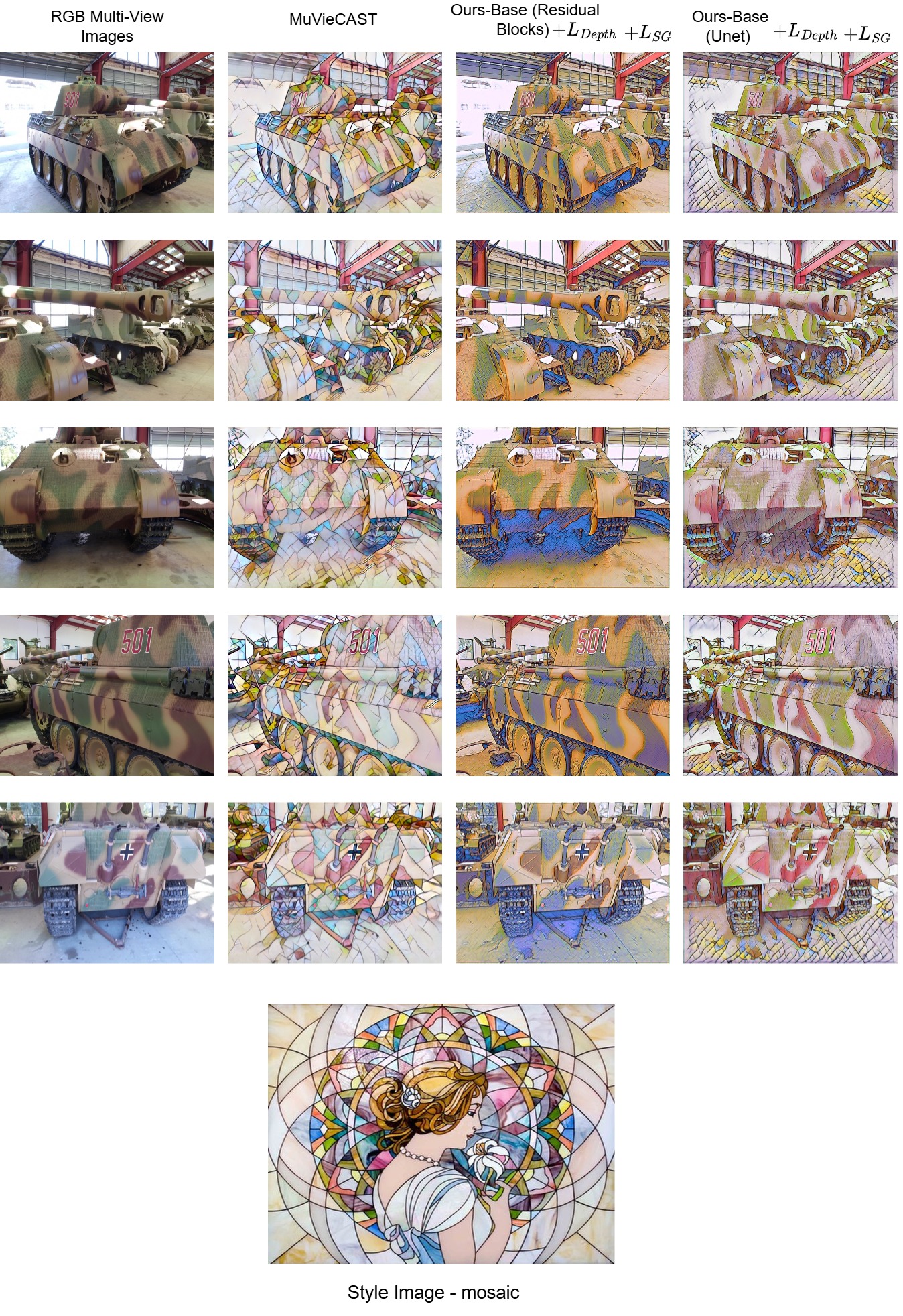}
\caption{Qualitative comparison on \textbf{Panther} (style: \textit{Mosaic}).Columns show the input RGB multi-view images, MuVieCAST, and our variants (Ours-Res and Ours-U-Net) trained with AdaIN+$\mathcal{L}_{Depth}$+$\mathcal{L}_{SG}$.}
\label{fig:qual_panther}
\end{figure*}

\begin{figure*}[p]
\centering
\includegraphics[width=0.9\textwidth]{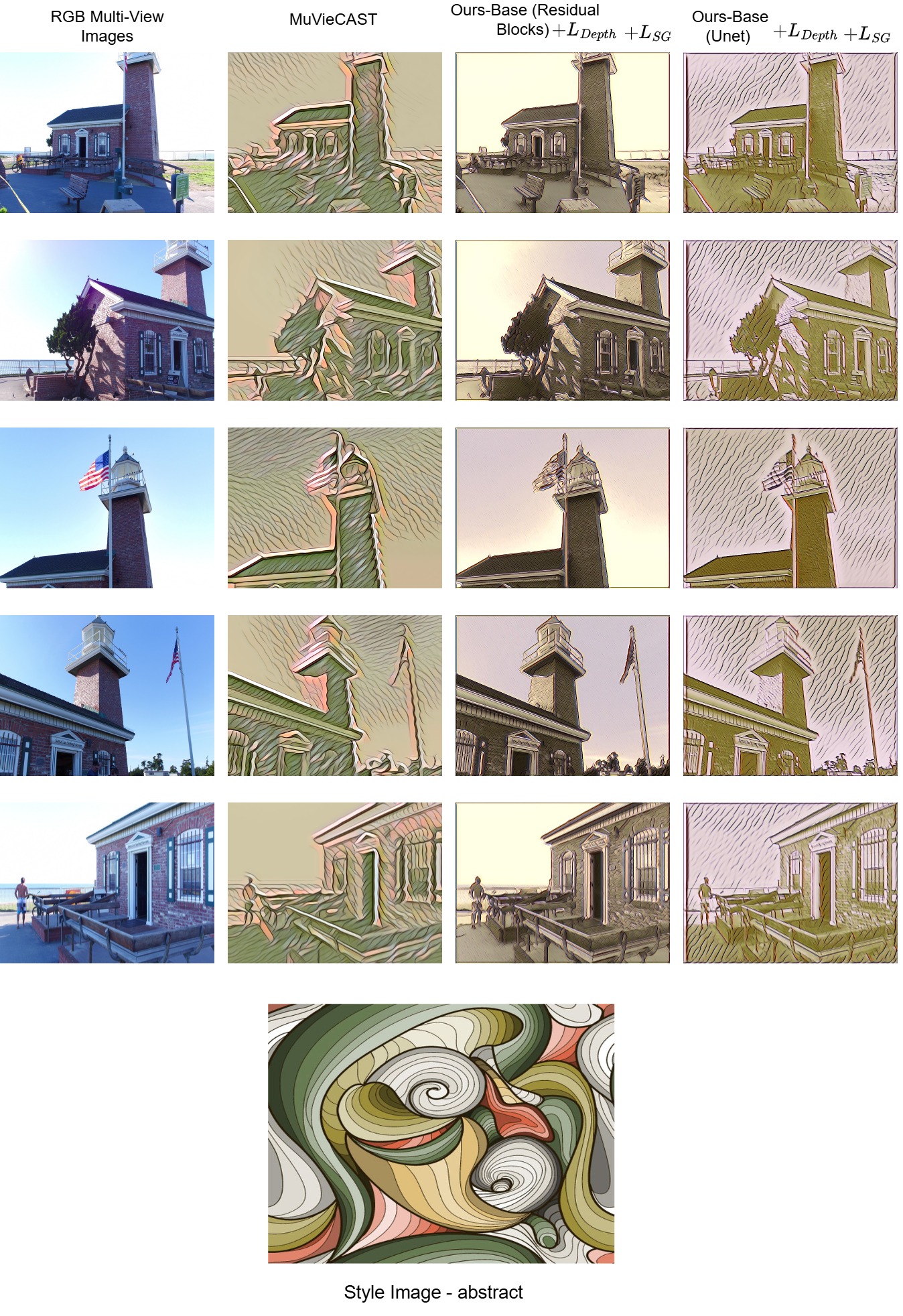}
\caption{Qualitative comparison on \textbf{Lighthouse} (style: \textit{Abstract}). Columns show the input RGB multi-view images, MuVieCAST, and our variants (Ours-Res and Ours-U-Net) trained with AdaIN+$\mathcal{L}_{Depth}$+$\mathcal{L}_{SG}$.}
\label{fig:qual_lighthouse}
\end{figure*}

\begin{figure*}[p]
\centering
\includegraphics[width=0.9\textwidth]{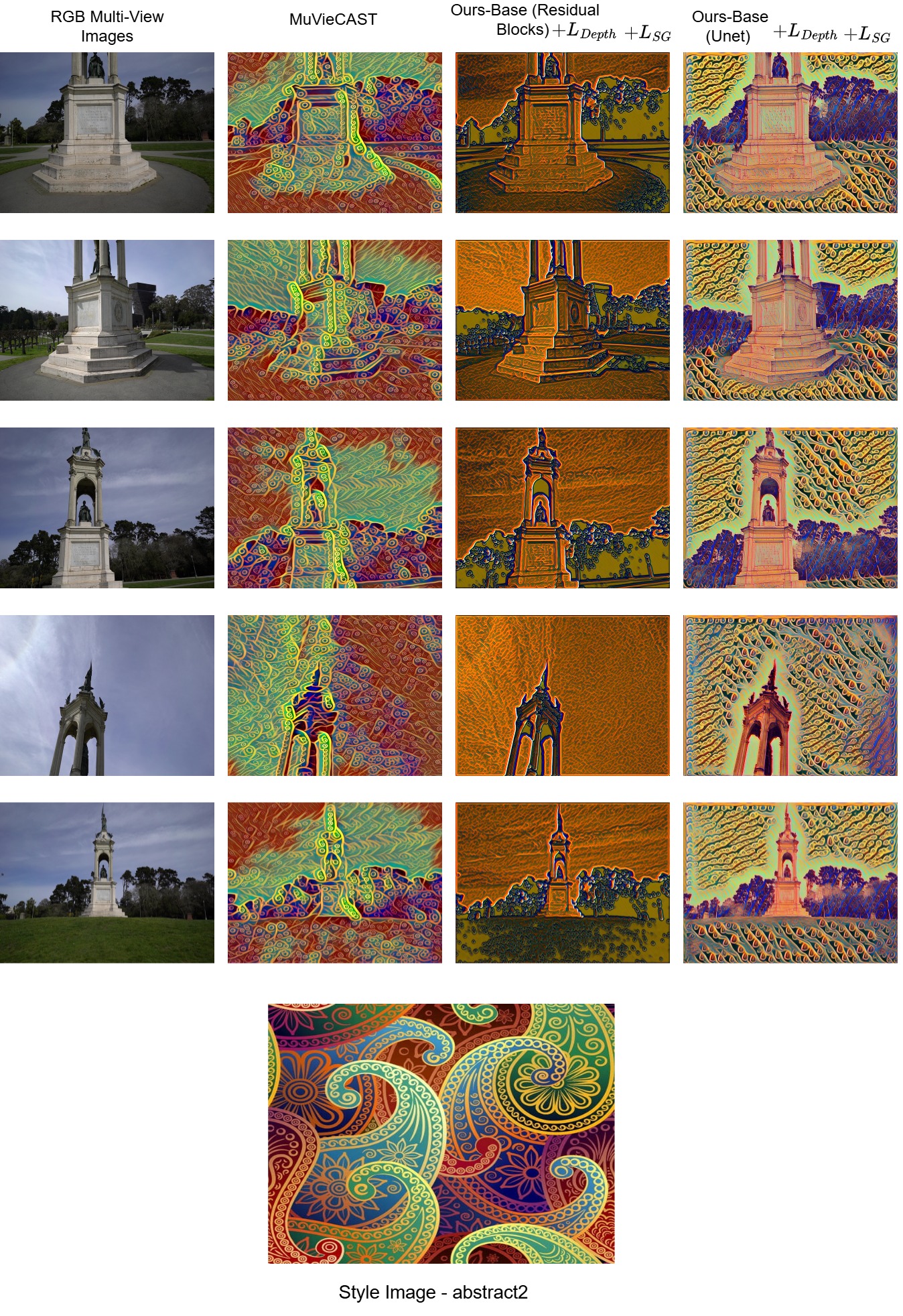}
\caption{Qualitative comparison on \textbf{Francis} (style: \textit{Abstract2}). Columns show the input RGB multi-view images, MuVieCAST, and our variants (Ours-Res and Ours-U-Net) trained with AdaIN+$\mathcal{L}_{Depth}$+$\mathcal{L}_{SG}$.}
\label{fig:qual_francis}
\end{figure*}

\begin{figure*}[p]
\centering
\includegraphics[width=0.9\textwidth]{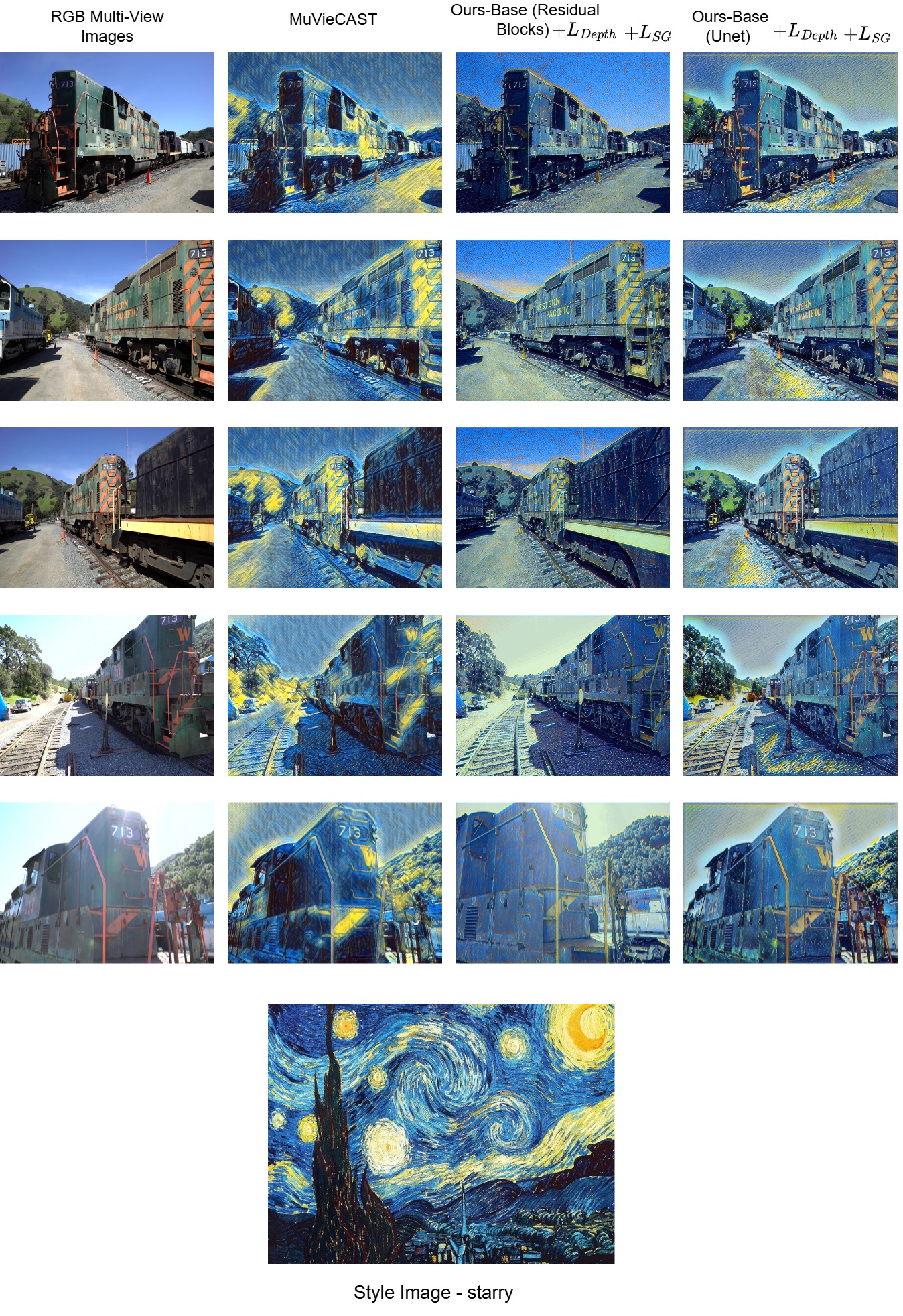}
\caption{Qualitative comparison on \textbf{Train} (style: \textit{Starry}). Columns show the input RGB multi-view images, MuVieCAST, and our variants (Ours-Res and Ours-U-Net) trained with AdaIN+$\mathcal{L}_{Depth}$+$\mathcal{L}_{SG}$.}
\label{fig:qual_train}
\end{figure*}

Figures~\ref{fig:qual_horse}--\ref{fig:qual_train} show representative multi-view results. Across scenes, MuVieCAST frequently produces strong per-image stylization but exhibits view-dependent inconsistencies in high-frequency regions (edge bending, texture “swimming”, and shifting patterns across viewpoints). Our outputs are typically more geometrically robust with stylized textures aligning more consistently with scene structure, and dominant edges remain stable across views. These qualitative behaviors mirror the SLAM-based improvements. The residual stylizer is typically the most stable, while the U-Net variant can be more expressive but may introduce style artifacts in smooth regions.

\paragraph{Scene-wise qualitative analysis.}
\textbf{Horse (\textit{greatwave}, Fig.~\ref{fig:qual_horse}).}
MuVieCAST introduces prominent stylized patterns that vary noticeably across views, especially around thin structures and the pedestal boundary, which can reduce keypoint repeatability. Both of our variants keep the statue silhouette, pedestal corners, and ground-circle boundaries sharper and more consistent. Visually, \textbf{Ours-Res} appears more conservative (less texture hallucination), while \textbf{Ours-U-Net} applies stronger global tone changes but remains relatively coherent.

\textbf{Panther (\textit{mosaic}, Fig.~\ref{fig:qual_panther}).}
MuVieCAST produces heavy mosaic fragmentation that can break long edges and introduce view-dependent piecewise patterns, especially on the tank body and background beams. Our results preserve the tank’s rigid geometry more faithfully: straight lines remain straight and local details stay aligned across views. \textbf{Ours-U-Net} tends to inject more visible mosaic-like line structures, whereas \textbf{Ours-Res} keeps textures more restrained and geometry-forward, consistent with its stronger geometric stability.

\textbf{Lighthouse (\textit{abstract}, Fig.~\ref{fig:qual_lighthouse}).}
MuVieCAST shows characteristic “wavy” deformations and contour drift around architectural boundaries (tower edges, roof lines), which is precisely the kind of high-frequency instability that harms tracking. Our stylized outputs preserve the lighthouse’s dominant lines and keep the appearance changes more consistent across viewpoints. \textbf{Ours-U-Net} adds more spatially varying hatch-like textures (especially in sky regions), while \textbf{Ours-Res} keeps a cleaner abstraction that better respects rigid geometry—matching its large SLAM advantage.

\textbf{Francis (\textit{abstract2}, Fig.~\ref{fig:qual_francis}).}
MuVieCAST exhibits severe view-dependent texture drift: swirling high-frequency patterns shift across viewpoints and can effectively “repaint” edges differently from frame to frame. This qualitative instability matches the large SLAM errors. Our models keep the monument boundaries and pillar edges much more stable. \textbf{Ours-U-Net} is visually the most style-adherent here (also reflected by the best CHD), while \textbf{Ours-Res} remains the most trajectory-stable, suggesting a geometry-vs-expressiveness trade-off.

\textbf{Train (\textit{starry}, Fig.~\ref{fig:qual_train}).}
MuVieCAST introduces strong starry-like strokes that vary in phase across views, especially along long surfaces and track regions, which can cause correspondence failures. Our variants preserve the train contours, rails, and strong perspective lines more consistently. \textbf{Ours-Res} is again the most geometrically stable, producing stylization that is visibly more anchored to object boundaries—consistent with the large improvement in 3D Chamfer distance.

Across scenes, MuVieCAST often produces strong stylization but introduces view-dependent deformations in high-frequency regions (e.g., thin structures, edges, and repeated textures). In contrast, our method tends to preserve sharper boundaries and more stable local patterns across views. This visual behavior is consistent with the improvements observed in DSD and SLAM-based metrics: preserving matchable local structure and maintaining depth-consistent appearance reduces geometry drift during reconstruction.

\paragraph{Residual stylizer (Ours-Res): conservative stylization with strong geometric retention.}
The residual stylizer tends to preserve spatial layout and low-level structure more faithfully. Its limited capacity encourages a style transfer that is closer to ``absorbing the essence'' of the style (palette shifts, edge emphasis, texture bias) without aggressively hallucinating new patterns in regions that lack strong geometric support. This conservative behavior is beneficial in multi-view settings: it reduces the chance of view-dependent artifacts and helps keep correspondences stable. In practice, Ours-Res often provides the best trade-off when the primary goal is geometry preservation for 3D pipelines.

\paragraph{U-Net stylizer (Ours-U-Net): stronger appearance changes but higher risk in low-frequency regions.}
The U-Net variant, due to its encoder--decoder structure and skip connections, can apply more global and spatially varying appearance transformations. This can produce visually richer stylization, but it can also introduce artifacts in \emph{low-frequency or weakly textured regions} where there is little geometric evidence to anchor the style. As observed (notably in Francis with Abstract2), the U-Net stylizer may lose some low-frequency components that do not carry strong geometric cues and may replace them with style-driven artifacts. Importantly, this does not typically break strong edges (which are still protected by $\mathcal{L}_{SG}$ and $\mathcal{L}_{Depth}$), but it can reduce the perceptual coherence of smooth regions and introduce minor inconsistencies across views.

\paragraph{Why MuVieCAST fails more severely than our U-Net in such cases.}
While the U-Net can occasionally introduce style artifacts in weak-structure regions, it is still constrained by our explicit correspondence and depth regularization, which limits geometric drift. MuVieCAST often optimizes for stylistic strength per-view, but is more vulnerable to cross-view drift in geometry-critical high-frequency regions as it is not explicitly tied to stable cross-view correspondences for local structure preservation. As a result the geometry breaks, when the style strongly perturbs edges and repeated patterns (as with Abstract2), MuVieCAST can fail more severely: it disrupts keypoint repeatability and produces view-dependent distortions that degrade reconstruction.

\paragraph{Why our method is more geometrically stable than MuVieCAST.}
MuVieCAST encourages multi-view consistency largely at the \emph{appearance level}, aiming for highly stylized but visually compatible renderings across viewpoints. However, it does not explicitly preserve the \emph{repeatable local evidence} that feature-based 3D pipelines depend on. Under strong stylization, high-frequency cues such as corners, junctions, thin edges, and small texture elements can shift in phase, become over-smoothed, or be replaced by style-induced patterns. These cues are precisely what keypoint detectors and descriptors require to remain stable across views; when they are perturbed, the stylized sequence becomes \emph{less matchable}, leading to noisier correspondences and higher ambiguity.

A related failure mode is \emph{view-dependent stylization drift}. Brush strokes, mosaic boundaries, or dense texture patterns may look plausible in each individual image, yet their placement can vary subtly with viewpoint (“texture swimming”). This breaks the standard multi-view assumption that appearance changes are explained primarily by geometry and illumination, so SLAM/MVS may interpret these deviations as motion or structural inconsistency. The effect becomes especially pronounced under wide baselines and in repetitive regions (e.g., architectural edges, fences, brick-like textures), where even small local shifts can cause large matching errors and downstream reconstruction drift.

Our approach is more geometrically stable because it targets these failure modes directly with two complementary constraints. 1) $\mathcal{L}_{SG}$ explicitly anchors stylization to \emph{cross-view feature correspondences} by enforcing descriptor consistency on high-confidence SuperPoint/SuperGlue matches, reducing keypoint drift and improving match repeatability under stylization. 2) $\mathcal{L}_{Depth}$ discourages appearance changes that distort depth cues, helping maintain depth-consistent structure and stabilizing 3D fusion. Together, these losses suppress view-dependent texture sliding and edge bending, producing stylized views whose high-frequency structure remains aligned across viewpoints and thereby improving cross-view tracking and yielding a more coherent reconstruction than MuVieCAST.

In summary, the comparison indicates that our method trades a small amount of pure style aggressiveness (in some scenes) for substantially improved multi-view geometric integrity. This trade-off is desirable for the thesis objective: producing stylized multi-view sets that remain compatible with downstream 3D reconstruction pipelines.

\subsection{Trajectory Visualization from DROID-SLAM Reconstructions}
\label{subsec:traj_viz}

To complement the SLAM-based scalar metrics (ATE/RTE) with an interpretable geometric diagnostic, we visualize the estimated camera trajectories produced by DROID-SLAM for the original RGB sequence, MuVieCAST stylizations, and our stylized outputs (Residual stylizer and U-Net stylizer). Each trajectory figure is a composite plot containing both \textbf{3D trajectories} and \textbf{2D projections} under two different alignment references:

\begin{enumerate}
    \item \textbf{Alignment to Ground Truth (GT) poses (left column).} When GT poses are available, we align each method’s trajectory to the GT coordinate frame (via a similarity alignment). This view highlights absolute geometric faithfulness to the true camera motion.
    \item \textbf{Alignment to RGB/DROID-SLAM poses (right column).} We also align trajectories to the pose sequence obtained by running DROID-SLAM on the original RGB images. This isolates the \emph{effect of stylization on SLAM stability} by comparing how much each stylized trajectory deviates from the ``best available'' monocular SLAM solution on the unmodified sequence.
\end{enumerate}

In each figure, we plot a 3D trajectory (top row) and three 2D projections (bottom rows: top-down $X$--$Y$, side $X$--$Z$, and front $Y$--$Z$). Start and end points are marked consistently (start as a circle and end as a cross), enabling quick inspection of drift, scale instability, loop deformation, or catastrophic tracking divergence. Across scenes, trajectories that remain close to the RGB/GT reference indicate that stylization preserved \emph{repeatable local structure}, \emph{stable correspondences}, and \emph{consistent depth cues} required by the SLAM backend.

\subsubsection{Per-scene observations}
\paragraph{Francis (style: abstract2).}
The Francis sequence reveals a clear failure mode for MuVieCAST: the estimated trajectory deviates substantially from both the GT-aligned and RGB-aligned references, indicating correspondence instability and drift under stylization. In contrast, our correspondence- and depth-regularized models maintain noticeably tighter agreement with the reference motion. Qualitatively, the Residual stylizer tends to preserve geometry-critical edges more conservatively, whereas the U-Net variant can apply stronger stylization on low-frequency regions, occasionally introducing mild appearance artifacts where geometric evidence is weak. Despite these stylistic differences, both of our variants yield significantly more stable trajectories than MuVieCAST, consistent with the lower ATE/RTE and improved reconstruction fidelity reported in the quantitative tables.

\paragraph{Lighthouse (style: abstract).}
Lighthouse is challenging due to large planar regions and repeated patterns, where viewpoint-consistent structure is essential. The MuVieCAST trajectory exhibits pronounced deviation and partial failure behavior, consistent with broken correspondences and unstable depth updates. Our method reduces this effect: the trajectory shape remains closer to the reference motion in both 3D and 2D projections, suggesting that the combined SuperGlue correspondence constraint and depth preservation help prevent the SLAM system from locking onto stylization-induced ambiguities.

\paragraph{Train (style: starry).}
For Train, the RGB-aligned plots are particularly informative. MuVieCAST shows a noticeable excursion away from the main path, suggesting partial tracking loss or accumulated drift after stylization. Both of our variants remain closer to the RGB reference trajectory, with the Residual stylizer typically showing the most conservative adherence. This supports the interpretation that the residual architecture often absorbs the essence of the style while keeping geometry-relevant cues (edges, corners, and repeatable patterns) intact, whereas stronger stylization capacity can sometimes suppress weak geometric evidence.

\paragraph{Horse (style: greatWave).}
In Horse, MuVieCAST again deviates more from the RGB reference than our outputs. Our trajectories overlap more consistently with both GT and RGB references in the 2D projections, indicating improved cross-view matchability and reduced pose drift under stylization. The Residual and U-Net variants remain close to each other, with small differences reflecting the trade-off between stronger stylization capacity (U-Net) and conservative structure retention (Residual).

\paragraph{Panther (style: mosaic).}
Panther is comparatively stable across methods: the trajectories remain close to the reference motion with only minor deviations. This is consistent with the quantitative results where error margins are already low. Even in this stable case, our variants remain competitive and generally show reduced drift relative to MuVieCAST, indicating that the added geometric constraints do not degrade performance when stylization is already SLAM-friendly.


\begin{figure}[p]
    \centering
    \includegraphics[width=\textwidth]{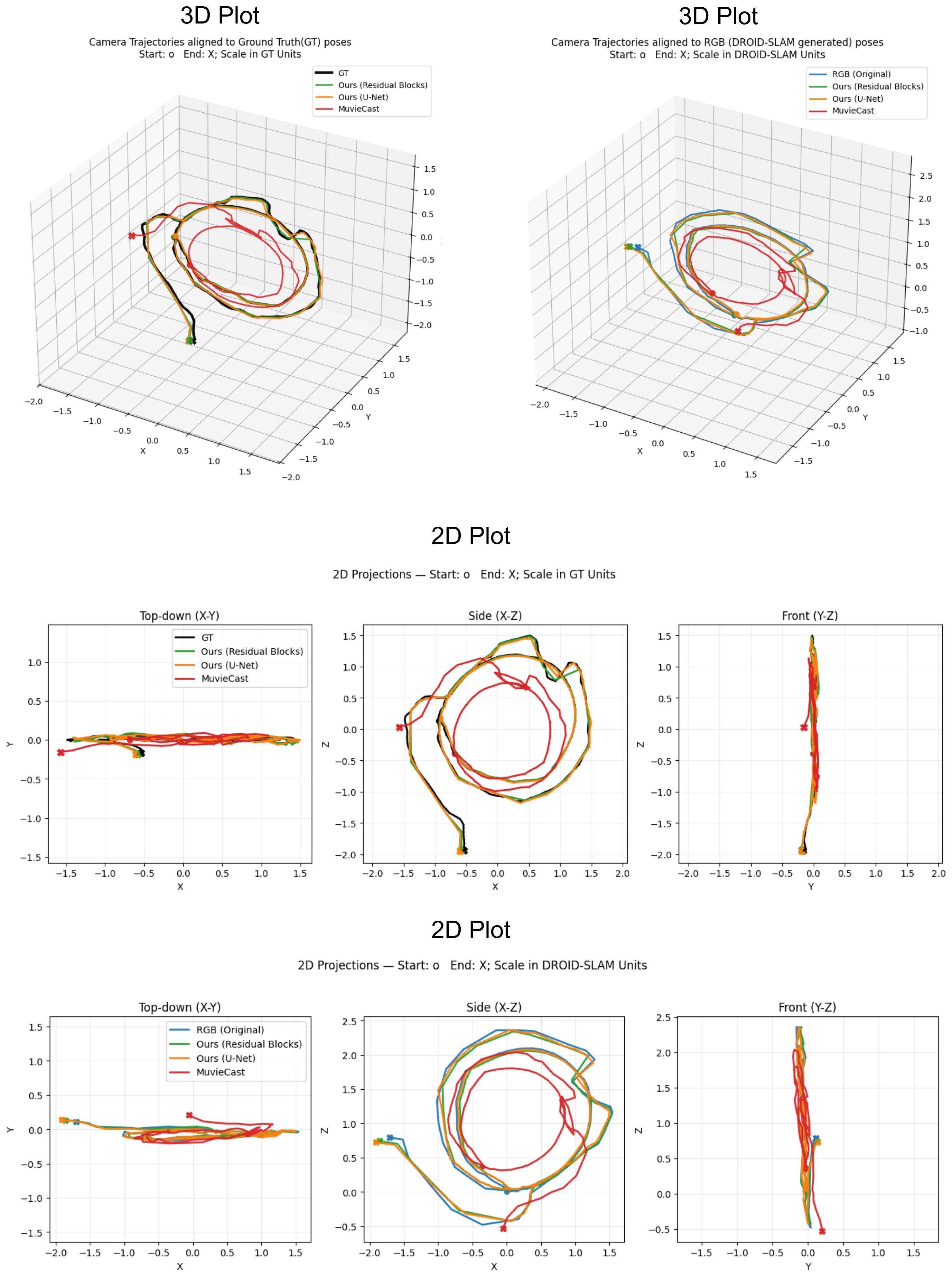}
    \caption{DROID-SLAM trajectory comparison on \textbf{Francis} (style: abstract2).}
    \label{fig:traj_francis}
\end{figure}
\clearpage

\begin{figure}[p]
    \centering
    \includegraphics[width=\textwidth]{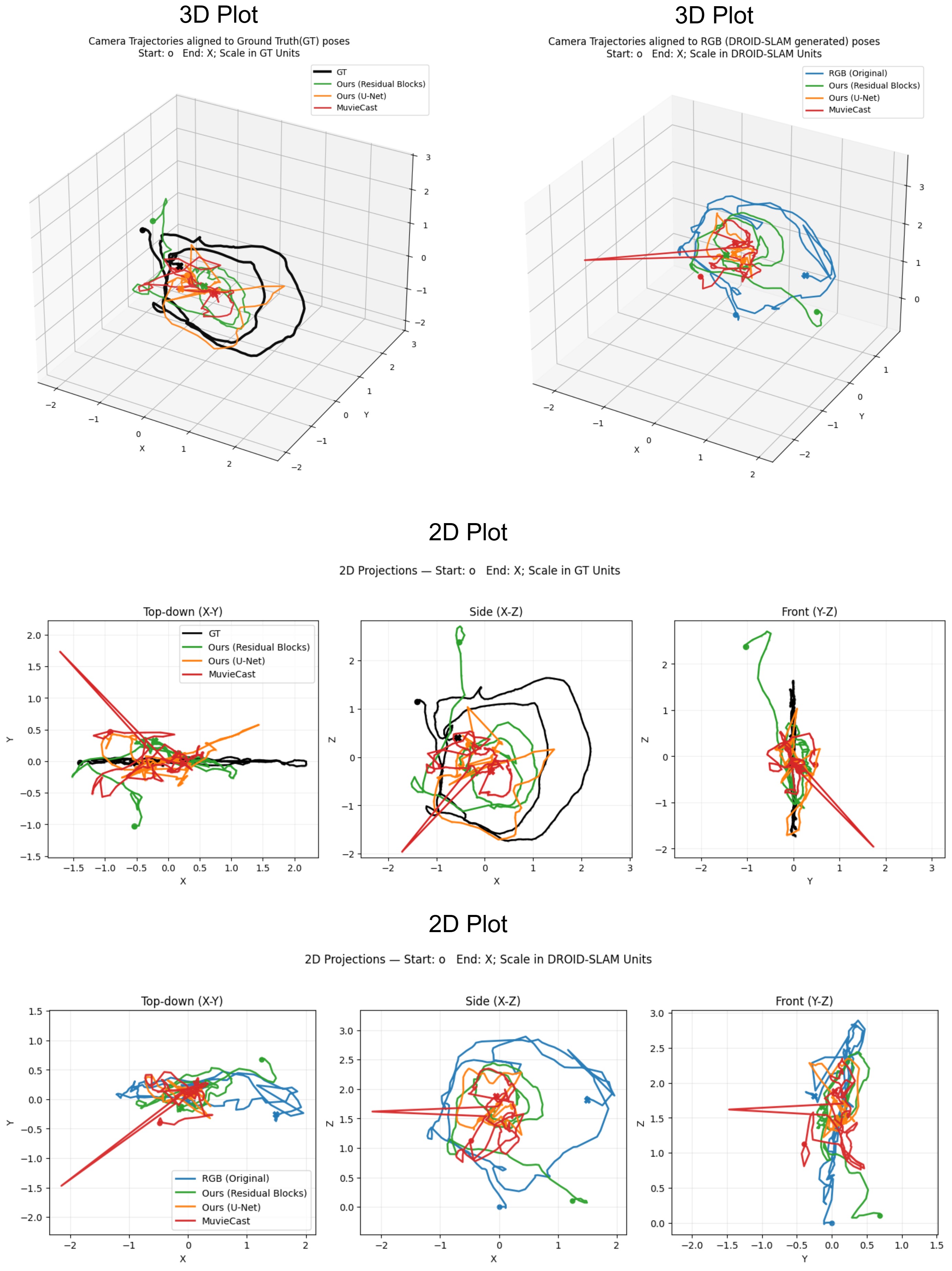}
    \caption{DROID-SLAM trajectory comparison on \textbf{Lighthouse} (style: abstract).}
    \label{fig:traj_lighthouse}
\end{figure}
\clearpage

\begin{figure}[p]
    \centering
    \includegraphics[width=\textwidth]{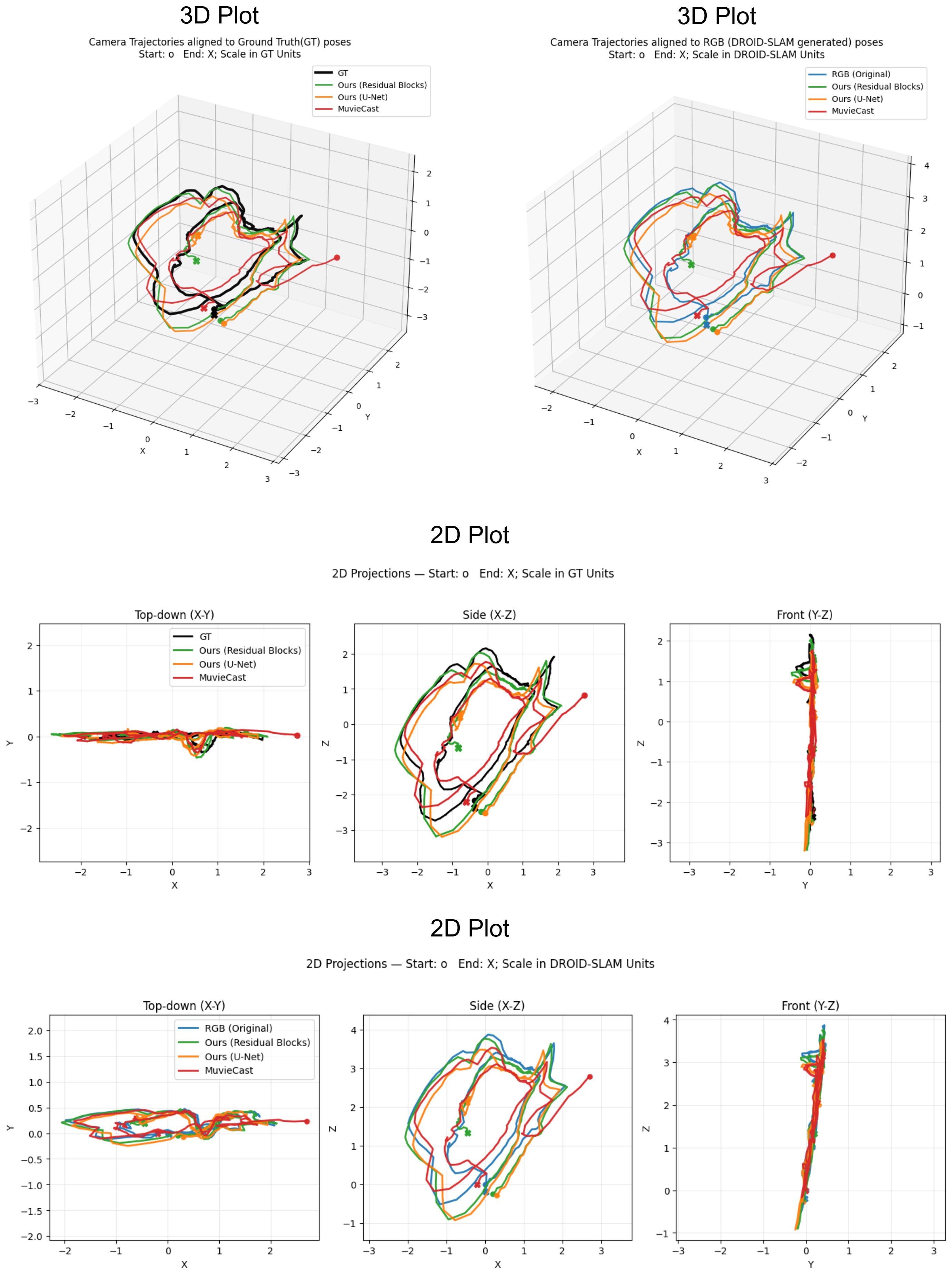}
    \caption{DROID-SLAM trajectory comparison on \textbf{Train} (style: starry).}
    \label{fig:traj_train}
\end{figure}
\clearpage

\begin{figure}[p]
    \centering
    \includegraphics[width=\textwidth]{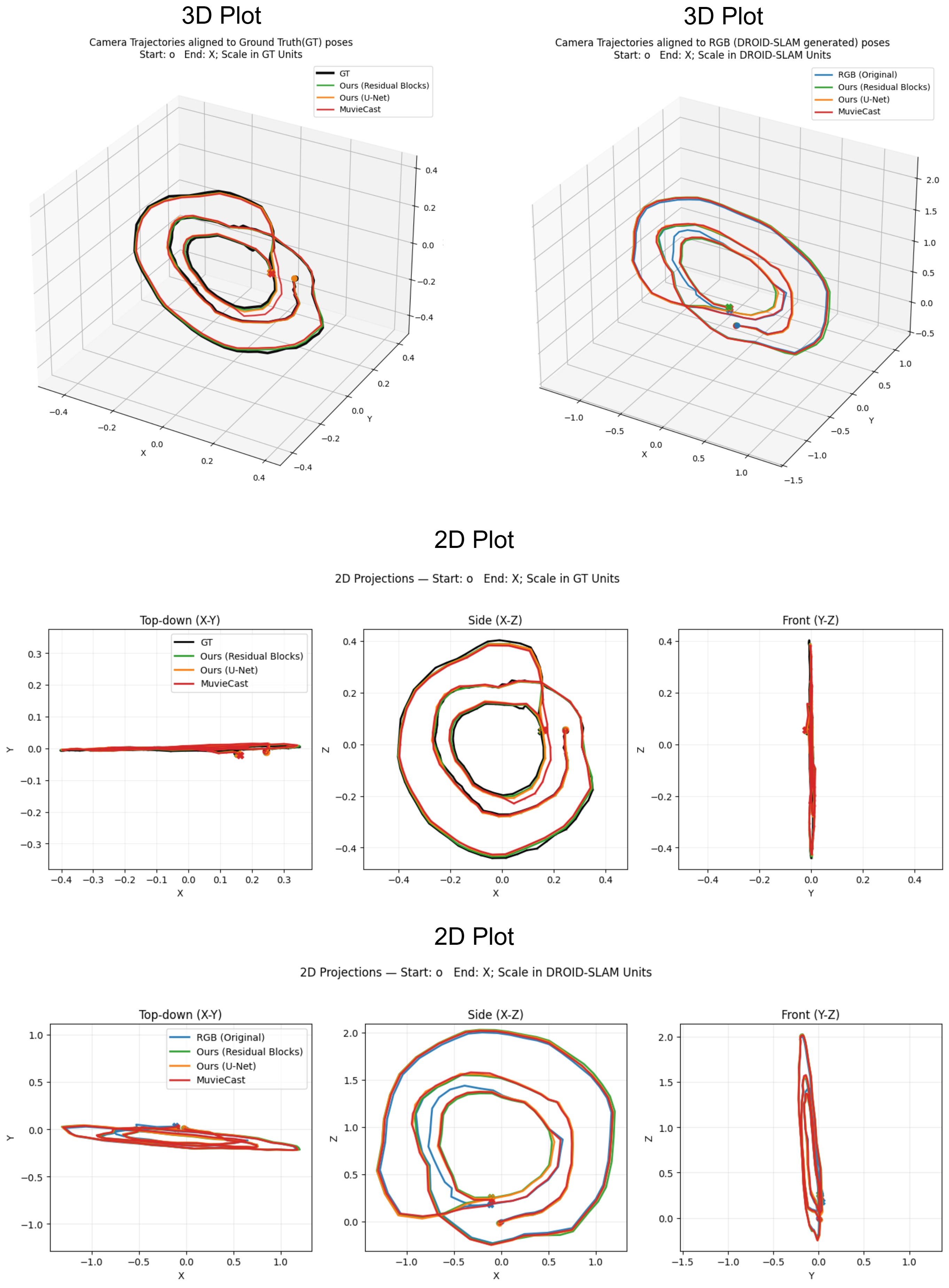}
    \caption{DROID-SLAM trajectory comparison on \textbf{Horse} (style: greatwave).}
    \label{fig:traj_horse}
\end{figure}
\clearpage

\begin{figure}[p]
    \centering
    \includegraphics[width=\textwidth]{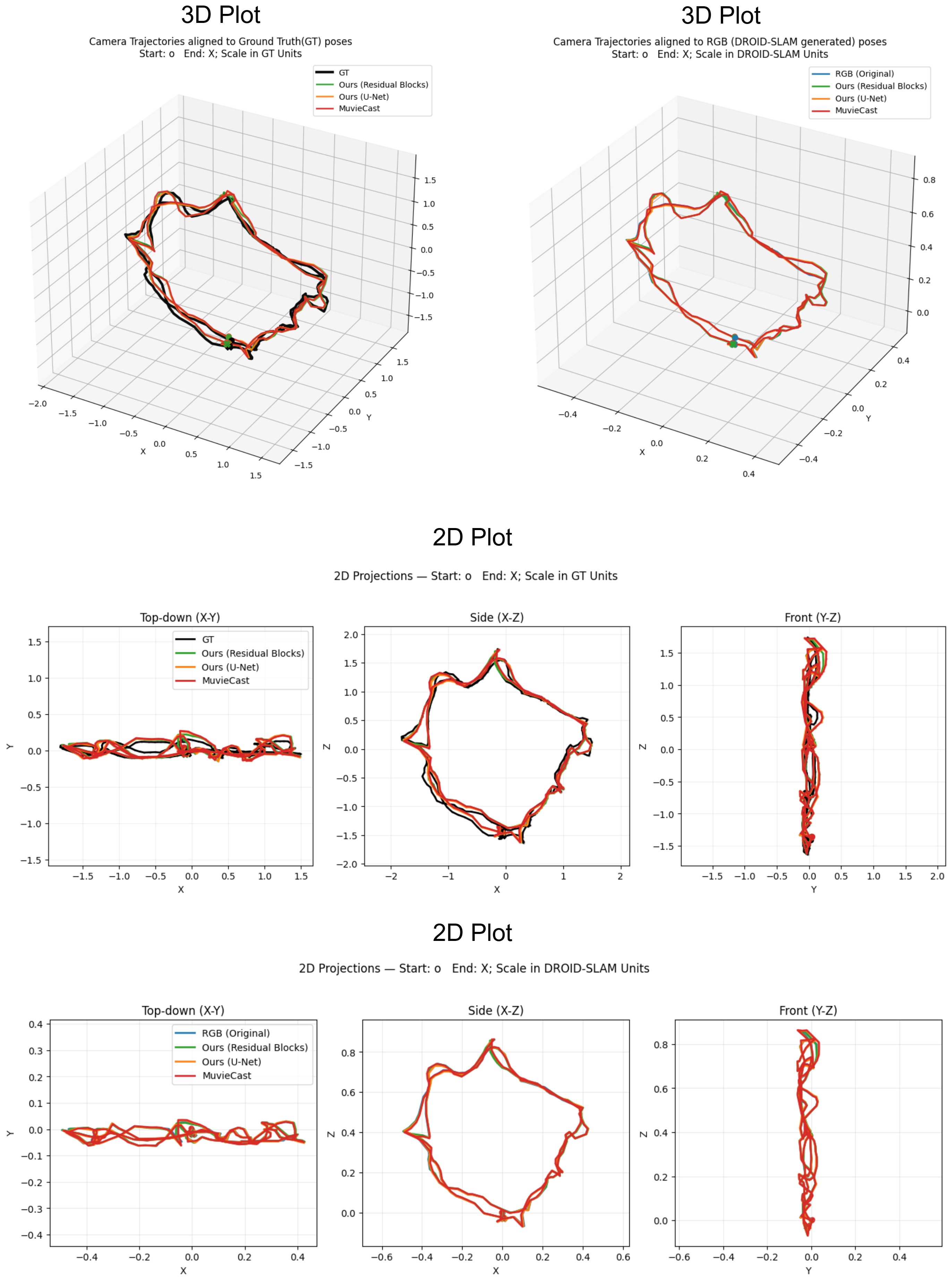}
    \caption{DROID-SLAM trajectory comparison on \textbf{Panther} (style: mosaic).}
    \label{fig:traj_panther}
\end{figure}
\clearpage


\section{Ablation Study: Effect of Depth and SuperGlue Constraints}
\label{sec:ablation_quant}

\paragraph{Setup.}
 The ablations are run on Tanks and Temples (\textit{Train}, \textit{Truck}) and Mip-NeRF 360 (\textit{Garden}, \textit{Bicycle}, \textit{Counter}) scenes.

We ablate the contributions of our two geometry-oriented regularizers: the depth-preservation term $\mathcal{L}_{Depth}$ and the correspondence consistency term $\mathcal{L}_{SG}$. We report (i) \textbf{static} image-level metrics (CHD/DSD) and (ii) \textbf{dynamic/3D} SLAM-consistency metrics (ATE/RTE/ChD), where lower is better in all cases. The \textbf{Base} denotes the stylizer model initialized with residual blocks and also corresponds to AdaIN-driven stylization (content + style statistics matching) without explicit geometric constraints. The other variants incrementally add $\mathcal{L}_{Depth}$ and/or $\mathcal{L}_{SG}$.

\begin{table}[t]
\centering
\footnotesize
\setlength{\tabcolsep}{2.2pt}
\renewcommand{\arraystretch}{1.15}
\caption{Ablation on \textbf{Train} and \textbf{Truck}: static metrics (lower is better). Each entry is \textbf{CHD/DSD}. Best (lowest) CHD and DSD per column are bolded.}
\label{tab:ablation_static_train_truck}
\begin{tabular}{lcccc}
\hline
\textbf{Method} &
\shortstack{\textbf{Train}\\\textit{style 19}\\\textbf{CHD/DSD}} &
\shortstack{\textbf{Train}\\\textit{starry}\\\textbf{CHD/DSD}} &
\shortstack{\textbf{Truck}\\\textit{style 19}\\\textbf{CHD/DSD}} &
\shortstack{\textbf{Truck}\\\textit{starry}\\\textbf{CHD/DSD}} \\
\hline
Ours (Base) &
\shortstack{\textbf{0.3268}/11.8940} &
\shortstack{\textbf{0.1620}/9.2920} &
\shortstack{\textbf{0.3060}/9.2320} &
\shortstack{\textbf{0.2121}/8.8169} \\
Ours (Base + $\mathcal{L}_{Depth}$) &
\shortstack{0.3492/9.4581} &
\shortstack{0.1767/8.6051} &
\shortstack{0.3177/8.9358} &
\shortstack{0.2347/8.5099} \\
Ours (Base + $\mathcal{L}_{SG}$) &
\shortstack{0.3521/8.4576} &
\shortstack{0.1843/8.4328} &
\shortstack{0.3409/8.8692} &
\shortstack{0.2464/7.8745} \\
Ours (Base + $\mathcal{L}_{Depth}$ + $\mathcal{L}_{SG}$) &
\shortstack{0.3568/\textbf{8.4146}} &
\shortstack{0.1895/\textbf{8.0876}} &
\shortstack{0.3488/\textbf{8.5543}} &
\shortstack{0.2581/\textbf{7.3641}} \\
\hline
\end{tabular}
\end{table}

\begin{table}[t]
\centering
\scriptsize
\setlength{\tabcolsep}{1.8pt}
\renewcommand{\arraystretch}{1.15}
\caption{Ablation on \textbf{Train} and \textbf{Truck}: DROID SLAM-based 3D consistency (lower is better). Each entry is \textbf{ATE/RTE/ChD}. Best (lowest) ATE, RTE, and ChD per column are bolded.}
\label{tab:ablation_slam_train_truck}
\begin{tabular}{lcccc}
\hline
\textbf{Method} &
\shortstack{\textbf{Train}\\\textit{style 19}\\\textbf{ATE/RTE/ChD}} &
\shortstack{\textbf{Train}\\\textit{starry}\\\textbf{ATE/RTE/ChD}} &
\shortstack{\textbf{Truck}\\\textit{style 19}\\\textbf{ATE/RTE/ChD}} &
\shortstack{\textbf{Truck}\\\textit{starry}\\\textbf{ATE/RTE/ChD}} \\
\hline
Ours (Base) &
\shortstack{0.824/26.418/580.51} &
\shortstack{0.854/23.904/103.82} &
\shortstack{0.0375/1.065/16.20} &
\shortstack{0.055/1.823/9.71} \\
Ours (Base + $\mathcal{L}_{Depth}$) &
\shortstack{0.809/20.273/70.14} &
\shortstack{0.755/22.045/66.75} &
\shortstack{0.029/0.625/15.25} &
\shortstack{0.021/0.971/8.33} \\
Ours (Base + $\mathcal{L}_{SG}$) &
\shortstack{0.526/16.697/56.71} &
\shortstack{0.536/17.233/44.63} &
\shortstack{0.027/0.609/14.21} &
\shortstack{0.019/0.881/7.56} \\
Ours (Base + $\mathcal{L}_{Depth}$ + $\mathcal{L}_{SG}$) &
\shortstack{\textbf{0.018}/\textbf{0.870}/\textbf{48.35}} &
\shortstack{\textbf{0.433}/\textbf{4.95}/\textbf{26.23}} &
\shortstack{\textbf{0.022}/\textbf{0.587}/\textbf{13.19}} &
\shortstack{\textbf{0.014}/\textbf{0.724}/\textbf{6.37}} \\
\hline
\end{tabular}
\end{table}

\subsection{Quantitative Analysis: Train and Truck Scenes}
Train is the most revealing case for understanding why explicit geometric constraints are necessary. With the \textbf{Base} stylizer, DROID-SLAM exhibits severe instability (e.g., Train--style\_19 yields very large ATE/RTE and an extremely high Chamfer distance), indicating that stylization alone can disrupt repeatable features and induce depth/pose drift. Adding $\mathcal{L}_{Depth}$ drastically reduces the Chamfer distance (from 580.51 to 70.14) and improves rotation error, showing that constraining monocular depth cues prevents large-scale volumetric deformation. Adding $\mathcal{L}_{SG}$ further improves all 3D metrics (e.g., Train--style\_19 Chamfer 56.71), consistent with our design goal: enforcing descriptor stability under SuperGlue matches preserves cross-view matchability needed for tracking and bundle adjustment. The \textbf{full model} (Base + $\mathcal{L}_{Depth}$ + $\mathcal{L}_{SG}$) achieves the best 3D consistency across both styles, demonstrating a strong synergy: depth regularization stabilizes global shape cues, while correspondence regularization stabilizes local, repeatable structure.

In the static metrics, CHD is lowest for the Base model across Train/Truck, while DSD improves monotonically as we add geometry losses. This reflects an expected trade-off: the geometry constraints restrict aggressive appearance changes that would otherwise push the output closer to the style color distribution, but in exchange they preserve structural layout and reduce distortions relevant for multi-view consistency.

Truck is comparatively easier: even the Base stylizer yields a trackable sequence, but both $\mathcal{L}_{Depth}$ and $\mathcal{L}_{SG}$ produce consistent gains. The full model yields the lowest ATE/RTE and Chamfer under both styles, confirming that the geometric constraints remain beneficial without harming stability in already well-conditioned sequences.

\begin{table}[t]
\centering
\footnotesize
\setlength{\tabcolsep}{1.6pt}
\renewcommand{\arraystretch}{1.15}
\caption{Ablation on \textbf{Garden} and \textbf{Counter}: static metrics (lower is better). Each entry is \textbf{CHD/DSD}. Best (lowest) CHD and DSD per column are bolded.}
\label{tab:ablation_static_garden_counter}
\begin{tabular}{lccccc}
\hline
\textbf{Method} &
\shortstack{\textbf{Garden}\\\textit{style 19}\\\textbf{CHD/DSD}} &
\shortstack{\textbf{Garden}\\\textit{starry}\\\textbf{CHD/DSD}} &
\shortstack{\textbf{Counter}\\\textit{style 19}\\\textbf{CHD/DSD}} &
\shortstack{\textbf{Counter}\\\textit{starry}\\\textbf{CHD/DSD}} &
\shortstack{\textbf{Counter}\\\textit{style\_2}\\\textbf{CHD/DSD}} \\
\hline
Ours (Base) &
\shortstack{\textbf{0.3215}/13.8940} &
\shortstack{\textbf{0.5205}/11.2367} &
\shortstack{0.3903/8.2029} &
\shortstack{0.2530/9.0322} &
\shortstack{\textbf{0.1867}/9.8447} \\
Ours (Base + $\mathcal{L}_{Depth}$) &
\shortstack{0.3287/13.6059} &
\shortstack{0.5353/10.1142} &
\shortstack{0.5464/9.0010} &
\shortstack{\textbf{0.2342}/9.0809} &
\shortstack{0.4229/9.1835} \\
Ours (Base + $\mathcal{L}_{SG}$) &
\shortstack{0.3351/12.7059} &
\shortstack{0.5424/9.8853} &
\shortstack{0.3872/\textbf{7.8421}} &
\shortstack{0.2572/9.3280} &
\shortstack{0.2516/8.2836} \\
Ours (Base + $\mathcal{L}_{Depth}$ + $\mathcal{L}_{SG}$) &
\shortstack{0.3370/\textbf{12.4668}} &
\shortstack{0.5434/\textbf{9.0919}} &
\shortstack{\textbf{0.3750}/7.9625} &
\shortstack{0.2564/\textbf{8.7614}} &
\shortstack{0.2771/\textbf{7.3569}} \\
\hline
\end{tabular}
\end{table}

\begin{table}[t]
\centering
\scriptsize
\setlength{\tabcolsep}{1.3pt}
\renewcommand{\arraystretch}{1.15}
\caption{Ablation on \textbf{Garden} and \textbf{Counter}: DROID SLAM-based 3D consistency (lower is better). Each entry is \textbf{ATE/RTE/ChD}. Best (lowest) ATE, RTE, and ChD per column are bolded.}
\label{tab:ablation_slam_garden_counter}
\begin{tabular}{lccccc}
\hline
\textbf{Method} &
\shortstack{\textbf{Garden}\\\textit{style 19}\\\textbf{ATE/RTE/ChD}} &
\shortstack{\textbf{Garden}\\\textit{starry}\\\textbf{ATE/RTE/ChD}} &
\shortstack{\textbf{Counter}\\\textit{style 19}\\\textbf{ATE/RTE/ChD}} &
\shortstack{\textbf{Counter}\\\textit{starry}\\\textbf{ATE/RTE/ChD}} &
\shortstack{\textbf{Counter}\\\textit{style\_2}\\\textbf{ATE/RTE/ChD}} \\
\hline
Ours (Base) &
\shortstack{0.106/2.225/1.34} &
\shortstack{0.125/2.625/0.74} &
\shortstack{0.063/3.058/12.28} &
\shortstack{0.024/1.721/9.70} &
\shortstack{0.146/7.468/20.33} \\
Ours (Base + $\mathcal{L}_{Depth}$) &
\shortstack{0.083/1.969/1.02} &
\shortstack{0.122/1.146/0.69} &
\shortstack{0.033/1.489/9.73} &
\shortstack{0.023/1.684/8.66} &
\shortstack{0.036/1.616/12.33} \\
Ours (Base + $\mathcal{L}_{SG}$) &
\shortstack{0.008/0.247/0.94} &
\shortstack{0.009/0.747/0.62} &
\shortstack{0.013/0.742/8.97} &
\shortstack{0.019/1.664/8.41} &
\shortstack{0.022/1.414/11.33} \\
Ours (Base + $\mathcal{L}_{Depth}$ + $\mathcal{L}_{SG}$) &
\shortstack{\textbf{0.001}/\textbf{0.163}/\textbf{0.73}} &
\shortstack{\textbf{0.006}/\textbf{0.546}/\textbf{0.59}} &
\shortstack{\textbf{0.011}/\textbf{0.655}/\textbf{8.40}} &
\shortstack{\textbf{0.014}/\textbf{1.179}/\textbf{7.48}} &
\shortstack{\textbf{0.016}/\textbf{0.987}/\textbf{9.75}} \\
\hline
\end{tabular}
\end{table}

\subsection{Quantitative Analysis: Garden and Counter scenes}
The ablations over the Garden and Counter scenes still show a consistent trend: $\mathcal{L}_{SG}$ and $\mathcal{L}_{Depth}$ improve trajectory stability and reconstruction faithfulness. In Garden, adding $\mathcal{L}_{SG}$ yields a large reduction in ATE/RTE (e.g., style\_19: ATE drops from 0.106 to 0.008), indicating that directly constraining cross-view matchability improves tracking even when the base sequence is already solvable. The full model achieves the best SLAM metrics across both styles, suggesting that depth and correspondence constraints are complementary: depth stabilizes global geometry cues while correspondence regularization protects local repeatable structure.

For Counter, the static metrics show a clearer style-structure trade-off depending on the style. For style\_19 and style\_2, the full model improves DSD substantially (structure preservation), while CHD is not always minimized by the same variant (e.g., Starry has the lowest CHD with $\mathcal{L}_{Depth}$). This indicates that constraints designed for geometric usability can mildly restrict style adherence in color-histogram space, but they consistently improve SLAM-based measures. Importantly, the full model improves ATE/RTE/ChD for all Counter styles, confirming that the geometric constraints increase robustness under larger appearance shifts.

\begin{table}[t]
\centering
\footnotesize
\setlength{\tabcolsep}{2.0pt}
\renewcommand{\arraystretch}{1.15}
\caption{Ablation on \textbf{Cycle}: static metrics (lower is better). Each entry is \textbf{CHD/DSD}. Best (lowest) CHD and DSD per column are bolded.}
\label{tab:ablation_static_cycle}
\begin{tabular}{lcccc}
\hline
\textbf{Method} &
\shortstack{\textbf{style 19}\\\textbf{CHD/DSD}} &
\shortstack{\textbf{starry}\\\textbf{CHD/DSD}} &
\shortstack{\textbf{style 15}\\\textbf{CHD/DSD}} &
\shortstack{\textbf{style 41}\\\textbf{CHD/DSD}} \\
\hline
Ours (Base) &
\shortstack{\textbf{0.2495}/17.6369} &
\shortstack{\textbf{0.2218}/12.7811} &
\shortstack{\textbf{0.1875}/18.2556} &
\shortstack{0.2951/15.9439} \\
Ours (Base + $\mathcal{L}_{Depth}$) &
\shortstack{0.3246/18.2966} &
\shortstack{0.2230/13.3792} &
\shortstack{0.2017/16.1685} &
\shortstack{0.2723/14.5357} \\
Ours (Base + $\mathcal{L}_{SG}$) &
\shortstack{0.3250/16.1209} &
\shortstack{0.2394/11.6278} &
\shortstack{0.2251/15.9774} &
\shortstack{0.2664/13.9665} \\
Ours (Base + $\mathcal{L}_{Depth}$ + $\mathcal{L}_{SG}$) &
\shortstack{0.3384/\textbf{15.6645}} &
\shortstack{0.2501/\textbf{10.4830}} &
\shortstack{0.2423/\textbf{14.9774}} &
\shortstack{\textbf{0.2468}/\textbf{13.0096}} \\
\hline
\end{tabular}
\end{table}

\begin{table}[t]
\centering
\scriptsize
\setlength{\tabcolsep}{1.7pt}
\renewcommand{\arraystretch}{1.15}
\caption{Ablation on \textbf{Cycle}: DROID SLAM-based 3D consistency (lower is better). Each entry is \textbf{ATE/RTE/ChD}. Best (lowest) ATE, RTE, and ChD per column are bolded.}
\label{tab:ablation_slam_cycle}
\begin{tabular}{lcccc}
\hline
\textbf{Method} &
\shortstack{\textbf{style 19}\\\textbf{ATE/RTE/ChD}} &
\shortstack{\textbf{starry}\\\textbf{ATE/RTE/ChD}} &
\shortstack{\textbf{style 15}\\\textbf{ATE/RTE/ChD}} &
\shortstack{\textbf{style 41}\\\textbf{ATE/RTE/ChD}} \\
\hline
Ours (Base) &
\shortstack{0.288/11.986/36.00} &
\shortstack{0.347/29.497/18.21} &
\shortstack{0.111/4.997/20.29} &
\shortstack{0.139/6.961/15.40} \\
Ours (Base + $\mathcal{L}_{Depth}$) &
\shortstack{0.173/7.856/28.92} &
\shortstack{0.149/6.483/17.80} &
\shortstack{0.100/4.702/15.33} &
\shortstack{0.133/5.291/9.44} \\
Ours (Base + $\mathcal{L}_{SG}$) &
\shortstack{0.1381/5.970/14.20} &
\shortstack{0.127/6.190/13.75} &
\shortstack{0.095/3.841/12.89} &
\shortstack{0.121/4.991/8.99} \\
Ours (Base + $\mathcal{L}_{Depth}$ + $\mathcal{L}_{SG}$) &
\shortstack{\textbf{0.072}/\textbf{3.502}/\textbf{12.30}} &
\shortstack{\textbf{0.088}/\textbf{3.119}/\textbf{12.36}} &
\shortstack{\textbf{0.077}/\textbf{3.058}/\textbf{12.48}} &
\shortstack{\textbf{0.111}/\textbf{4.554}/\textbf{7.44}} \\
\hline
\end{tabular}
\end{table}

\subsection{Quantitative Analysis: Bicycle scene}
Bicycle is the hardest ablation setting in our study: it exhibits larger motion and texture variations, so SLAM depends strongly on stable correspondences across views. This is reflected in the Base variant, which yields relatively high ATE/RTE and large Chamfer distances for several styles. Adding $\mathcal{L}_{Depth}$ improves the 3D metrics in most cases (e.g., style\_19 ATE 0.288 $\rightarrow$ 0.173), indicating that preserving monocular depth cues already mitigates some volumetric drift. However, the most significant gains occur when adding $\mathcal{L}_{SG}$: the correspondence constraint sharply reduces Chamfer distance (e.g., style\_19 36.00 $\rightarrow$ 14.20) and improves both ATE and RTE, confirming that explicitly regularizing descriptor stability under confident cross-view matches targets the primary failure mode for multi-view stylization.

The full model provides the best SLAM metrics for all Bicycle styles, demonstrating that $\mathcal{L}_{Depth}$ and $\mathcal{L}_{SG}$ are complementary rather than redundant. In the static metrics, CHD often increases when constraints are added, while DSD decreases substantially across styles. This pattern is consistent with our objective: geometric usability for reconstruction is improved by enforcing structure-preserving constraints, even if it slightly restricts aggressive style shifts measured purely in global color-histogram space.

\subsection{Qualitative Analysis: Train Scene}
\begin{figure}[p]
    \centering
    \includegraphics[width=\textwidth]{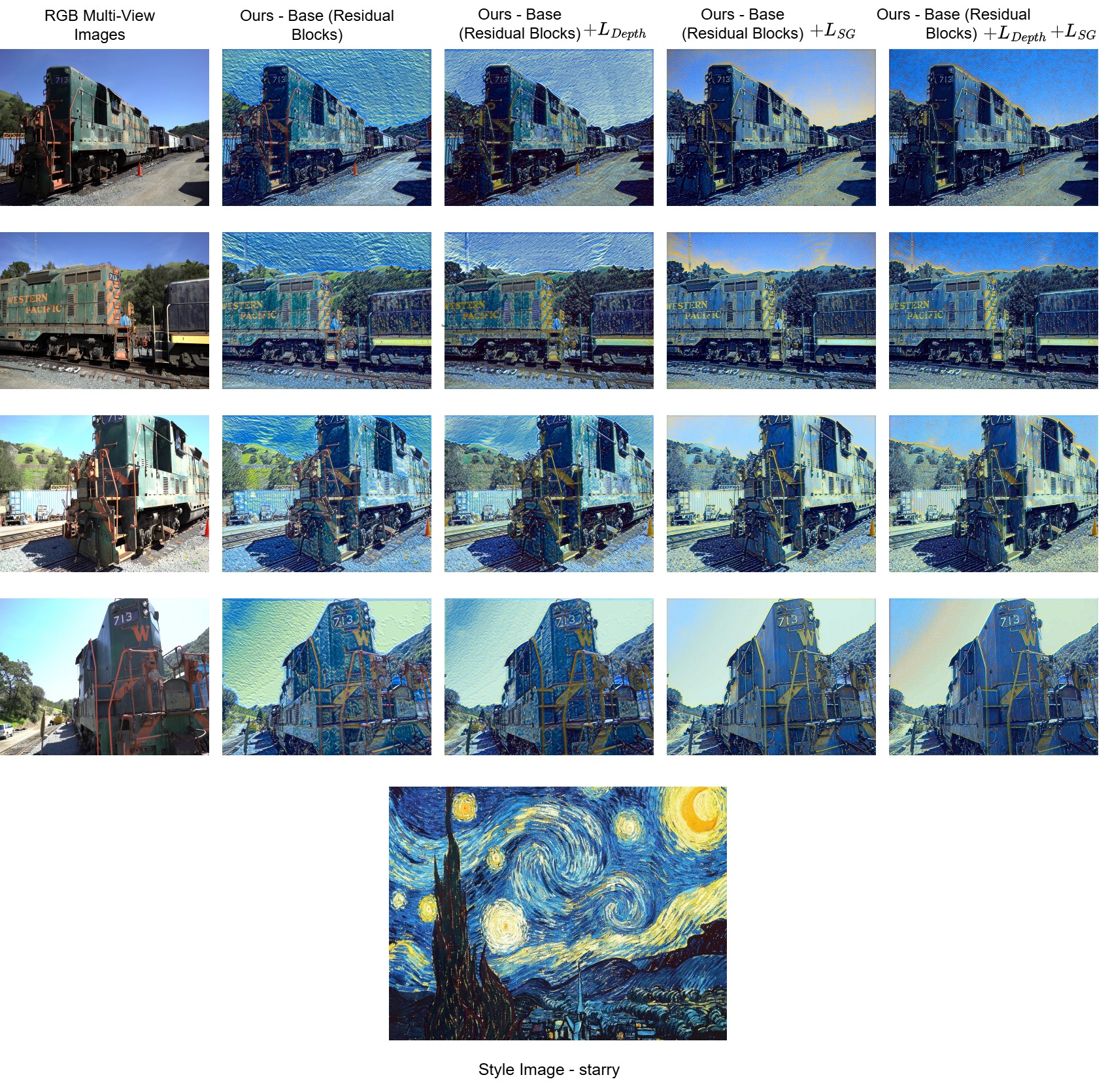}
    \caption{Qualitative ablation on \textbf{Train} with \textbf{starry} style. From left to right:
    input RGB, Base (AdaIN), Base+$\mathcal{L}_{Depth}$, Base+$\mathcal{L}_{SG}$, Base+$\mathcal{L}_{Depth}+\mathcal{L}_{SG}$.}
    \label{fig:ablation_train_starry}
\end{figure}

\begin{figure}[p]
    \centering
    \includegraphics[width=\textwidth]{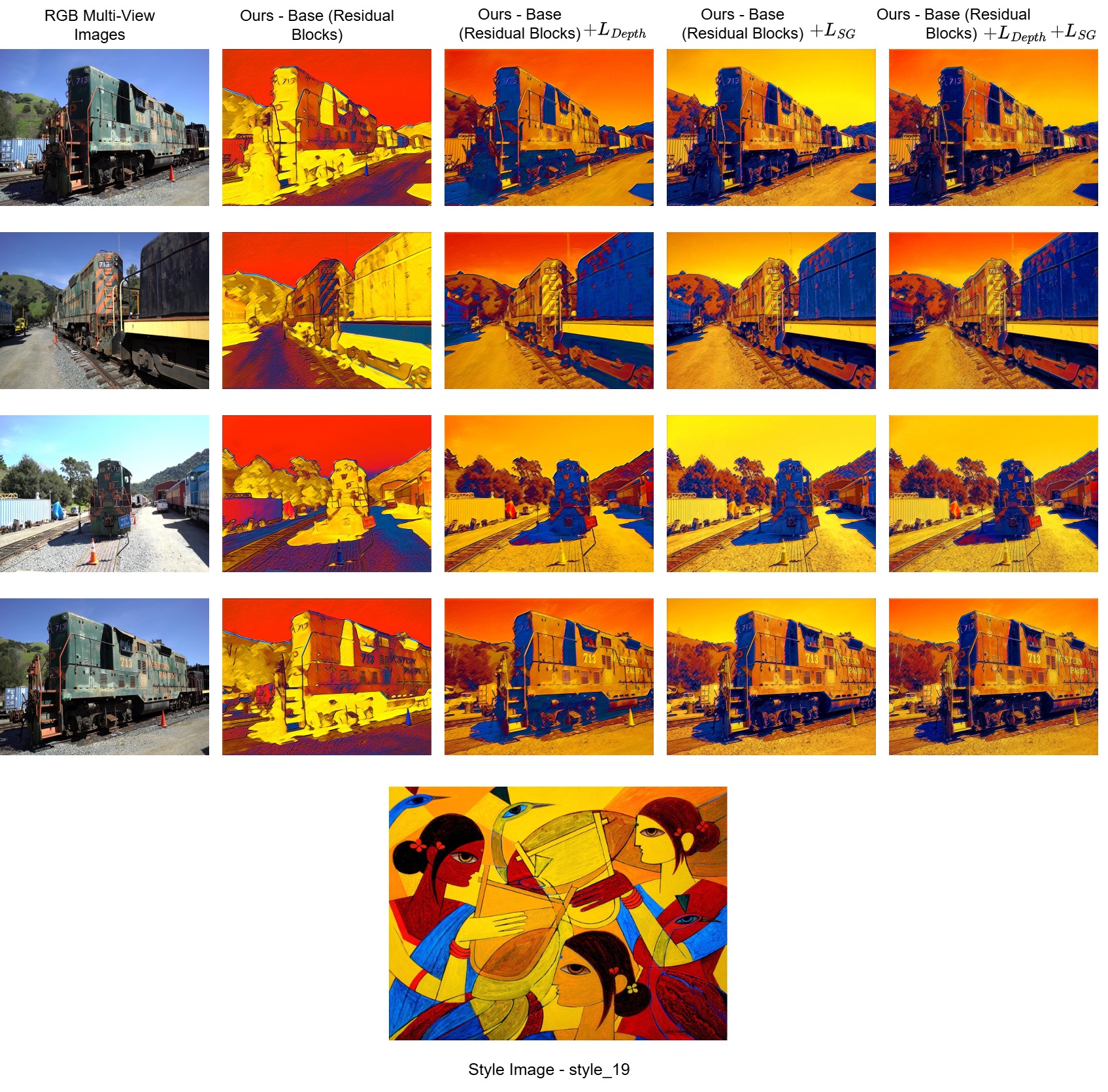}
    \caption{Qualitative ablation on \textbf{Train} with \textbf{style\_19}. From left to right:
    input RGB, Base (AdaIN), Base+$\mathcal{L}_{Depth}$, Base+$\mathcal{L}_{SG}$, Base+$\mathcal{L}_{Depth}+\mathcal{L}_{SG}$.}
    \label{fig:ablation_train_19}
\end{figure}

Figures~\ref{fig:ablation_train_starry} and \ref{fig:ablation_train_19} compare (from left to right) the input multi-view RGB frames, our baseline residual stylizer trained only with AdaIN losses (\textbf{Base}), and three progressively constrained variants: \textbf{Base+$\mathcal{L}_{Depth}$}, \textbf{Base+$\mathcal{L}_{SG}$}, and \textbf{Base+$\mathcal{L}_{Depth}+\mathcal{L}_{SG}$}. Across both styles, the same qualitative trend emerges: \emph{adding constraints reduces view-dependent stylization artifacts and stabilizes 3D-relevant structure, but can slightly attenuate aggressive stylistic distortions}.

\paragraph{Train + starry.}
In Fig.~\ref{fig:ablation_train_starry}, the \textbf{Base} model produces strong painterly texture and high-frequency brush patterns, but these patterns are not equally repeatable across viewpoints: thin edges (e.g., train contours, rails, and window boundaries) exhibit small shifts and local “texture swimming” across views. Introducing \textbf{$\mathcal{L}{Depth}$} suppresses some of the most unstable appearance changes by discouraging stylization that perturbs monocular depth cues (contrast transitions, shading boundaries). This typically makes large planar regions and major silhouettes more stable, but also slightly reduces stylistic strength in regions where depth relies heavily on appearance cues (sky/ground separation and distant background). Adding \textbf{$\mathcal{L}_{SG}$} yields a more direct improvement in multi-view repeatability: edge-like structures and local patterns that SuperPoint consistently detects become more consistent across views, reducing correspondence-breaking hallucinations. The combined \textbf{Base+$\mathcal{L}_{Depth}+\mathcal{L}_{SG}$} model provides the most reliable compromise: the overall Starry appearance is maintained, while the outlines and mid-frequency structures stay coherent across views.

\paragraph{Train + style\_19.}
Fig.~\ref{fig:ablation_train_19} shows a similar but more pronounced trade-off. The \textbf{Base} stylizer strongly pushes color and saturation, and introduces stylized edges that can drift slightly between views. With \textbf{$\mathcal{L}_{Depth}$}, the model becomes more conservative with global appearance warps that alter depth predictions; the result is often a more stable global layout but with reduced stylistic exaggeration in certain regions. With \textbf{$\mathcal{L}_{SG}$}, repeated structures (train face, rails, high-contrast boundaries) become more consistently stylized across views, producing fewer viewpoint-dependent discontinuities. The \textbf{combined} model yields the most stable geometry-preserving stylization: the scene remains clearly stylized, but the stylization is better “locked” to the underlying 3D structure.

\subsection{Qualitative Analysis: Truck Scene}
\begin{figure}[p]
    \centering
    \includegraphics[width=\textwidth]{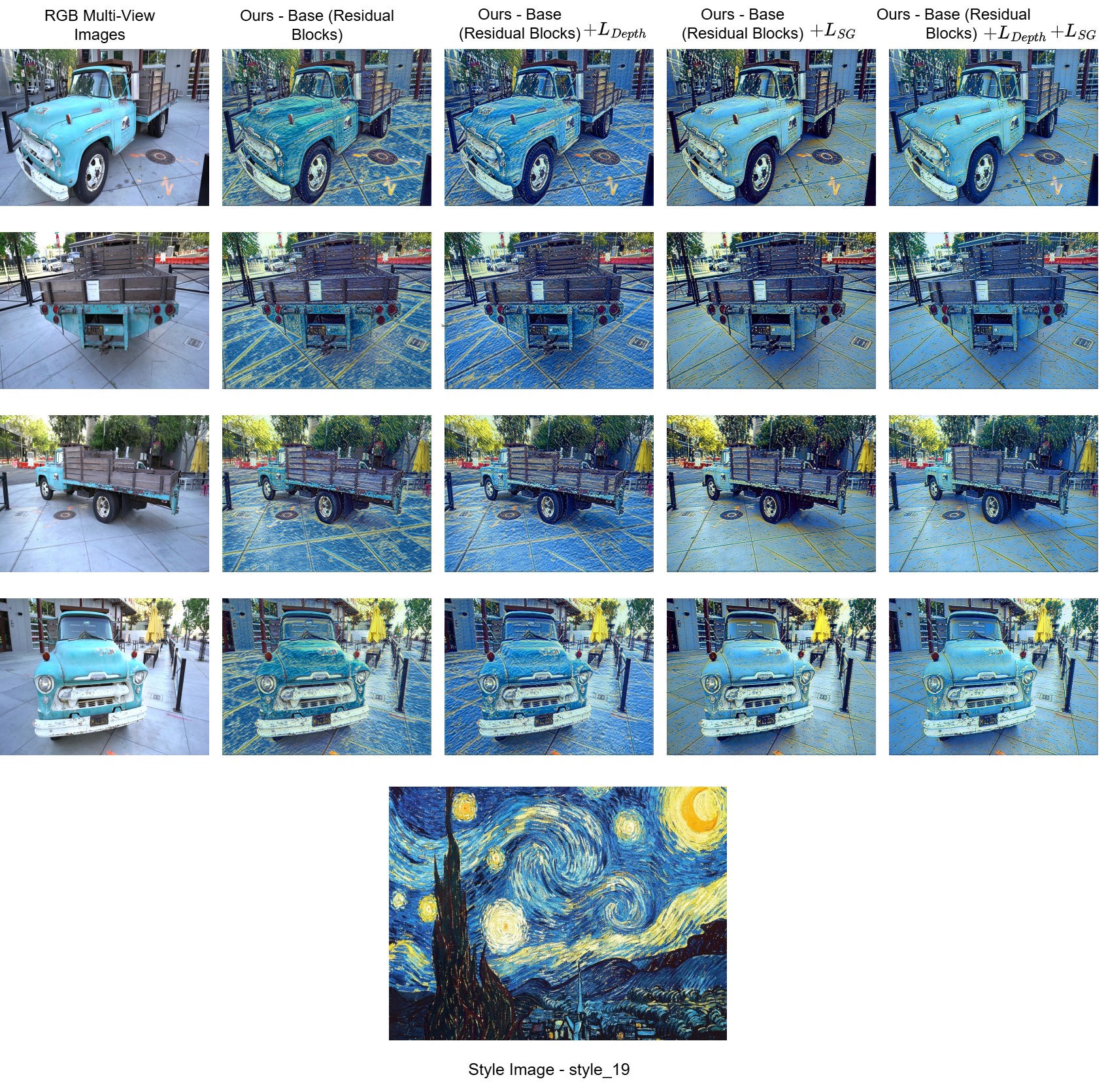}
    \caption{Qualitative ablation on \textbf{Truck} with \textbf{starry} style. From left to right:
    input RGB, Base (AdaIN), Base+$\mathcal{L}_{Depth}$, Base+$\mathcal{L}_{SG}$, Base+$\mathcal{L}_{Depth}+\mathcal{L}_{SG}$.}
    \label{fig:ablation_truck_starry}
\end{figure}

\begin{figure}[p]
    \centering
    \includegraphics[width=\textwidth]{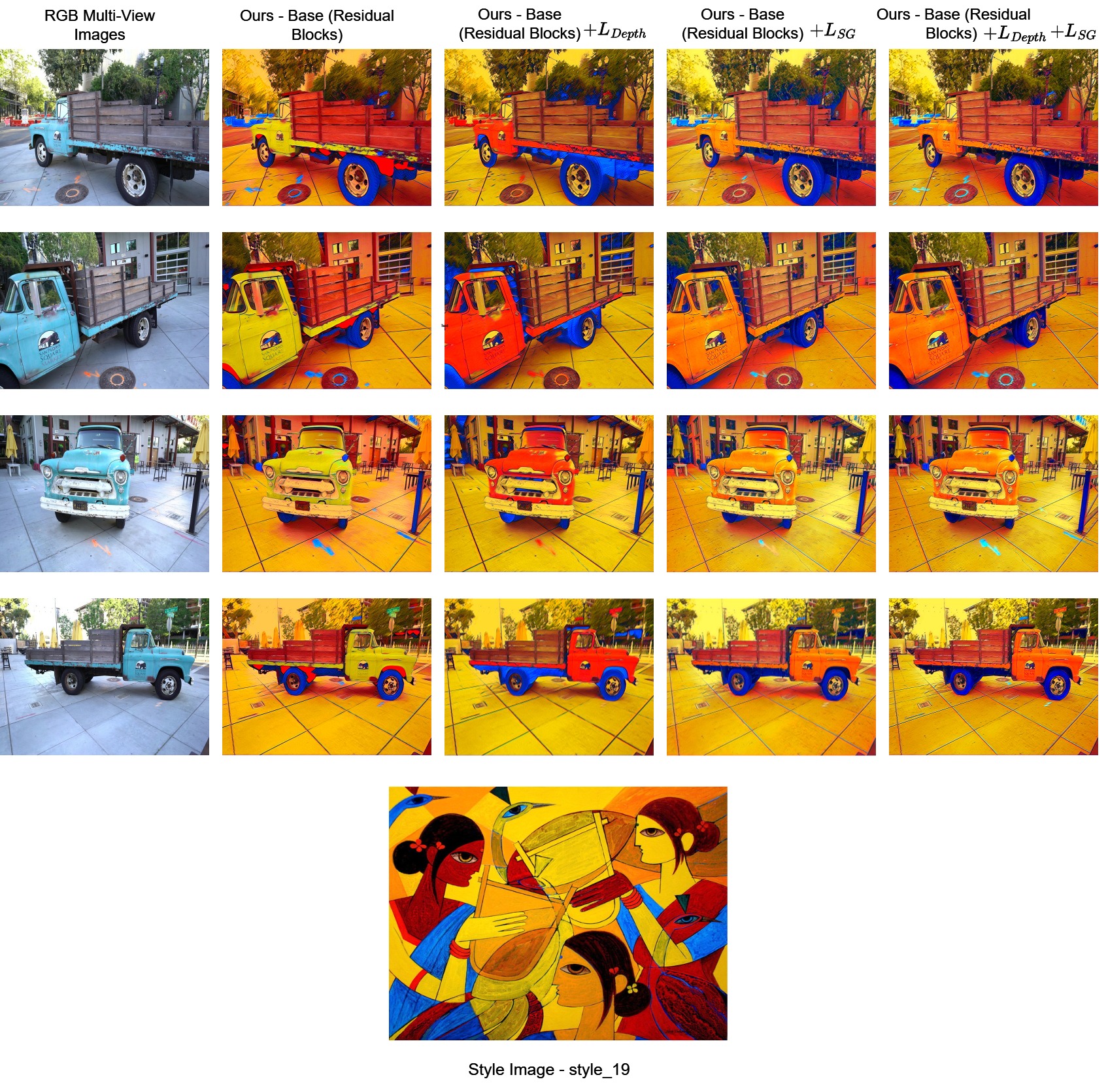}
    \caption{Qualitative ablation on \textbf{Truck} with \textbf{style\_19}. From left to right:
    input RGB, Base (AdaIN), Base+$\mathcal{L}_{Depth}$, Base+$\mathcal{L}_{SG}$, Base+$\mathcal{L}_{Depth}+\mathcal{L}_{SG}$.}
    \label{fig:ablation_truck_19}
\end{figure}

Figures~\ref{fig:ablation_truck_starry} and \ref{fig:ablation_truck_19} repeat the same ablation for the Truck scene. This scene highlights a key practical property of our formulation: \textbf{$\mathcal{L}_{SG}$} stabilizes \emph{repeatable local structure}, while \textbf{$\mathcal{L}_{Depth}$} stabilizes \emph{global volumetric consistency}; the two constraints are complementary.

\paragraph{Truck + starry.}
In Fig.~\ref{fig:ablation_truck_starry}, \textbf{Base} yields strong stylization, but fine structures (edges around the truck body, grill contours, and background boundaries) can fluctuate across views. With \textbf{$\mathcal{L}_{Depth}$}, large-scale structure becomes more stable, but the method can become overly conservative (reducing stylistic deformation) or exhibit local color instabilities if depth gradients dominate. With \textbf{$\mathcal{L}_{SG}$}, view-to-view repeatability improves notably: stylized edges adhere more consistently to the same physical boundaries across viewpoints. The \textbf{combined} variant produces the most robust results preserving the Starry “texture essence” while keeping viewpoint consistent placement of mid range frequency details.

\paragraph{Truck + style\_19.}
Fig.~\ref{fig:ablation_truck_19} illustrates that \textbf{Base} tends to over-amplify global palette shifts (strong reds/yellows) and can introduce stylization that is not perfectly consistent across frames. \textbf{$\mathcal{L}_{Depth}$} reduces some of these unstable appearance changes by discouraging depth-inconsistent shading artifacts. \textbf{$\mathcal{L}_{SG}$} further improves stability of edges and repeated local patterns across views. The \textbf{combined} model again offers the best balance: it remains clearly stylized while exhibiting the most coherent geometry-aligned appearance across all shown viewpoints.

\clearpage

\clearpage

\clearpage

\clearpage

\subsection{Qualitative Analysis: Garden and Counter scenes}
\begin{figure}[p]
    \centering
    \includegraphics[width=\textwidth]{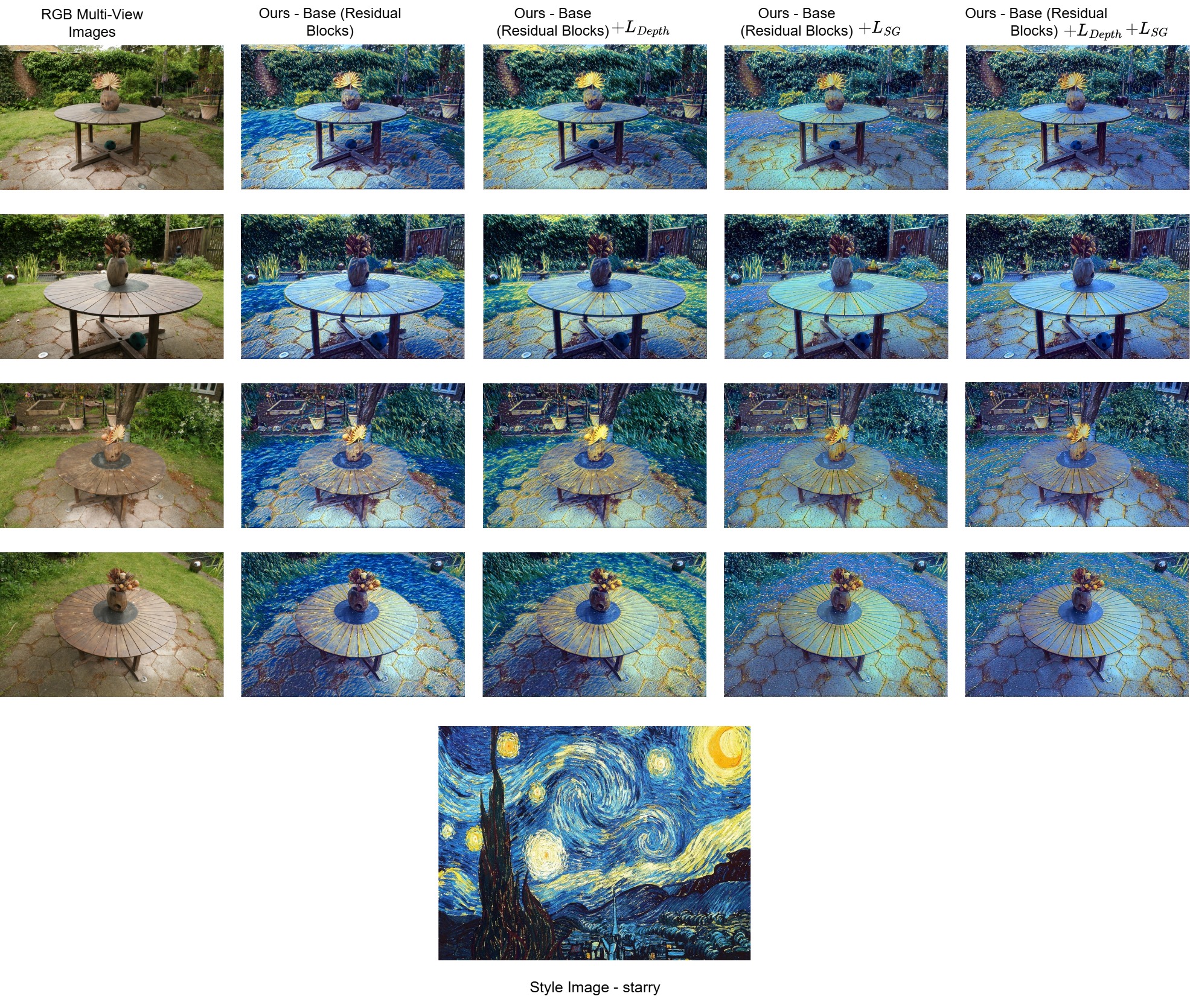}
    \caption{Qualitative ablation on \textbf{Garden} with style \textbf{starry} Columns: RGB input, Base, Base+$\mathcal{L}_{Depth}$, Base+$\mathcal{L}_{SG}$, Base+$\mathcal{L}_{Depth}$+$\mathcal{L}_{SG}$.}
    \label{fig:ablation_garden_starry}
\end{figure}

\begin{figure}[p]
    \centering
    \includegraphics[width=\textwidth]{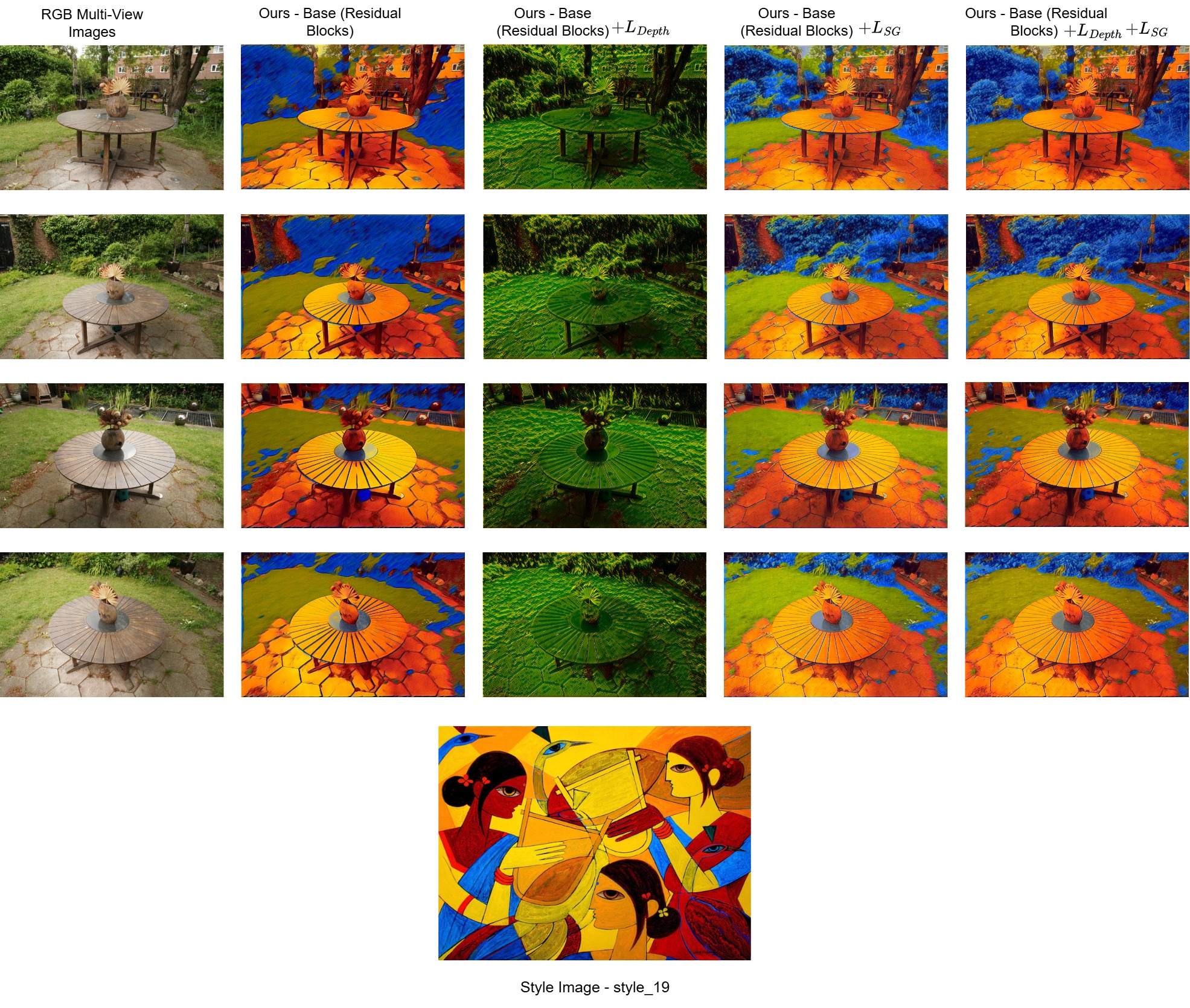}
    \caption{Qualitative ablation on \textbf{Garden} with \textbf{style\_19} Columns: RGB input, Base, Base+$\mathcal{L}_{Depth}$, Base+$\mathcal{L}_{SG}$, Base+$\mathcal{L}_{Depth}$+$\mathcal{L}_{SG}$.}
    \label{fig:ablation_garden19}
\end{figure}

\begin{figure}[p]
    \centering
    \includegraphics[width=\textwidth]{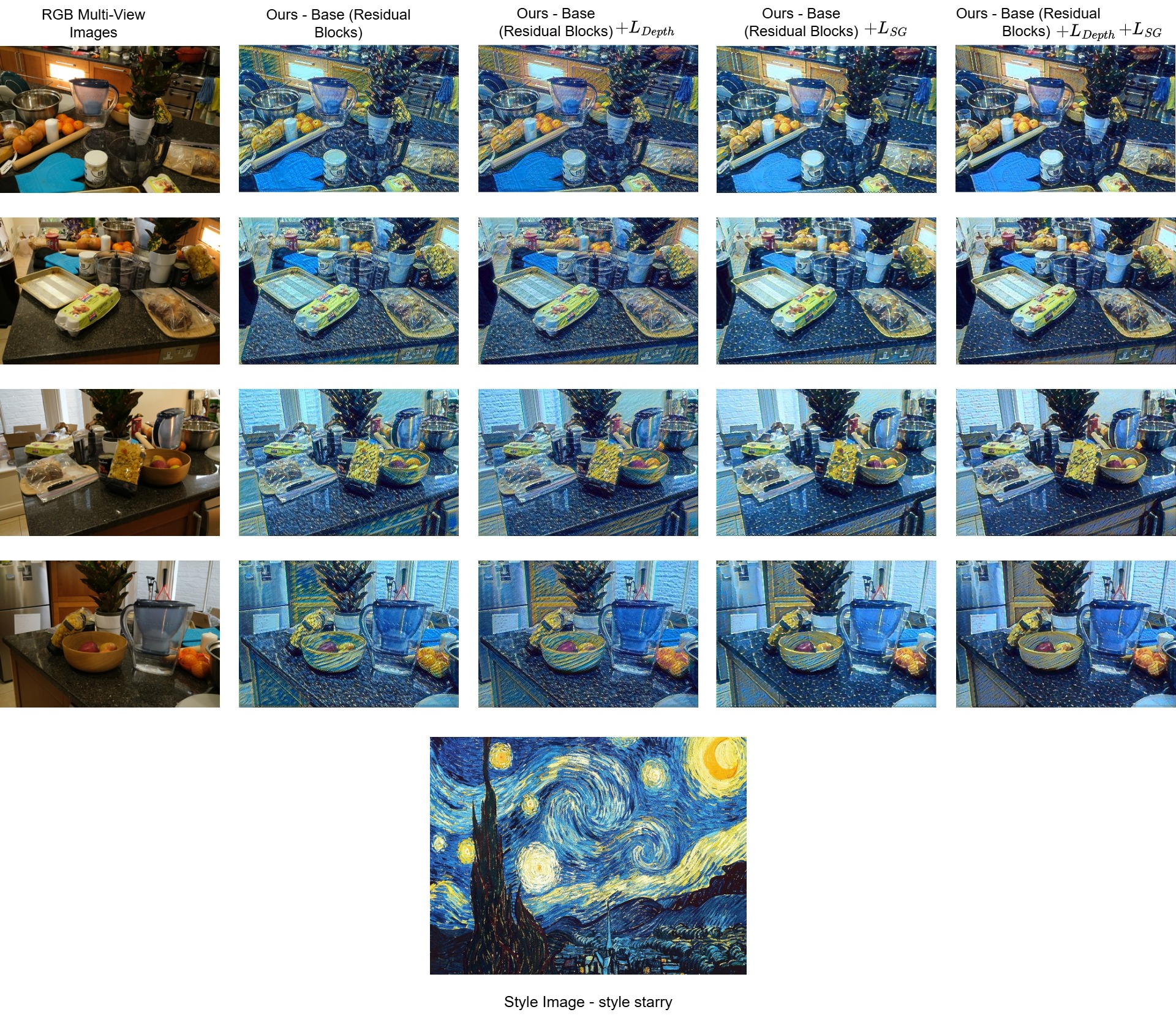}
    \caption{Qualitative ablation on \textbf{Counter} with style \textbf{starry} Columns: RGB input, Base, Base+$\mathcal{L}_{Depth}$, Base+$\mathcal{L}_{SG}$, Base+$\mathcal{L}_{Depth}$+$\mathcal{L}_{SG}$.}
    \label{fig:ablation_counter_starry}
\end{figure}

\begin{figure}[p]
    \centering
    \includegraphics[width=\textwidth]{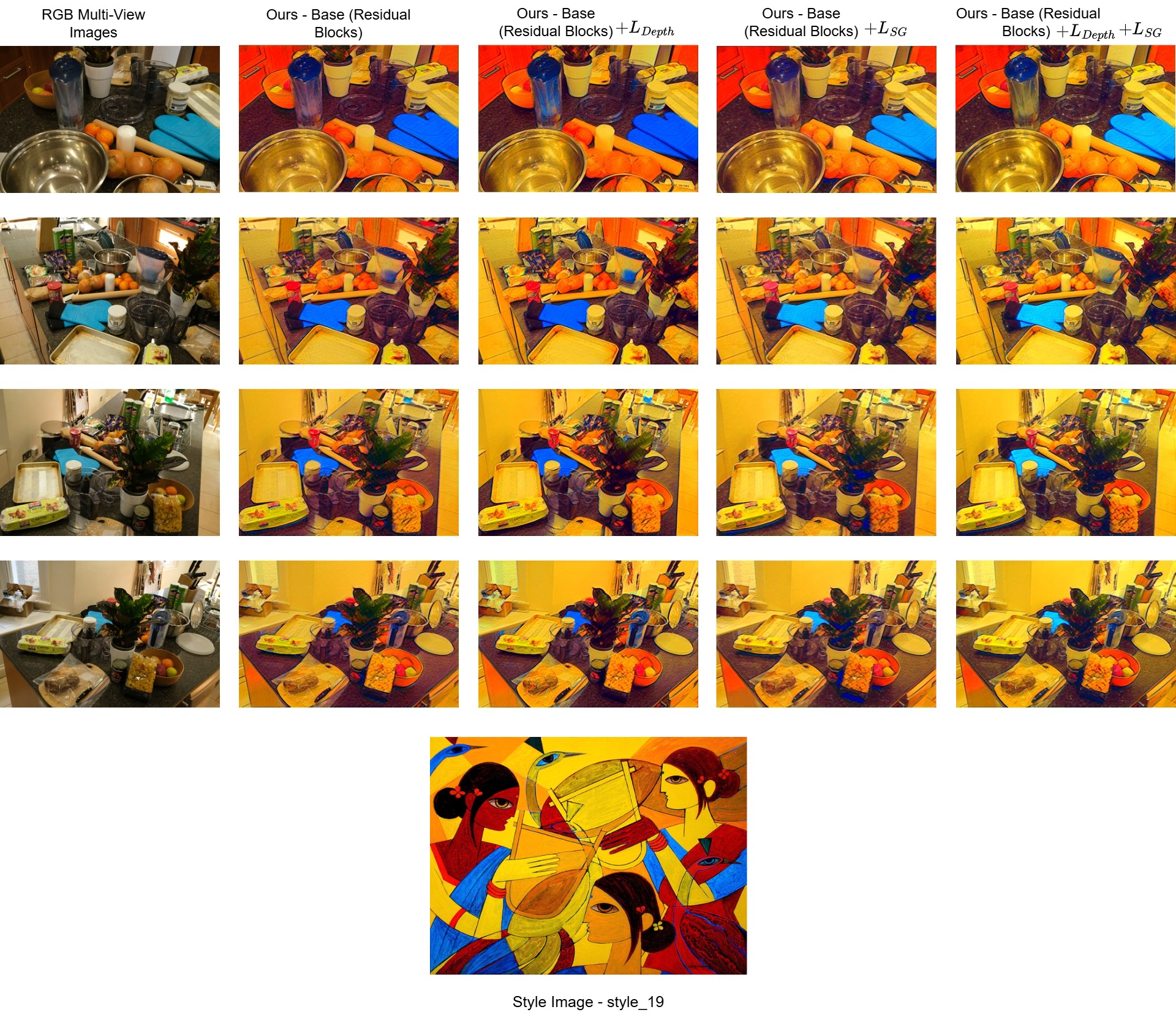}
    \caption{Qualitative ablation on \textbf{Counter} with \textbf{style\_19} Columns: RGB input, Base, Base+$\mathcal{L}_{Depth}$, Base+$\mathcal{L}_{SG}$, Base+$\mathcal{L}_{Depth}$+$\mathcal{L}_{SG}$.}
    \label{fig:ablation_counter19}
\end{figure}

\begin{figure}[p]
    \centering
    \includegraphics[width=\textwidth]{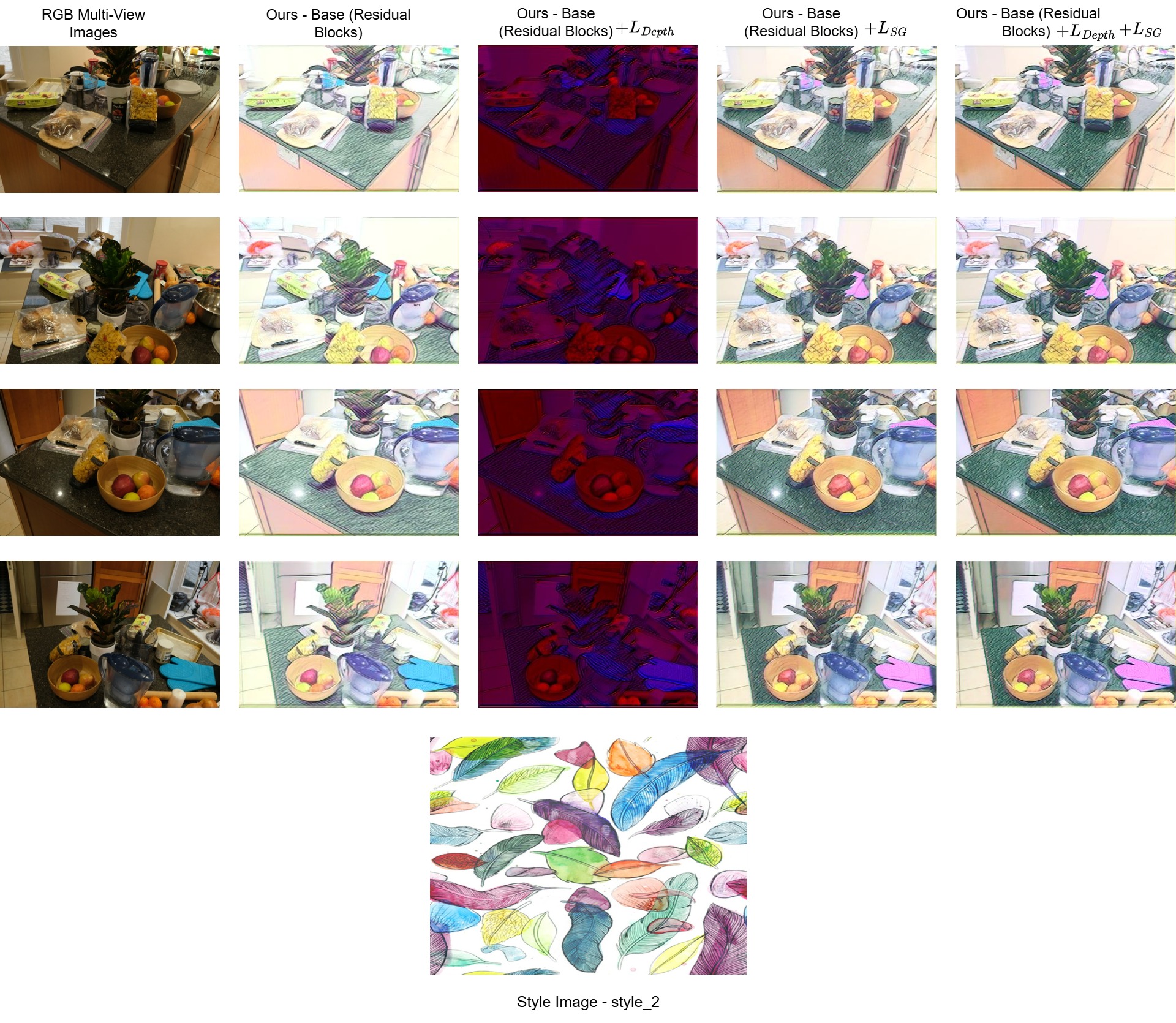}
    \caption{Qualitative ablation on \textbf{Counter} with \textbf{style\_2} Columns: RGB input, Base, Base+$\mathcal{L}_{Depth}$, Base+$\mathcal{L}_{SG}$, Base+$\mathcal{L}_{Depth}$+$\mathcal{L}_{SG}$.}
    \label{fig:ablation_counter_style2}
\end{figure}

Figures~\ref{fig:ablation_garden_starry}--\ref{fig:ablation_counter_style2} visualize the qualitative effect of progressively adding the two geometry-centric regularizers to the base stylizer: the monocular depth consistency loss ($\mathcal{L}_{Depth}$) and the cross-view correspondence loss ($\mathcal{L}_{SG}$). Across both Garden and Counter, we observe a consistent trend: \textit{adding geometric constraints increases multi-view stability and structural faithfulness, at the cost of slightly weaker stylization strength}. This trade-off aligns with the quantitative behavior reported earlier (CHD tends to rise modestly while DSD and SLAM errors improve), and is expected because both regularizers explicitly penalize view-dependent appearance changes that are common in aggressive texture transfer.

\paragraph{Garden (starry).}
In Fig.~\ref{fig:ablation_garden_starry}, the \textbf{Base} stylizer produces visually strong painterly strokes, but some regions exhibit view-dependent high-frequency patterns (notably on the stone pavement and foliage), which can subtly shift between frames. Adding \textbf{$\mathcal{L}_{Depth}$} dampens these instabilities by discouraging stylization changes that alter the depth network’s response; this improves the coherence of large planar surfaces (the patio stones) and reduces ``floating'' texture behavior. Adding \textbf{$\mathcal{L}_{SG}$} further anchors the stylization to repeatable cross-view keypoint correspondences, making edges of the table and flowerpot more consistent across viewpoints. Finally, \textbf{$\mathcal{L}_{Depth}+\mathcal{L}_{SG}$} yields the most stable configuration: strokes remain stylistic, but they better respect object boundaries and maintain viewpoint-consistent placement.

\paragraph{Garden (style\_19).}
Fig.~\ref{fig:ablation_garden19} highlights a key failure mode of using depth alone in a test-time stylization loop. With \textbf{$\mathcal{L}_{Depth}$ only}, the output collapses into an unnatural, nearly monochromatic \emph{green-dominant} appearance that looks like a ``garbage'' color solution rather than a meaningful rendition of the target style. This behavior is characteristic of \textbf{OOD depth regularization}: MiDaS/DPT is trained on natural images, and strongly stylized distributions (especially those with extreme palette shifts) can lie far outside its training domain. When the depth term is applied without an additional cross-view anchor, optimization can converge to a \textbf{degenerate local minimum} where global color/contrast is distorted in a way that stabilizes the depth network response, even if it destroys the intended style palette. In contrast, \textbf{$\mathcal{L}_{SG}$} provides a complementary constraint that forces patch-level correspondences to remain geometrically meaningful; this prevents the optimizer from ``escaping'' into trivial color casts. As a result, \textbf{$\mathcal{L}_{SG}$ only} already restores a more plausible stylization, and the \textbf{combined} objective achieves the best balance between geometry and recognizable stylistic coloration.

\paragraph{Counter (starry).}
The Counter scene (Fig.~\ref{fig:ablation_counter_starry}) is dominated by clutter, specular highlights, and many small objects (bowls, containers, utensils), which are challenging for multi-view stylization because fine textures are easily hallucinated in a view-dependent way. The \textbf{Base} stylizer tends to introduce strong strokes and micro-texture on the countertop fabric and object surfaces; these patterns can drift across frames and reduce matchability. Adding \textbf{$\mathcal{L}_{Depth}$} improves \emph{global surface consistency} (e.g., planar regions on the counter and cabinet faces), while \textbf{$\mathcal{L}_{SG}$} improves \emph{edge and object-boundary repeatability} (e.g., container rims, fruit edges, and strong silhouette contours). The \textbf{combined} model yields the most stable depiction: stylization remains present, but textures are less likely to ``swim'' across viewpoints, which is precisely the behavior required for downstream tracking and reconstruction.

\paragraph{Counter (style\_2 and style\_19).}
Figures~\ref{fig:ablation_counter_style2} and \ref{fig:ablation_counter19} further illustrate that \textbf{$\mathcal{L}_{Depth}$ alone can become brittle under large style-domain shifts}. In Style\_2 (Fig.~\ref{fig:ablation_counter_style2}), the depth-only variant collapses to a dark purple/magenta cast with suppressed style structure, again indicating a \textbf{shortcut} where the optimizer prioritizes a depth-network-consistent appearance over stylization fidelity. This instability is substantially reduced once \textbf{$\mathcal{L}_{SG}$} is introduced: correspondences act as a geometric ``backbone'' that keeps the stylizer aligned to repeatable scene structure, limiting the space of solutions that the depth term can exploit. The \textbf{combined} objective consistently produces the most reconstruction-friendly appearance while retaining a recognizable imprint of the target style.

\clearpage

\clearpage

\clearpage

\clearpage

\clearpage

\subsection{Qualitative Analysis: Bicycle scene}
\begin{figure}[p]
    \centering
    \includegraphics[width=\textwidth]{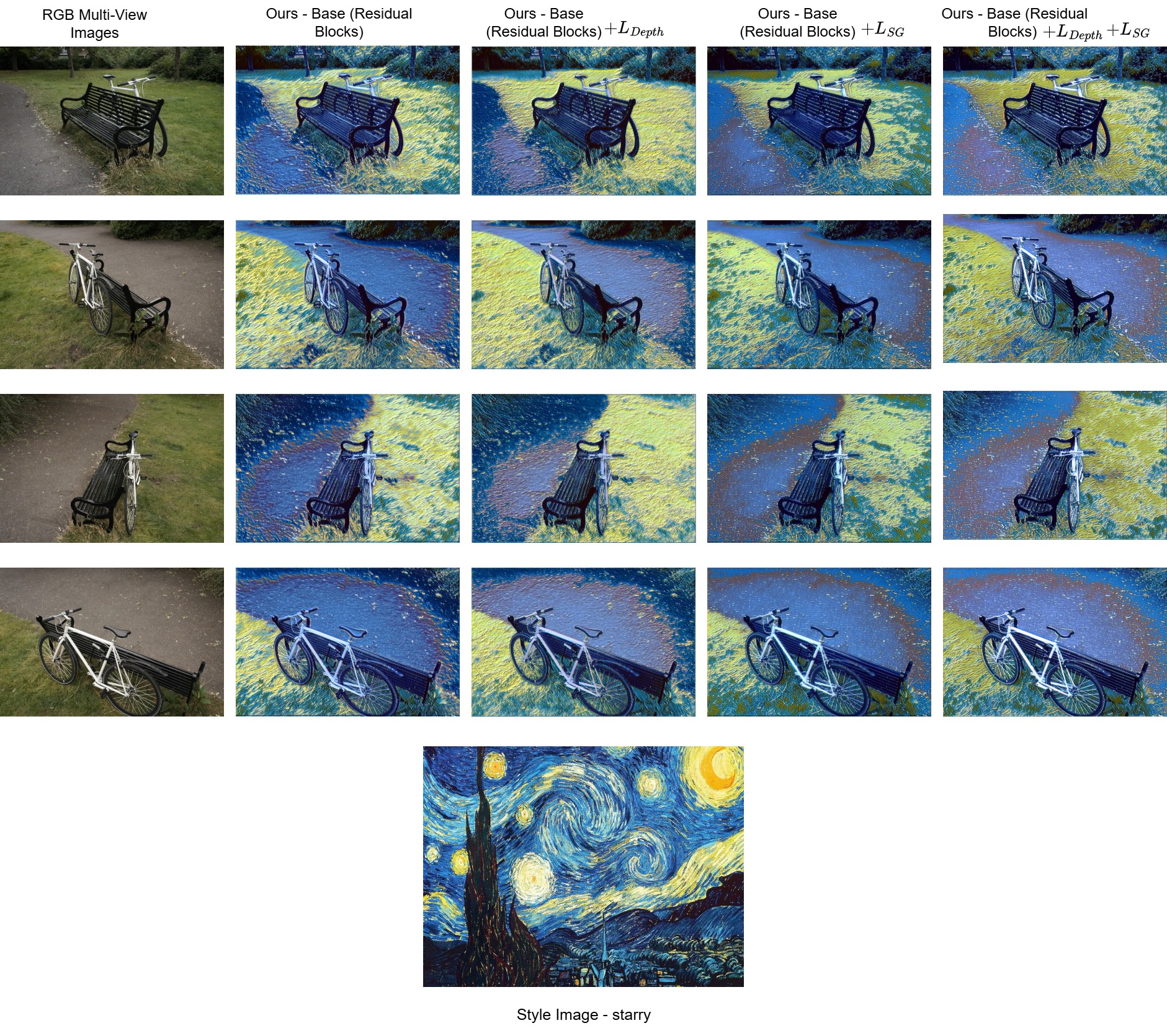}
    \caption{Qualitative ablation on \textbf{Bicycle} with style \textbf{starry} Columns: RGB input, Base, Base+$\mathcal{L}_{Depth}$, Base+$\mathcal{L}_{SG}$, Base+$\mathcal{L}_{Depth}$+$\mathcal{L}_{SG}$.}
    \label{fig:ablation_cycle_starry}
\end{figure}

\begin{figure}[p]
    \centering
    \includegraphics[width=\textwidth]{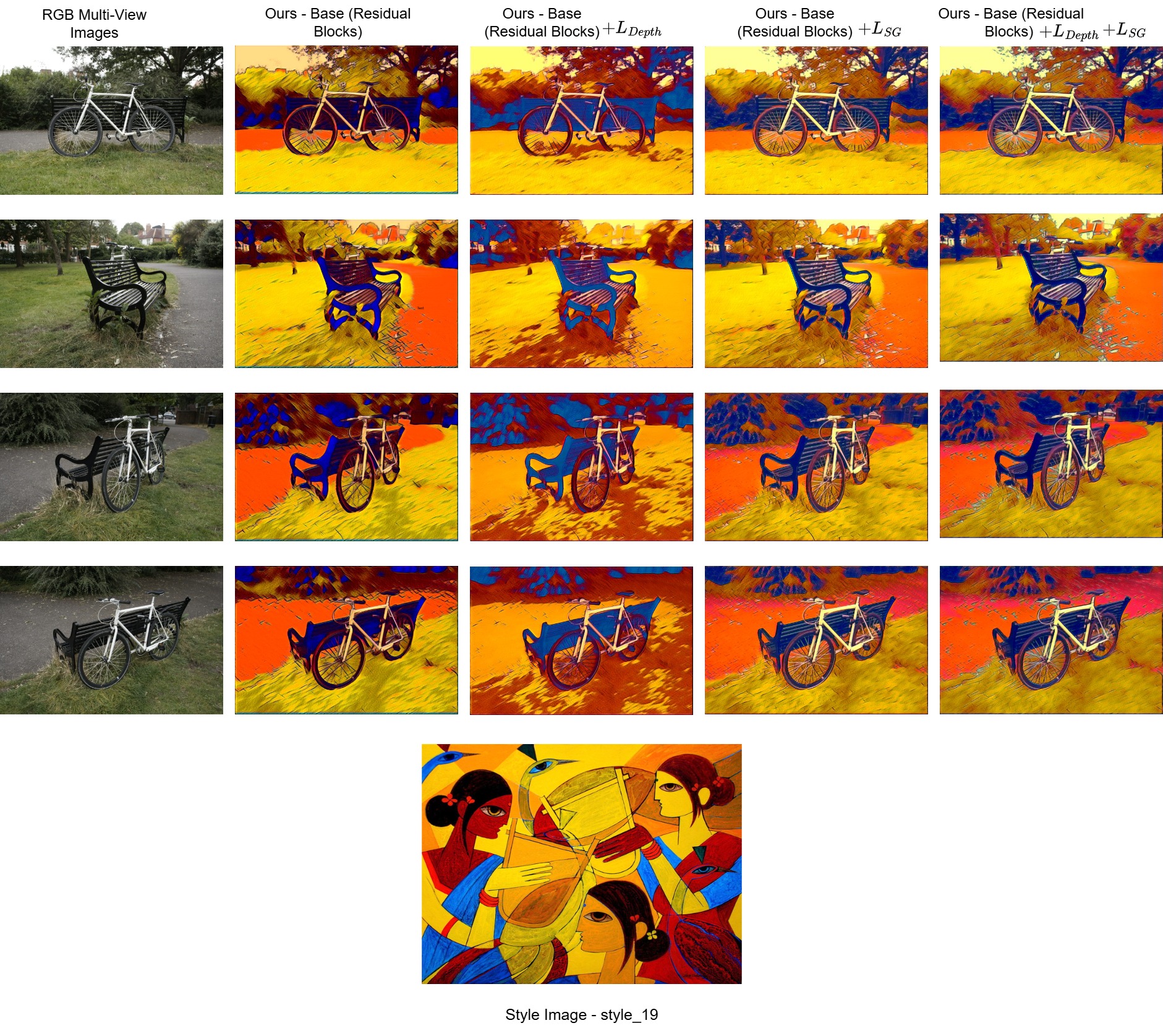}
    \caption{Qualitative ablation on \textbf{Bicycle} with style \textbf{style\_19} Columns: RGB input, Base, Base+$\mathcal{L}_{Depth}$, Base+$\mathcal{L}_{SG}$, Base+$\mathcal{L}_{Depth}$+$\mathcal{L}_{SG}$.}
    \label{fig:ablation_cycle19}
\end{figure}

\begin{figure}[p]
    \centering
    \includegraphics[width=\textwidth]{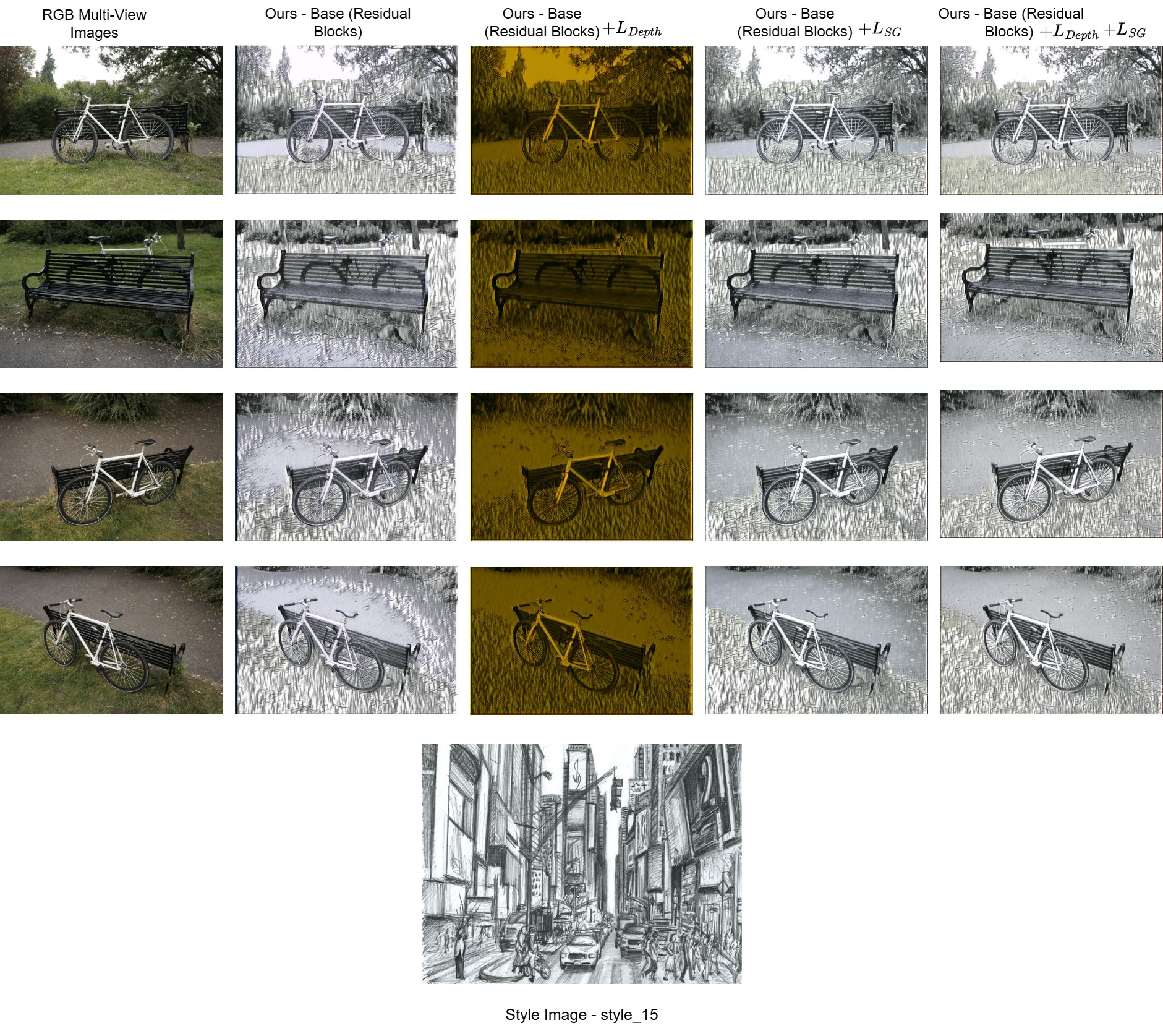}
    \caption{Qualitative ablation on \textbf{Bicycle} with style \textbf{style\_15} Columns: RGB input, Base, Base+$\mathcal{L}_{Depth}$, Base+$\mathcal{L}_{SG}$, Base+$\mathcal{L}_{Depth}$+$\mathcal{L}_{SG}$.}
    \label{fig:ablation_cycle15}
\end{figure}

\begin{figure}[p]
    \centering
    \includegraphics[width=\textwidth]{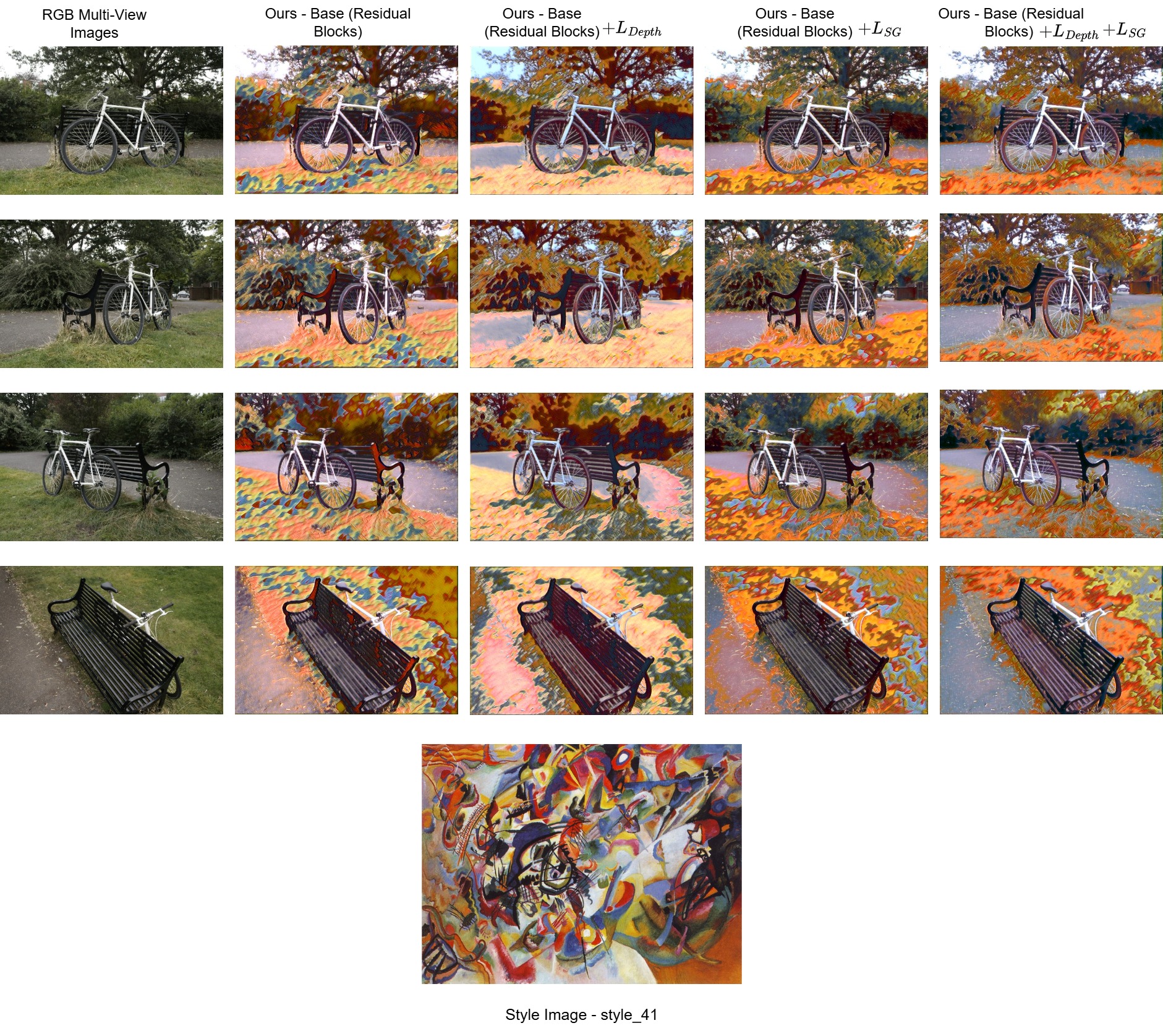}
    \caption{Qualitative ablation on \textbf{Bicycle} with style \textbf{style\_41} Columns: RGB input, Base, Base+$\mathcal{L}_{Depth}$, Base+$\mathcal{L}_{SG}$, Base+$\mathcal{L}_{Depth}$+$\mathcal{L}_{SG}$.}
    \label{fig:ablation_cycle41}
\end{figure}

Figure~\ref{fig:ablation_cycle_starry}--\ref{fig:ablation_cycle41} shows the Bicycle scene, which contains many strong geometric cues (bike frame, spokes, bench slats) that are highly relevant for correspondence-based constraints. Consistent with our design, $\mathcal{L}_{SG}$ is particularly effective here: it preserves the repeatability of thin structures and reduces cross-view jitter on elongated edges, while $\mathcal{L}_{Depth}$ stabilizes large regions such as the ground plane and background.

\paragraph{Bicycle (starry).}
In Fig.~\ref{fig:ablation_cycle_starry}, all variants preserve the overall composition, but differences appear in \emph{stroke stability} on the ground and in the crispness of the bicycle silhouette. The \textbf{Base} output is highly stylized yet slightly more ``textured'' in a way that can drift between views. Adding \textbf{$\mathcal{L}_{Depth}$} makes the ground appearance more coherent and reduces over-texturing. Adding \textbf{$\mathcal{L}_{SG}$} strengthens thin-structure consistency (bike frame, bench outline). The \textbf{combined} version best suppresses view-dependent brush patterns while keeping the Starry palette.

\paragraph{Bicycle (style\_19).}
Fig.~\ref{fig:ablation_cycle19} demonstrates a clearer style--geometry trade-off. The \textbf{Base} variant applies strong, high-contrast style fields but can slightly deform thin structures (e.g., bike spokes and bench slats appear less repeatable). With \textbf{$\mathcal{L}_{Depth}$}, large regions become more stable but the style becomes less aggressive. With \textbf{$\mathcal{L}_{SG}$}, silhouettes and thin structures align more consistently across viewpoints. The \textbf{combined} model retains a recognizable Style\_19 color scheme while producing the most structurally reliable images for downstream matching.

\paragraph{Bicycle (style\_15: sketch-like).}
In Fig.~\ref{fig:ablation_cycle15}, the target style is predominantly line-based and low-frequency in color. Here, \textbf{$\mathcal{L}_{Depth}$ only} again shows vulnerability to palette collapse (sepia/olive cast) because the depth constraint dominates and the depth network response can be stabilized by global color shifts that do not correspond to the intended sketch appearance. In contrast, \textbf{$\mathcal{L}_{SG}$} maintains clean structural lines on the bike and bench, and the \textbf{combined} objective yields the most faithful sketch-like rendering while remaining geometrically stable.

\paragraph{Bicycle (style\_41).}
Fig.~\ref{fig:ablation_cycle41} shows that for abstract, high-color styles, the \textbf{Base} model tends to maximize stylization strength but introduce viewpoint-dependent color patches along boundaries. Depth regularization reduces such patchiness by discouraging depth-inconsistent appearance shifts, while SuperGlue regularization improves boundary locking (bike outline and bench geometry). The \textbf{combined} model provides the cleanest compromise: stable geometry with an abstract color field that remains consistent across views.

\subsection{Why \textbf{Base+$\mathcal{L}_{Depth}$} can fall into “garbage” color regimes.}
In some runs of scenes, the depth-only penalized model can converge to visually implausible or “stuck” colorizations. This behavior is consistent with the fact that the depth term is computed through a \emph{frozen} monocular depth network, which introduces two failure modes during optimization:
\begin{itemize}
\item \textbf{Out-of-Domain(OOD) sensitivity of monocular depth:} stylized images are far from the depth network’s training distribution. Gradients from the depth model can therefore push the stylizer toward unnatural color/contrast configurations that nonetheless reproduce similar depth responses.
\item \textbf{Underdetermined constraint:} matching depth alone does not uniquely constrain appearance. Many different RGB renderings can yield similar normalized depth maps, so the optimizer may find a low-loss but visually degenerate solution (e.g., strong global hue shifts, saturation collapse, or flat low-frequency color fields).
\item \textbf{Lack of cross-view anchoring:} without $\mathcal{L}_{SG}$, nothing explicitly enforces that stylization remains \emph{correspondence-consistent} across views. As a result, the model can satisfy depth while drifting in per-view texture placement or global coloration.
\end{itemize}
Empirically, adding \textbf{$\mathcal{L}_{SG}$} significantly reduces this degeneracy because correspondence matching penalizes solutions that destroy repeatable keypoint structure; this effectively anchors stylization to stable, trackable image evidence.

\clearpage

\clearpage

\clearpage

\clearpage

\subsection{Real-World Style Images for Domain Translation}
Beyond artistic paintings, we also test our framework with \emph{real photographs} as style references (e.g., autumn, rainy, snow) and transfer them onto multi-view scenes (Playground, Francis, Lighthouse). This setting is closer to \emph{domain translation} than classical style transfer: the target style encodes real illumination and seasonal/weather statistics rather than purely painterly textures. The central requirement is that appearance may change substantially while \emph{multi-view geometry must remain stable}, otherwise correspondence matching and camera tracking degrade.

Our method is well-suited for this regime because it separates appearance adaptation from geometry preservation. The AdaIN-based stylization objective rapidly injects low-frequency domain characteristics (global color temperature, contrast, saturation, and mild texture statistics). In parallel, the multi-view constraints (SuperPoint/SuperGlue correspondences and depth loss) discourage view-dependent changes that would ``repaint'' or deform high-frequency structures. As a result, the stylizer behaves like a domain translator: it matches the target condition photometrically while retaining edges, corners, and fine details needed for consistent geometry.

\begin{figure}[p]
  \centering
  \includegraphics[width=\linewidth]{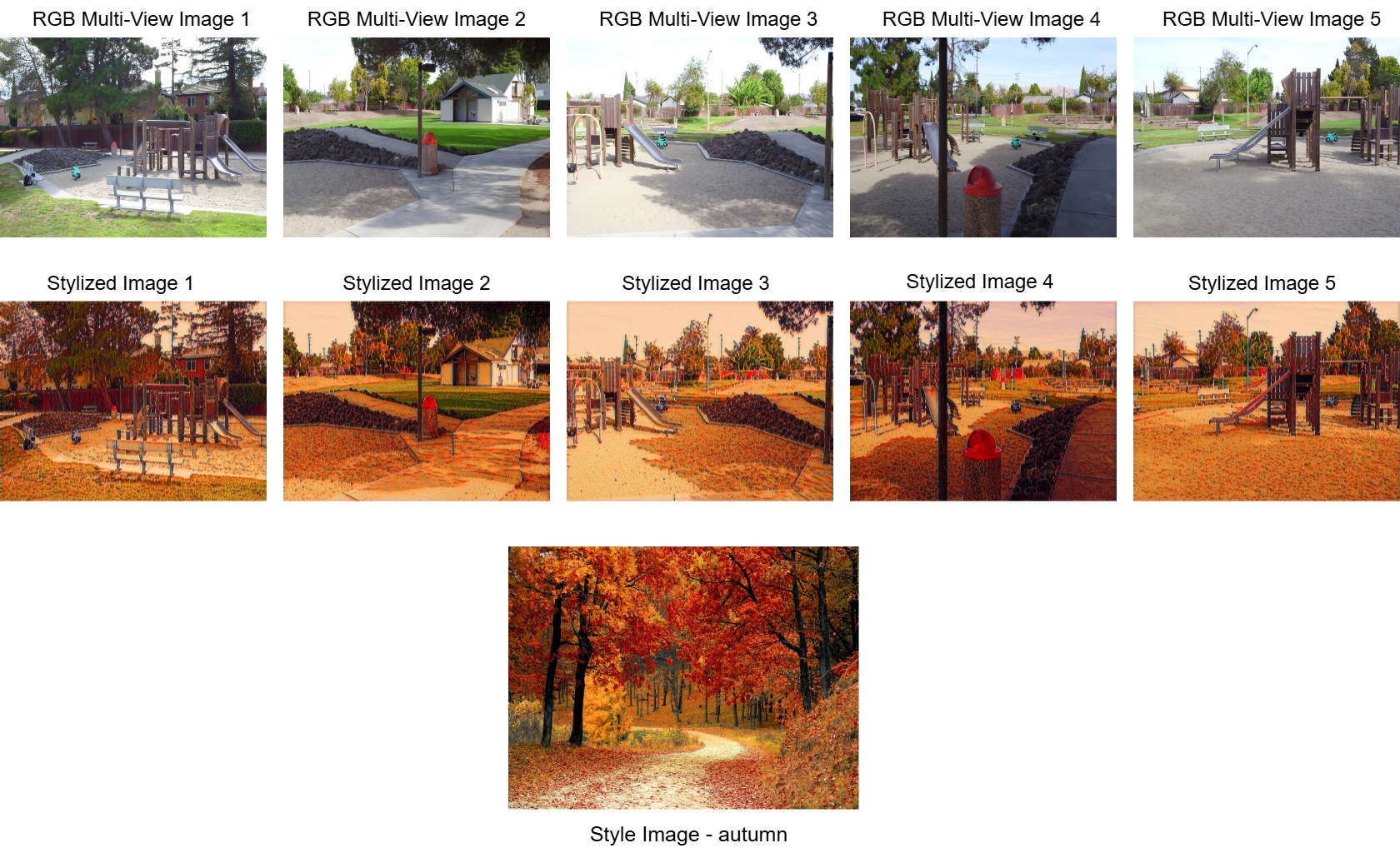}
  \caption{Real-world style transfer for \textbf{Autumn} on the \textbf{Playground} scene. The method applies a coherent seasonal appearance while preserving sharp geometric boundaries across views.}
  \label{fig:realstyle_autumn_playground}
\end{figure}

\paragraph{Autumn $\rightarrow$ Playground.}
As in Figure~\ref{fig:realstyle_autumn_playground}, the autumn translation yields a coherent warm palette with consistent seasonal tones across all viewpoints. Crucially, playground structures retain sharp silhouettes and stable high-frequency boundaries, indicating that the transformation acts primarily on appearance rather than geometry.

\begin{figure}[p]
  \centering
  \includegraphics[width=\linewidth]{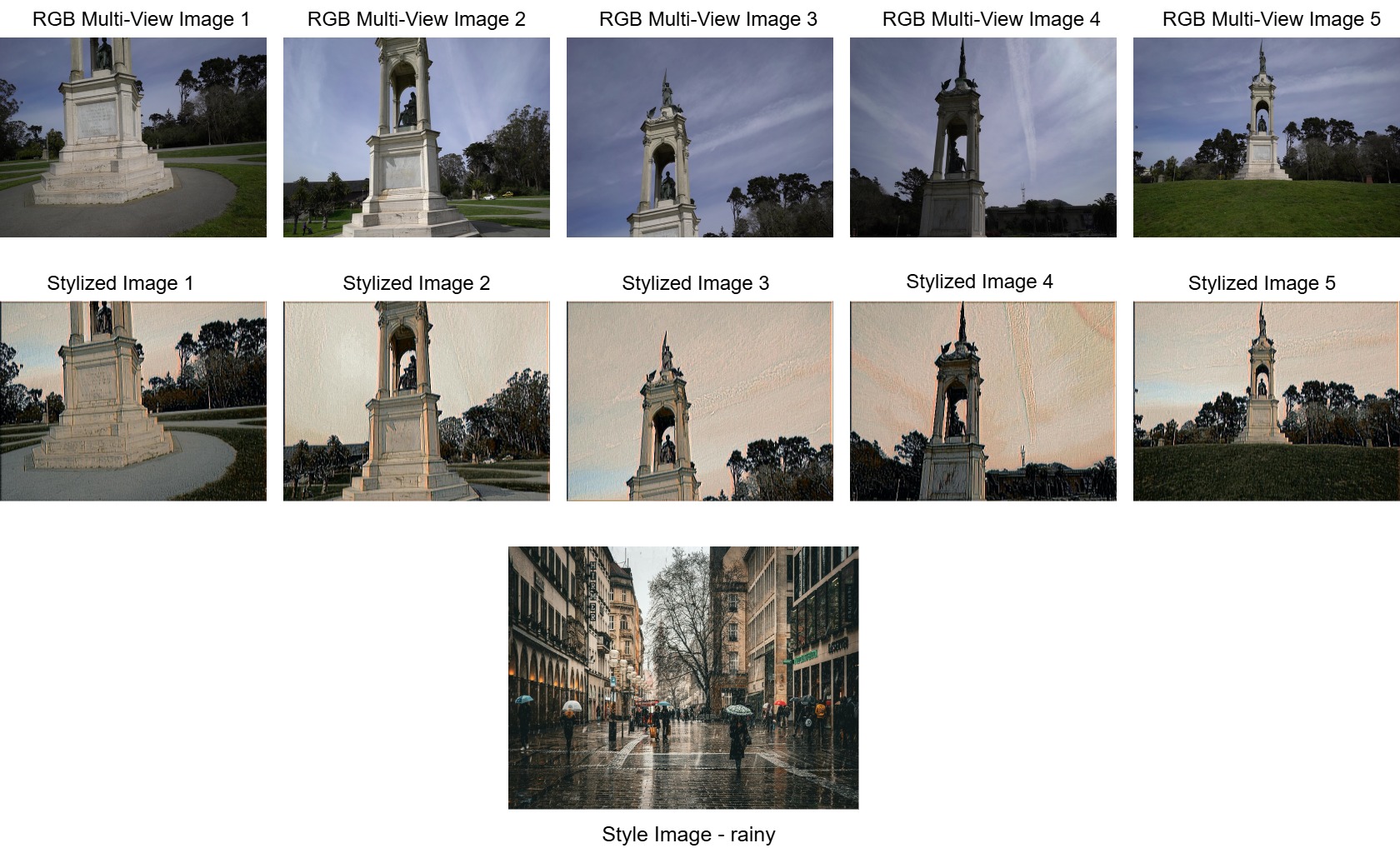}
  \caption{Real-world style transfer for \textbf{Rainy} on the \textbf{Francis} scene. The output exhibits an overcast, low-saturation domain shift while maintaining stable architectural structure across views.}
  \label{fig:realstyle_rainy_francis}
\end{figure}

\paragraph{Rainy $\rightarrow$ Francis.}
Similarly in Figure~\ref{fig:realstyle_rainy_francis}, the rainy style produces a stable overcast look with reduced saturation and contrast. Despite strong photometric shift, architectural contours remain crisp and consistent across views, supporting reliable cross-view correspondences.

\begin{figure}[p]
  \centering
  \includegraphics[width=\linewidth]{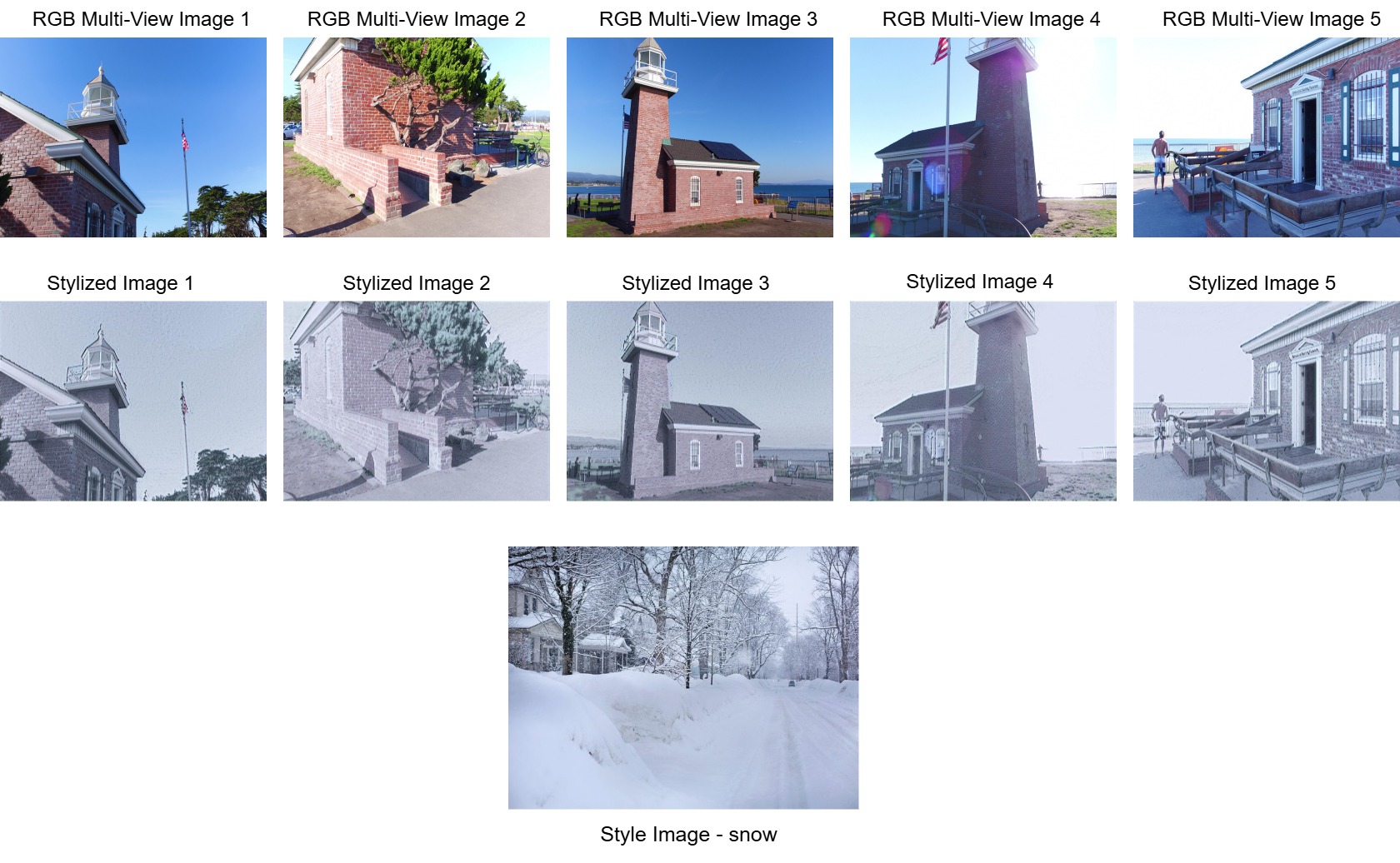}
  \caption{Real-world style transfer for \textbf{Snow} on the \textbf{Lighthouse} scene. The method achieves a cold, desaturated winter appearance while preserving key high-frequency geometric cues required for multi-view consistency.}
  \label{fig:realstyle_snow_lighthouse}
\end{figure}

\paragraph{Snow $\rightarrow$ Lighthouse.}
The snow translation as in Figure~\ref{fig:realstyle_snow_lighthouse}, introduces a cold, desaturated appearance while preserving dominant geometric cues such as brick edges, window frames, and railings. These stable high-frequency structures are critical for downstream pose estimation and reconstruction.

\clearpage

\clearpage

\clearpage

\section{Multi-Style Stylization with Conditional U-Net}
\label{sec:multistyle_cond_unet}

While our per-style, per-scene stylizer provides strong control over the stylization--geometry trade-off, it requires training a separate network for each target style. To study whether a \emph{single} stylizer can support \emph{multiple} styles without sacrificing multi-view consistency, we train a \textbf{conditional U-Net} that is explicitly conditioned on the desired style identity, using Conditional Instance Normalization. Concretely, the network receives an RGB view and a discrete style code and predicts the stylized output. This design forces the model to \emph{factorize} the problem into (i) \emph{content/geometry} extraction and preservation and (ii) \emph{style-specific} appearance modulation.

A key advantage of the U-Net backbone is its multi-scale skip connections, which directly carry high-frequency image components (edges, corners, and fine contours) from encoder to decoder. In practice, these high-frequency structures are the primary cues that anchor geometry across views: they define object boundaries and stable keypoints that reappear under viewpoint changes. By injecting the style code through conditional instance normalization, the network is encouraged to apply style predominantly through low- and mid-frequency channels (color statistics, shading, texture patterns), while preserving high-frequency content signals that are critical for view consistency and downstream pose estimation. As a result, the conditional U-Net can produce diverse stylizations \emph{within one model} while maintaining stable scene structure.

\subsection{Multi-style results on Train and Panther}
\begin{figure}[p]
  \centering
  \includegraphics[width=\linewidth]{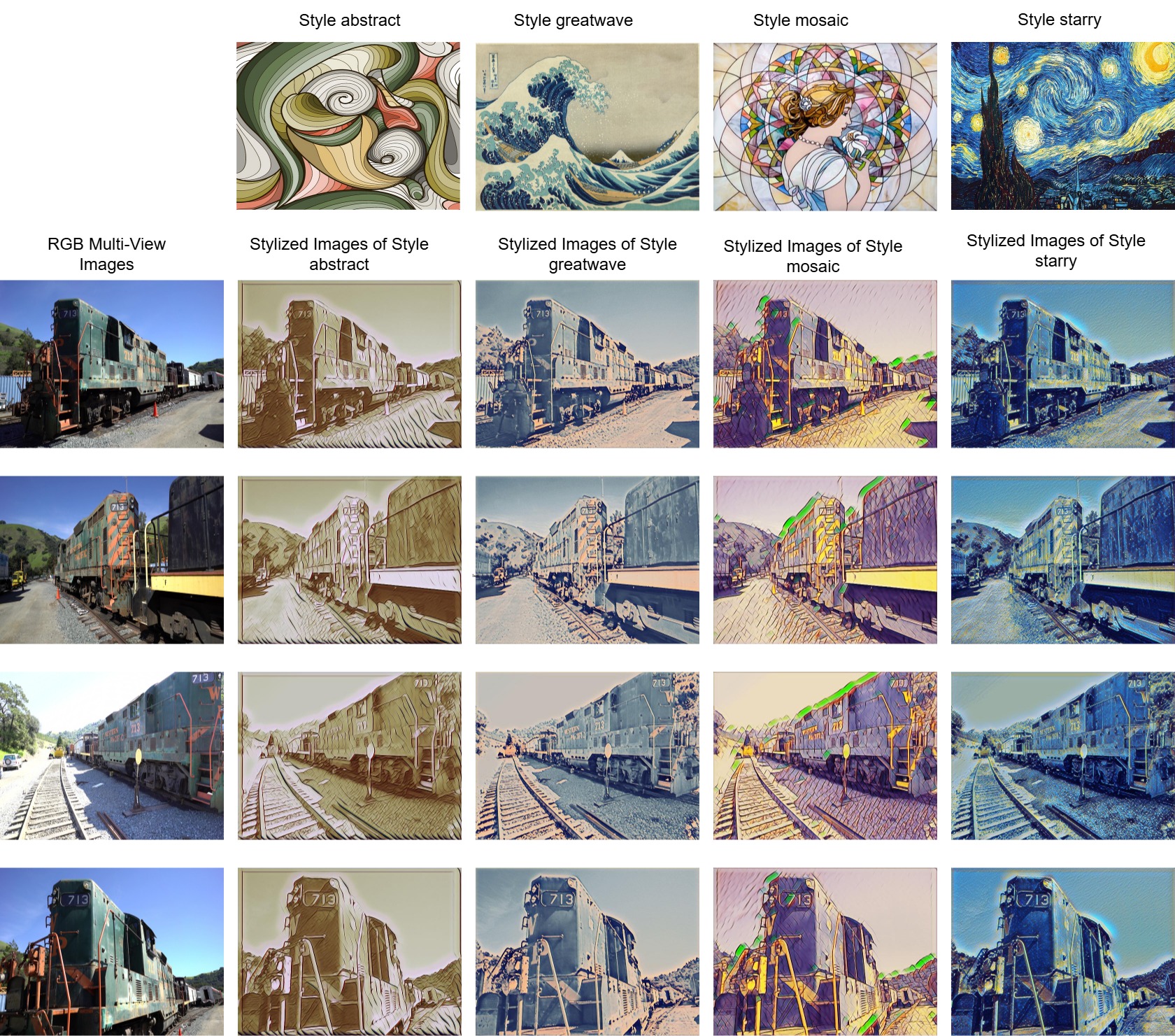}
  \caption{\textbf{Multi-style conditional U-Net on Train.} The network produces multiple target styles from a single model while preserving view-consistent geometry. Across styles, high-frequency structures (edges, corners, repeated details) remain stable, while global appearance (color/texture) changes according to the conditioning code.}
  \label{fig:multistyle_train}
\end{figure}

\begin{figure}[p]
  \centering
  \includegraphics[width=\linewidth]{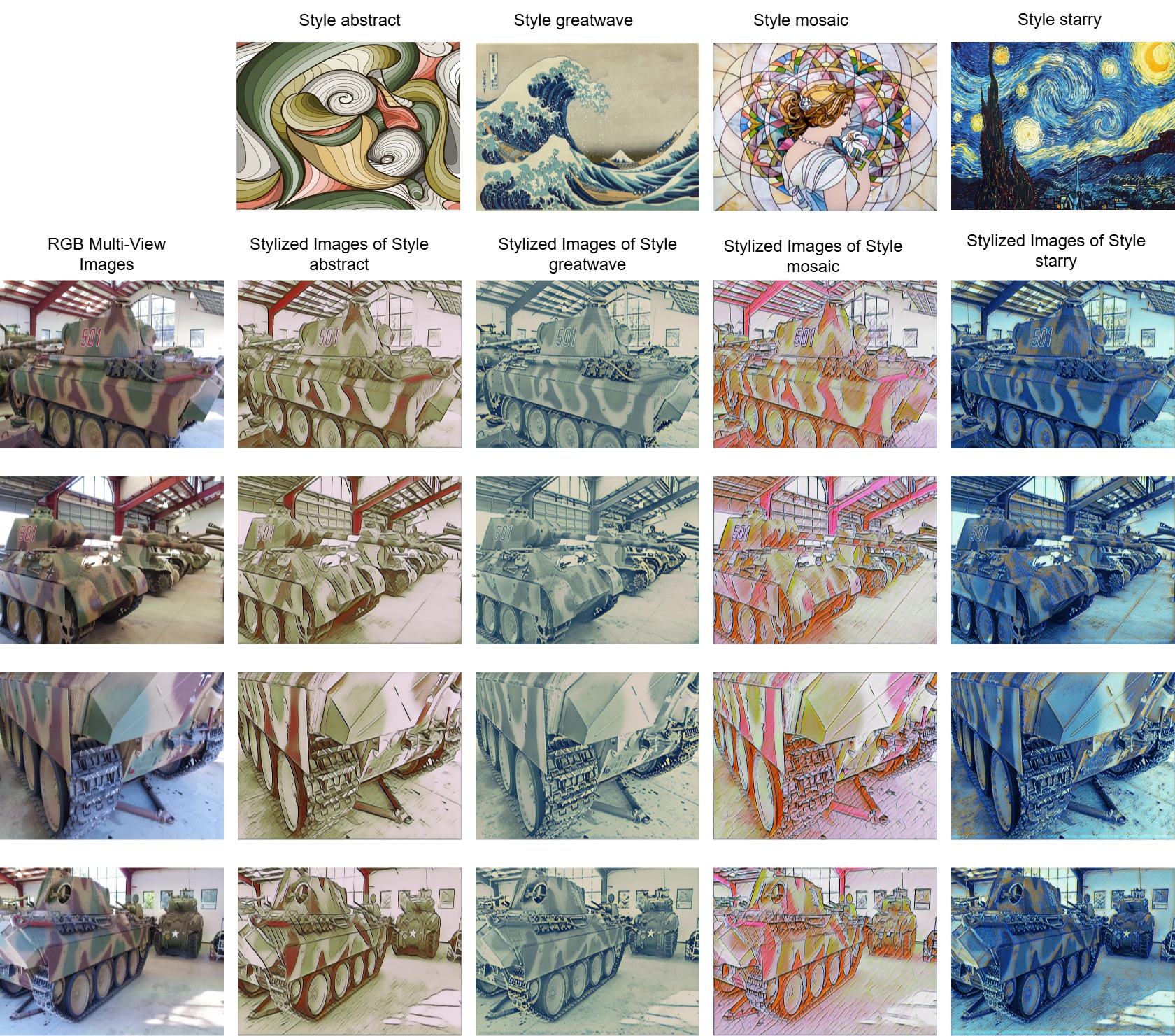}
  \caption{\textbf{Multi-style conditional U-Net on Panther.} Despite large stylistic differences (e.g., Mosaic vs.\ Starry), the turret silhouette, track boundaries, and wheel structures remain consistent across viewpoints, illustrating that the conditional model preserves geometry-bearing high-frequency cues while modulating style-dependent appearance.}
  \label{fig:multistyle_panther}
\end{figure}

Figures~\ref{fig:multistyle_train} and~\ref{fig:multistyle_panther} show multi-style stylization on the \emph{Train} and \emph{Panther} scenes for four distinct styles (abstract, greatWave, mosaic, starry). Across styles, the outputs remain visually consistent across viewpoints: straight edges (e.g., train chassis lines, turret/track boundaries), repeated structures (wheels, track elements), and prominent contours are preserved with minimal view-dependent warping. Importantly, although each style induces a different global appearance, from line-art abstractions to heavy color remapping and textured brush patterns, the \emph{geometry-bearing details} remain stable. This behavior aligns with the intended intention of the loss selections: the model learns a shared geometry-preserving representation which is consistent across views and uses the conditioning signal primarily to steer the stylization manifold.

\subsection{Cross-scene generalization: Horse $\rightarrow$ Lighthouse}
\begin{figure}[p]
  \centering
  \includegraphics[width=\linewidth]{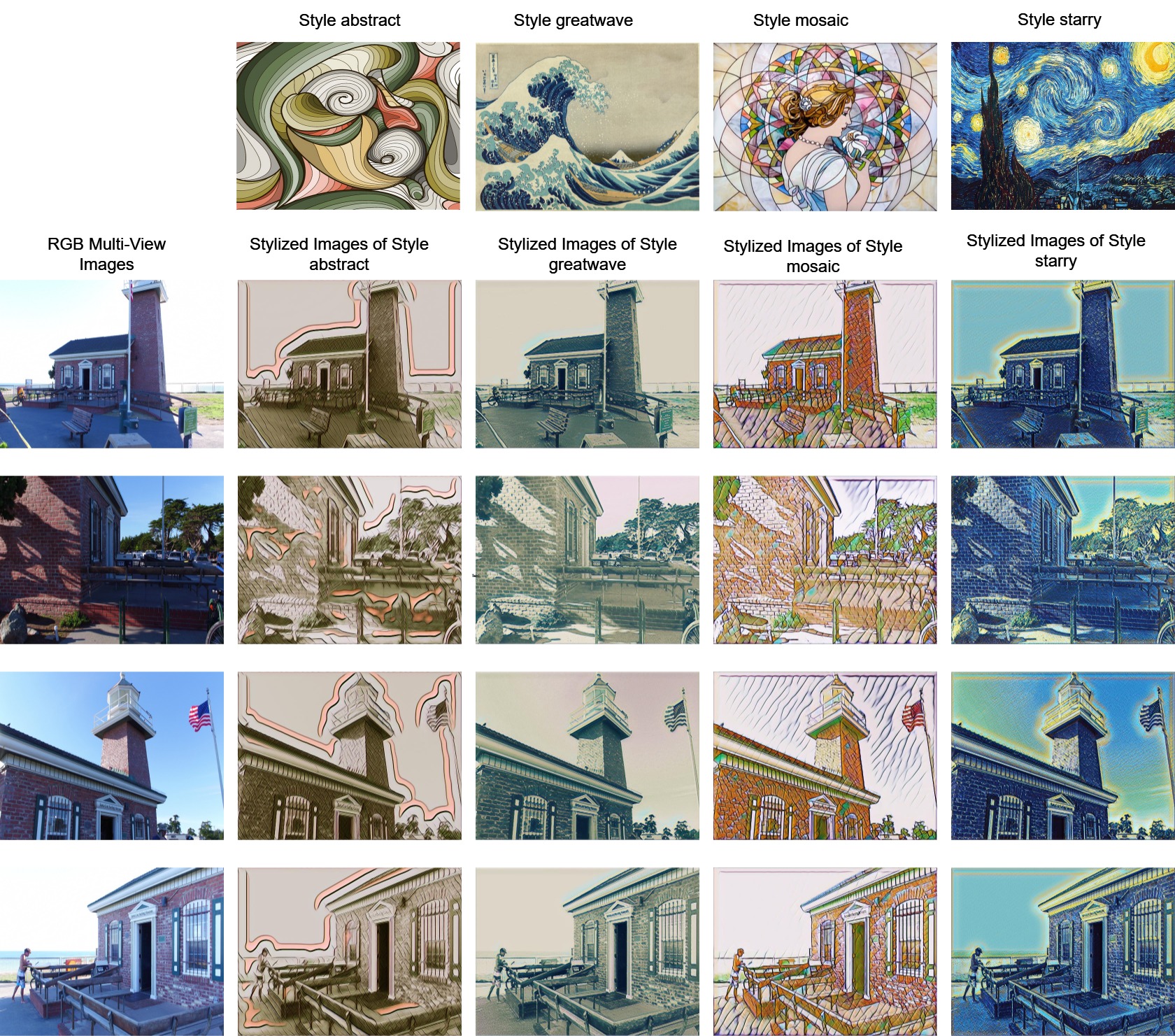}
  \caption{\textbf{Cross-scene generalization (trained on Horse, tested on Lighthouse).} The conditional U-Net transfers the target styles to an unseen scene while preserving lighthouse geometry (tower edges, facade boundaries, and walkway structure).}
  \label{fig:multistyle_horse_to_lighthouse}
\end{figure}

To test whether the conditional U-Net learns \emph{transferable} style control (rather than memorizing a particular scene), we train the multi-style model on one scene (Horse) and evaluate it on an unseen scene (Lighthouse). Figure~\ref{fig:multistyle_horse_to_lighthouse} demonstrates that the model still generates the target styles with coherent appearance while preserving lighthouse-specific geometric structures (e.g., tower edges, brick boundaries, railing and walkway lines). The persistence of crisp contours and consistent structural alignment across multiple views indicates that the conditional U-Net has learned (i) a style-conditioned rendering behavior that generalizes beyond the training scene and (ii) a stable prior to retain high-frequency, geometry-defining components even under domain shift. In other words, the model does not only learn ``what the style looks like''; it also learns \emph{how to apply} that style while respecting structural content cues that support multi-view consistency.


\clearpage

\clearpage

\begin{figure}[p]
  \centering
  \includegraphics[width=\linewidth]{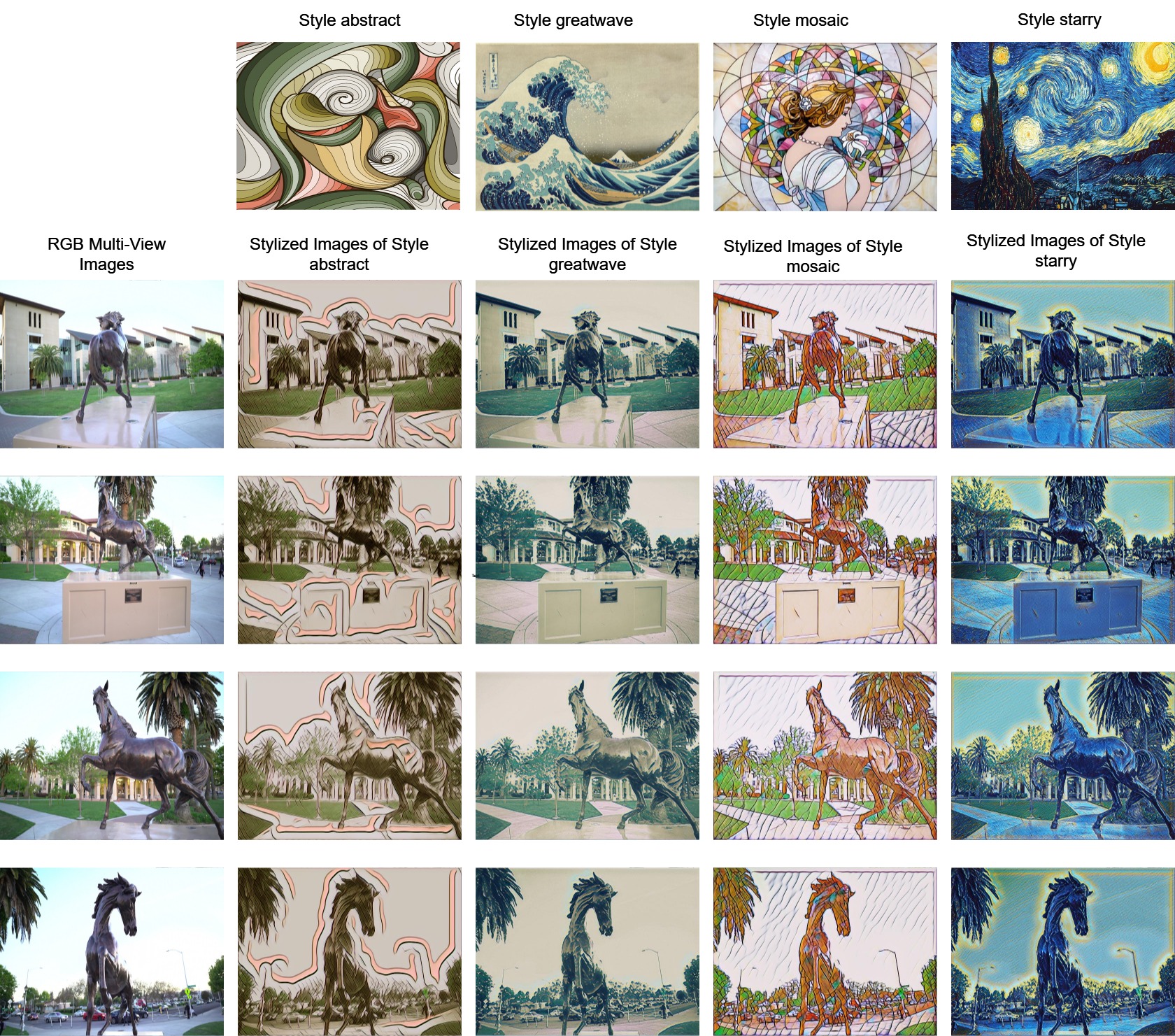}
  \caption{\textbf{Multi-style conditional U-Net on Horse (training scene).} Stylizations remain coherent across views, with stable statue contours and consistent background layout across different style conditions.}
  \label{fig:multistyle_horse}
\end{figure}
\clearpage

\clearpage

\section{Depth Consistency of Stylized Outputs}
\label{sec:depth_consistency_vis}

\begin{figure*}[p]
\centering
\includegraphics[width=0.95\textwidth]{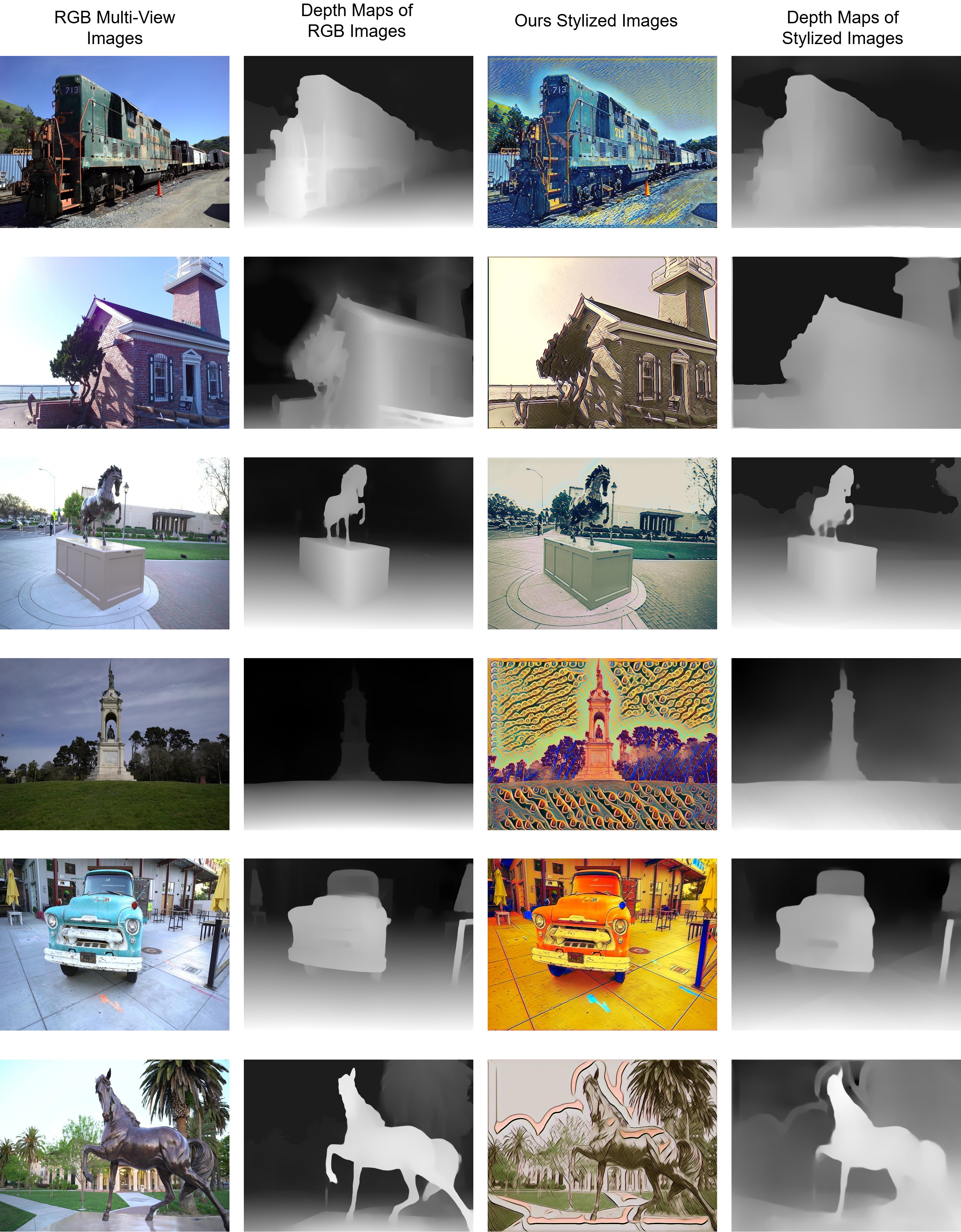} 
\caption{\textbf{Depth consistency under stylization.} We show (left to right) the RGB multi-view inputs, monocular depth predictions of the RGB images, our stylized outputs, and monocular depth predictions of the stylized images.}
\label{fig:depth_compare}
\end{figure*}

To verify that our stylization remains \emph{geometrically usable} (and not only perceptually pleasing), we visualize monocular depth predictions on both the original RGB views and their stylized counterparts (Fig.~\ref{fig:depth_compare}). Depth maps are computed with a frozen off-the-shelf depth estimator (DPT/MiDaS), and our method explicitly enforces depth stability through the depth-consistency objective $\mathcal{L}_{Depth}$: the stylizer is penalized if stylization alters the predicted depth structure. Intuitively, this encourages the network to treat style as an \emph{appearance transformation} (color/texture/stroke statistics) rather than a geometric deformation that bends edges, shifts silhouettes, or changes surface ordering.

\paragraph{Scene-wise qualitative consistency.}
In the \textit{Train} example, despite the strong painterly texture and color shifts, the locomotive silhouette and the depth step between the train body and the background remain nearly unchanged, showing that style is layered on top of stable geometry rather than redefining it. In the \textit{Lighthouse} and \textit{Francis} scenes, straight architectural edges remain structurally coherent in depth: the tower is still recovered as a dominant foreground structure with consistent separation from sky and surrounding vegetation/ground. The \textit{Horse} statue examples further highlight boundary preservation: the statue contours and the pedestal depth are maintained, indicating that stylization does not erode thin structures or merge foreground elements into the background. Finally, even in the \textit{Truck} scene, the depth shape of the vehicle remains consistent, demonstrating that our constraints do not require aggressive stylization to maintain depth stability, they enforce it whenever style is applied.

\paragraph{Why this aligns with our methodology.}
These depth visualizations directly reflect the complementary roles of our geometry-aware losses. $\mathcal{L}_{Depth}$ anchors the stylization to a stable, view-consistent notion of scene layout by discouraging changes that would perturb depth prediction (typically caused by edge bending, silhouette drift, or introducing view-dependent texture cues that mimic shape). Meanwhile, the SuperPoint/SuperGlue correspondence loss $\mathcal{L}_{SG}$ protects \emph{repeatable local structure} (corners, junctions, textured keypoints) across views. Together, they reduce view-dependent “texture swimming” and preserve both (i) the global structure captured by depth and (ii) the local high-frequency evidence needed for matching. This is precisely the combination required for downstream usability: stylized images should still support stable tracking, correspondence estimation, and reconstruction.

\paragraph{Practical implication for domain translation.}
In real deployments (AR/VR capture, scene documentation, robotics), one often wants to translate appearance (season, lighting mood, artistic style) without collecting new multi-view training data in that target domain. The depth-consistency results indicate that our method can perform such domain translation while keeping geometry intact: stylization modifies the image domain, but the underlying 3D layout remains interpretable and consistent for reconstruction pipelines.

\chapter{Conclusion}

This thesis studied \emph{multi-view artistic stylization} under a constraint that is often ignored by image-only style transfer: the stylized outputs must remain \emph{geometrically usable}. When stylization is applied independently per view, even visually appealing results can break cross-view correspondences, disrupt monocular depth cues, and ultimately degrade 3D reconstruction. To address this, we proposed a lightweight, scene wise training framework that explicitly balances stylization strength with 3D-consistency requirements, without assuming access to camera poses or explicit geometry during training.

Our methodology learns a feed-forward stylizer $g_{\theta}$ using an AdaIN based feature statistics objective for efficient style transfer, while introducing two geometry-preserving constraints that directly target 3D failure modes. First, we enforced cross-view correspondence stability using a SuperPoint/SuperGlue-based loss that encourages matched local structures to remain descriptor-consistent under stylization. This loss provides a pose-free, correspondence-driven geometric signal that is tightly aligned with the requirements of SfM/SLAM systems. Second, we introduced a depth preservation loss using a frozen MiDaS/DPT model to penalize stylization-induced distortions in monocular depth predictions; a simple dataset-to-style color alignment step mitigated depth-model domain shift and improved stability. Because strong geometric constraints can hinder early stylization learning, we adopted a warmup-and-ramp scheduling strategy that stages the optimization: the stylizer first learns style statistics, and only then is progressively constrained toward geometry and depth consistency.

We evaluated the proposed approach with a protocol that goes beyond single-image metrics. Static metrics (CHD and DSD) captured style adherence and per-view structural faithfulness, while dynamic SLAM-based metrics derived from DROID-SLAM measured whether the \emph{entire} stylized set remains reconstructable. In particular, trajectory alignment errors (ATE/RTE) and point-cloud Chamfer distance quantified camera tracking stability and 3D geometric agreement between reconstructions from original and stylized sequences. Across ablations, adding $\mathcal{L}_{SG}$ and $\mathcal{L}_{Depth}$ consistently improved geometric consistency, most strongly reflected in reduced SLAM trajectory errors and lower Chamfer distances, confirming that correspondence- and depth-based constraints translate into measurable downstream 3D benefits. Comparisons against MuVieCAST further showed that, while our approach may trade off some pure style alignment in certain cases, it delivers markedly stronger reconstruction stability and structural integrity, especially in challenging scenes where view-dependent artifacts are amplified.

Finally, we extended the framework toward multi-style learning using a conditional U-Net with Conditional Instance Normalization, demonstrating that a single model can support multiple styles for a scene and can transfer stylization to unseen scenes while maintaining high-frequency structure that is critical for geometry. We also showed that the method can be applied with photorealistic images (e.g., seasonal or weather domains such as snow/rain/autumn), highlighting its potential for domain translation scenarios where maintaining geometry is essential for downstream perception and reconstruction.

Overall, the central contribution of this thesis is a practical, camera pose-free formulation of multi-view stylization that is explicitly optimized for \emph{geometric consistency}, together with an evaluation protocol that measures this consistency directly through SLAM trajectories and reconstructed 3D geometry. The results support the thesis claim that stylization can be made substantially more compatible with 3D pipelines when correspondence stability and depth cues are treated as first-class objectives rather than side effects of per-image structure preservation.

\paragraph{Limitations and future work.}
Despite strong improvements in 3D consistency, the method still exhibits a style--geometry trade-off in some regimes: highly abstract styles, low-texture regions, or extreme appearance shifts can reduce match reliability and constrain stylization richness. Future work could incorporate more robust or dense correspondence families (e.g., detector-free matchers), confidence-aware match filtering, or multi-anchor sampling strategies, and could explore joint learning of a style-robust feature space that remains stable under stronger stylization.



\microtypesetup{protrusion=false}
\listoffigures{}
\listoftables{}
\microtypesetup{protrusion=true}
\printbibliography{}

\end{document}